%% file: acl_latex.tex
\pdfoutput=1
\documentclass[11pt]{article}

\usepackage[final]{acl}

\usepackage{times}
\usepackage{latexsym}
\usepackage[T1]{fontenc}
\usepackage[utf8]{inputenc}
\usepackage{microtype}
\usepackage{inconsolata}
\usepackage{graphicx}
\usepackage{pifont}
\usepackage{booktabs}
\usepackage{amsmath}
\usepackage{amsfonts}
\usepackage{amsthm}
\usepackage{mathtools}
\usepackage{multirow}
\usepackage{subfig}

\title{Identifying Semantic Induction Heads to Understand In-Context Learning}

\author{Jie Ren\thanks{This work is done during the internship at Shanghai Artificial Intelligence Laboratory.}\textsuperscript{1,2}, Qipeng Guo\thanks{Corresponding author.}\textsuperscript{2}, Hang Yan\textsuperscript{2,4}, Dongrui Liu\textsuperscript{1},
\\
\textbf{
Quanshi Zhang\textsuperscript{1}, Xipeng Qiu\textsuperscript{3}, Dahua Lin\textsuperscript{2,4}}
\\
\\
\textsuperscript{1}Shanghai Jiao Tong University
~~\textsuperscript{2}Shanghai Artificial Intelligence Laboratory\\
\textsuperscript{3}Fudan University
~~\textsuperscript{4}The Chinese University of Hong Kong
\\
\texttt{\{ariesrj, drliu96, zqs1022\}@sjtu.edu.cn},\\ \texttt{\{guoqipeng,yanhang,lindahua\}@pjlab.org.cn}, \texttt{xpqiu@fudan.edu.cn}
}

\begin{document}
\maketitle

\begin{abstract}
Although large language models (LLMs) have demonstrated remarkable performance, the lack of transparency in their inference logic raises concerns about their trustworthiness. To gain a better understanding of LLMs, we conduct a detailed analysis of the operations of attention heads and aim to better understand the in-context learning of LLMs. Specifically, we investigate whether attention heads encode two types of relationships between tokens in natural languages: the syntactic dependency parsed from sentences and the relation within knowledge graphs.  We find that certain attention heads exhibit a pattern where,  when attending to head tokens, they recall tail tokens and increase the output logits of those tail tokens. More crucially, the formulation of such semantic induction heads has a close correlation with the emergence of the in-context learning ability of language models.
The study of semantic attention heads advances our understanding of the intricate operations of attention heads in transformers, and further provides new insights into the in-context learning of LLMs.
\end{abstract}

\section{Introduction}

In recent years, the transformer-based large language models (LLMs)~\citep{kaplan2020scaling,brown2020language,touvron2023llama,bubeck2023sparks} have rapidly emerged as one of the mainstreams in the field of natural language processing (NLP). While these models demonstrate emergent abilities as they scale~\cite {brown2020language,wei2022emergent}, they become less interpretable due to the vast number of parameters and complex architectures, which emphasizes LLMs' safety and trustworthiness~\citep{carlini2021extracting,manakul-etal-2023-selfcheckgpt,ren2024exploring}.
Thus, beyond classical gradient-based explanations~\citep{simonyan2013deep,li2015visualizing}, and perturbation-based explanations~\citep{ribeiro2016should,lundberg2017unified,sundararajan2017axiomatic}, recent studies in mechanistic interpretability~\citep{cammarata2020thread, elhage2021mathematical} attempt to reverse engineer the computations in transformers (particularly attention layers).

\begin{figure*}
    \centering
    \setlength{\abovecaptionskip}{3pt}
    \includegraphics[width=\linewidth]{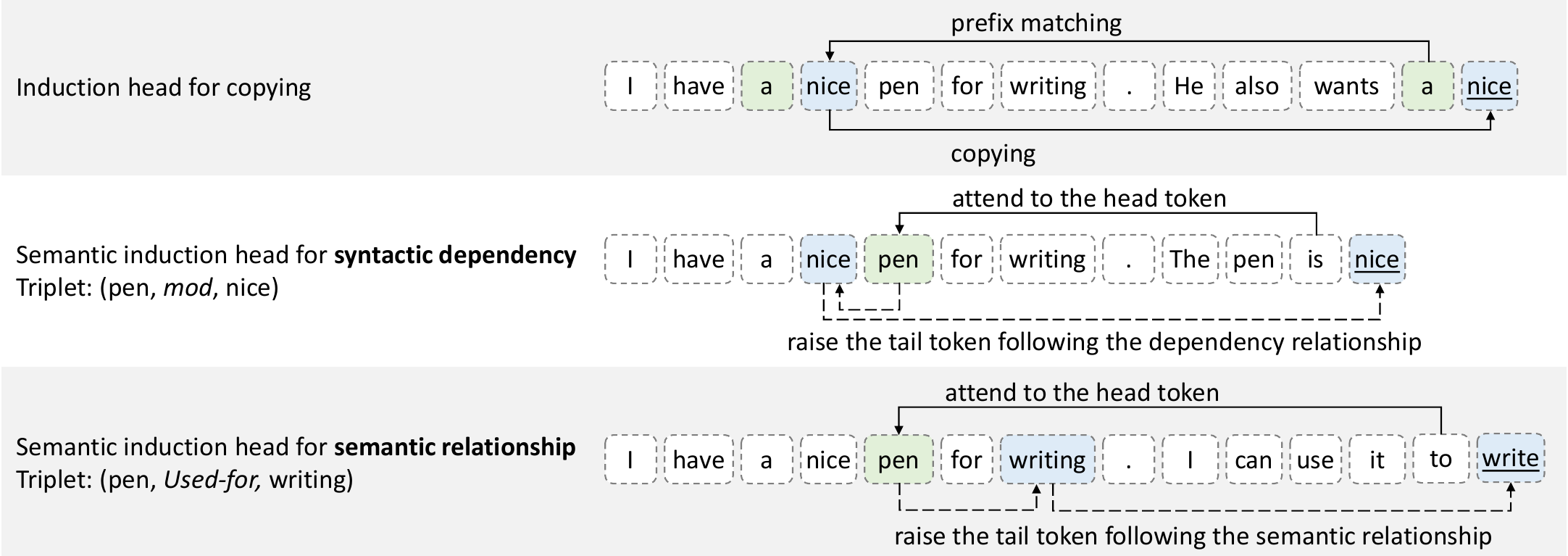}
    \caption{Induction heads and semantic induction heads. For the sequence ``... a nice ... a'', an induction head finds a place where the current token ``a'' occurred, attends to its next token ``nice'' (\textit{prefix matching}), and then copies ``nice'' to the output \textit{(copying}).
    In contrast, the semantic induction head raises the output logits of tail tokens (``nice'' in the \textit{mod} dependency and ``writing'' in the \textit{Used-for} relation) when attending to the head token ``pen''.}
    \label{fig: induction_head}
\end{figure*}

The mechanistic interpretability on transformer language models was first performed by \citet{elhage2021mathematical}. They disentangle two circuits from the operation of each attention head in transformers: Query-Key circuit (determines which token the head prefers to attend to) and Output-Value circuit (determines how the head affects the output logits of the next token).
Then, \citet{elhage2021mathematical} discover that some attention heads prefer to search for a previous occurrence of the current token in context and copy the next token associated with that occurrence, as shown in Figure~\ref{fig:  induction_head}.
The attention heads performing such operations are termed \textit{induction heads}.
Taking a step further, \citet{olsson2022context,bansal-etal-2023-rethinking} have discovered that the presence of induction heads has a close correlation with the in-context learning (ICL) ability of LLMs. This finding highlights the importance of understanding the behavior of attention heads to the overall learning capabilities of LLMs.

On the other hand, semantic relationships have a vital importance on natural language understanding and processing.
However, \citet{elhage2021mathematical} only focus on whether the attention heads copy the attended token, without studying semantic relationships between tokens.
Another major limitation of previous studies is that \citet{olsson2022context} does not explain the popular few-shot in-context learning schema.
Instead, they study the loss decreasing along with the increase of token indices. This setting does not fully capture the complete ability of LLMs to learn from the context.

In this work, beyond simple copying, we delve deeper into high-level relationships encoded in attention heads.
We focus on two types of relationships: (1) syntactic dependencies in the sentence and (2) semantic relationships between entities.
Please refer to Figure~\ref{fig: induction_head} for examples.
Each relation is represented as a triplet: (\textit{head}, \textit{relation}, \textit{tail}). We find that when attending to head tokens, some attention heads prefer to raise the output logits of tail tokens associated with specific relations.
Such attention heads encoding semantic relationships are termed \textit{semantic induction heads}.
Unlike conventional induction heads, semantic induction heads learn and leverage the semantic relationships between words to infer the output, thereby providing a better understanding of the behavior of networks.

Inspired by the study of induction heads and in-context learning, we further explore the correlation between semantic induction heads and in-context learning.
We first categorize the in-context learning ability into three basic levels: loss reduction, format compliance, and pattern discovery. These three levels progressively increase in difficulty, with each subsequent level building upon the achievements of the previous one.
The experimental results are consistent with our hypothesis, demonstrating the emergence of three levels of ICL in a sequential manner. Specifically, we observe the emergence of loss reduction from the beginning of the training, followed by the emergence of format compliance at around 1.6B tokens, and finally, the emergence of pattern discovery after training on approximately 4B tokens. 
Moreover, we find semantic induction heads mainly emerge around the same time as pattern discovery.
Based on this finding, we infer that the emergence of semantic induction heads plays a crucial role in facilitating the ICL of LLMs.

Our contributions can be summarized as follows. 

\textbullet~ We unveil the existence of semantic induction heads in LLMs that extract semantic relationships within the context. This discovery deepens the study of mechanistic interpretability and enhances our understanding of transformer-based models.

\textbullet~ To study the ICL in LLMs, we categorize it into three different levels and observe the gradual emergence of different levels of ICL during the early training stage of LLMs.

\textbullet~ Through a meticulous analysis of early checkpoints in the training of LLMs, we establish a close correlation between semantic induction heads and the occurrence of ICL.

\section{Related Works}

In this section, we provide an overview of recent advancements in the interpretability of neural networks, particularly mechanistic interpretability.
On the other hand, previous studies~\citep{petroni-etal-2019-language, zhang-etal-2022-probing} have also explored the topic of semantic relationships in models. However, our study distinguishes itself by focusing on mechanistic interpretability. 

Previous studies in interpretability can be roughly categorized into the following four types: estimating the attribution of input features to the network output~\citep{ribeiro2016should, sundararajan2017axiomatic,lundberg2017unified,yang-etal-2023-local,modarressi-etal-2023-decompx}, discovering interaction patterns between input features~\citep{ren2021towards,ren2023defining, liu2024towards, zhou2024explaining}, extracting concepts from intermediate-layer features~\citep{kim2018tcav,fel2023craft, qian2024towards}, and designing self-explainable architectures~\citep{oscar2018deep, das-etal-2022-enolp}.
As transformer-based models become mainstream, recent works focus on understanding the attention mechanism. The most direct approach is to visualize the attention using bipartite graphs~\citep{vig-2019-multiscale,yeh2023attentionviz} or heatmaps~\citep{park2019sanvis}.
Another line of research aims to reverse engineer the operation of attention heads, called mechanistic interpretability~\citep{cammarata2020thread, elhage2021mathematical}.

\citet{elhage2021mathematical} proposed the circuit analysis (introduced in Section~\ref{subsec: preliminary}) to examine the operation of attention heads, and they found induction heads in attention-only models.
\citet{olsson2022context} further investigated the correlation between the formation of induction heads and ICL.
\citet{bansal-etal-2023-rethinking} observed an overlap between the set of induction heads and the set of important attention heads for ICL. 
Using circuit analysis, \citet{wang2023interpretability} also found some attention heads performing the function of identifying/removing names in the indirect object identification task.
Other studies~\citep{lieberum2023does,geva-etal-2023-dissecting,mohebbi-etal-2023-quantifying} intervened the attention or FFN layers to study their functions. 
In this paper, we leverage the circuit analysis to investigate semantic relationships in attention heads.

\section{Semantic Induction Head}



\paragraph{Preliminary.}
\label{subsec: preliminary}
\citet{elhage2021mathematical} rewrite the operation of a multi-head attention (MHA) layer containing $h$ attention heads as follows.
\begin{equation}
\begin{aligned}
\label{eq: circuit}
    & {\sum}_{h=1}^H \textit{softmax}\left(\boldsymbol{x}W^h_q(\boldsymbol{x}W^h_k)^T/\sqrt{d_h}\right)\boldsymbol{x}W^h_vW_o^h\\
    = & {\sum}_{h=1}^H \textit{softmax}\left(\boldsymbol{x}W^h_{QK}\boldsymbol{x}^T/\sqrt{d_h}\right)\boldsymbol{x}W^h_{OV}
\end{aligned}
\end{equation}
where $\boldsymbol{x}\!=\![x^T_1,x^T_2,\ldots,x^T_N]^T\in\mathbb{R}^{N\times d}$ denotes the embedding sequence, and $x_i=t_i W_e\in\mathbb{R}^{1\times d}$ is the embedding of the $i$-th input token $t_i$.
$W_e\in\mathbb{R}^{\vert \mathcal{V}\vert\times d}$ denotes the embedding layer over a vocabulary $\mathcal{V}$.
$W^h_q,W^h_k,W^h_v\in\mathbb{R}^{d\times d_h}$ denote the query, key, and value transformations in the $h$-th attention head. The output transformation $W_o$ can be decomposed as $W_o\!=\![(W_o^1)^T ~  (W_o^2)^T ~ \ldots ~ (W_o^H)^T]^T$, where $W_o^h\in\mathbb{R}^{d_h\times d}$.

In Equation~\eqref{eq: circuit}, $W^h_{QK}=W^h_q(W^h_k)^T$, termed the Query-Key (QK) circuit, is responsible for computing the attention pattern of the head, thus determining the head prefers to attend to which token.
On the other hand, the matrix $W^h_{OV}=W^h_vW_o^h$, termed the Output-Value (OV) circuit, computes the independent output of each head at the current token regardless of the attention pattern.
The output of the OV circuit can be projected back to the vocabulary as $\boldsymbol{x}W^h_{OV}W_u$ by the unembedding transformation $W_u\in\mathbb{R}^{d\times \vert \mathcal{V}\vert}$.
The projected vector represents the influence of the attention head on the output.
Importantly, according to \citep{elhage2021mathematical}, both the QK circuit and OV circuit are directly performed on input embeddings, facilitating the understanding of operations in attention heads.

Based on the above decomposition, \citet{elhage2021mathematical} identify a specific behavior in attention heads, which they refer to as \textit{induction heads}. They observe this behavior in attention heads when presented with sequences like ``[A] [B] $\cdots$ [A]''. In these induction heads, the QK circuit causes the attention head to attend to the token [B], which appears next to the previous occurrence of the current token [A]. This behavior is termed prefix matching.
Then, the OV circuit increases the output logit of the attended token [B], termed \textit{copying}.
This mechanism is shown in Figure~\ref{fig: induction_head}.

\paragraph{Main experimental setup.}
We use the open-sourced InternLM2-1.8B\footnote{\url{https://huggingface.co/internlm}}, which contains 24 layers and each layer consists of 16 attention heads.
We use the Abstract GENeration DAtaset (AGENDA)\footnote{\url{https://github.com/rikdz/GraphWriter}}~\citep{koncel-kedziorski-etal-2019-text} for testing because it contains well-annotated relations between entities. The test set of the AGENDA dataset has a total of 1,000 samples, each consisting of a knowledge graph and a corresponding paragraph that describes relations in the knowledge graph.
For syntactic dependencies, we split paragraphs in the AGENDA dataset into individual sentences, and use spaCy~\citep{Honnibal2020spaCy} to extract dependencies between tokens.

\subsection{Syntactic Dependency in Attention Heads}
\label{subsec: syntactic_dependency}
In this section, we explore whether attention heads encode more complex knowledge beyond copying. We first study a basic pairwise relation inherent in natural language, syntactic dependency, representing the grammatical structure of a sentence.

\begin{figure*}
    \setlength{\abovecaptionskip}{1pt}
    \setlength{\belowcaptionskip}{-6pt}
    \centering
    \begin{minipage}{\linewidth}
        \centering
        \includegraphics[width=0.245\linewidth]{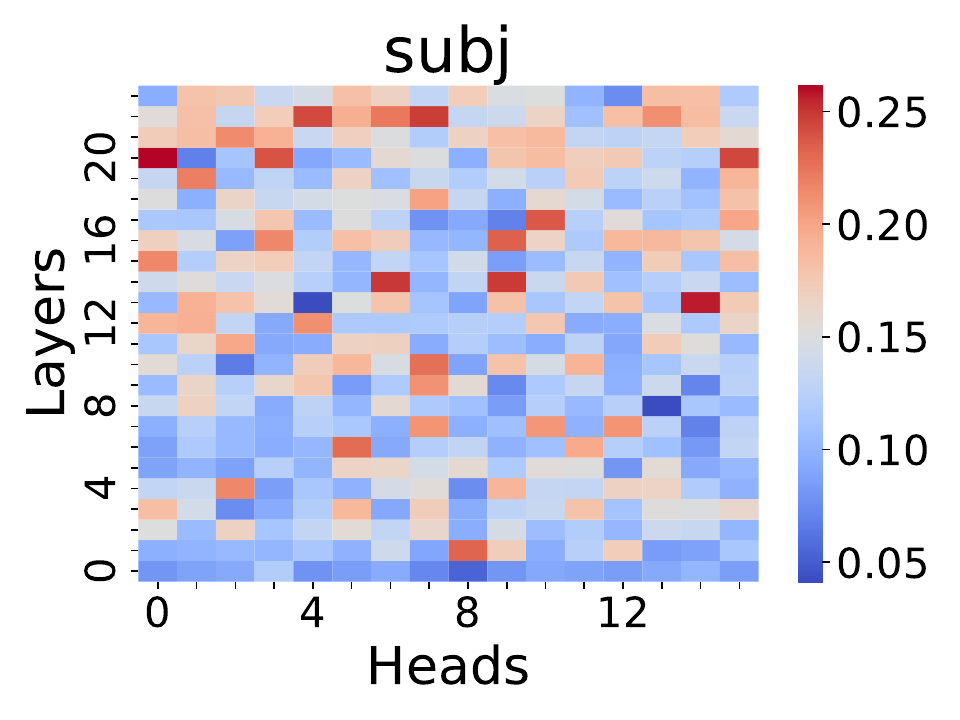}
        \hfill
        \includegraphics[width=0.245\linewidth]{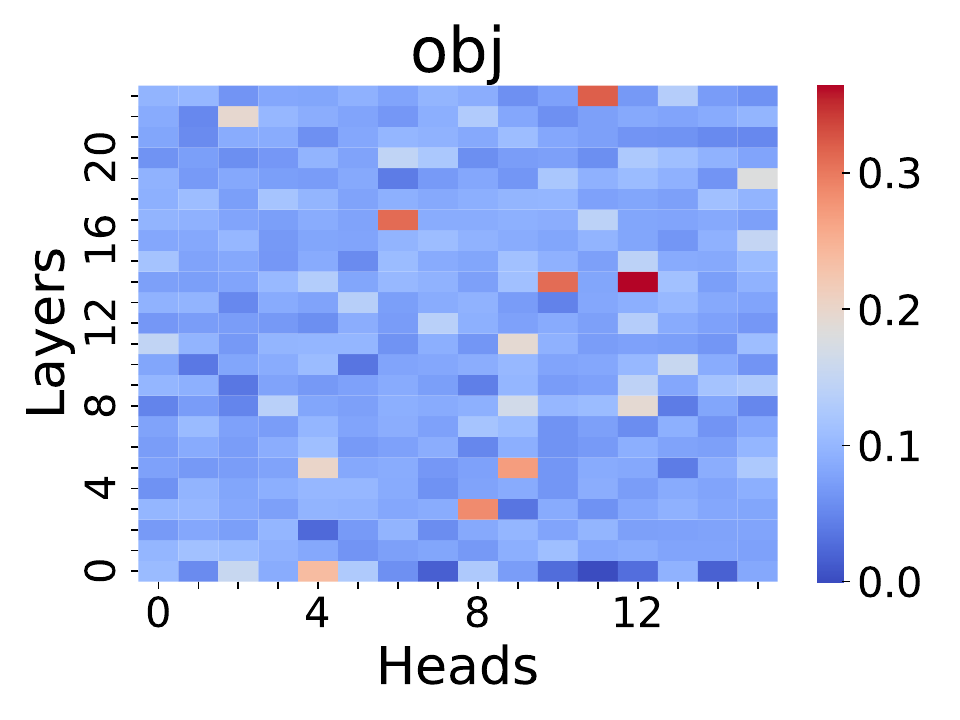}
        \hfill
        \includegraphics[width=0.245\linewidth]{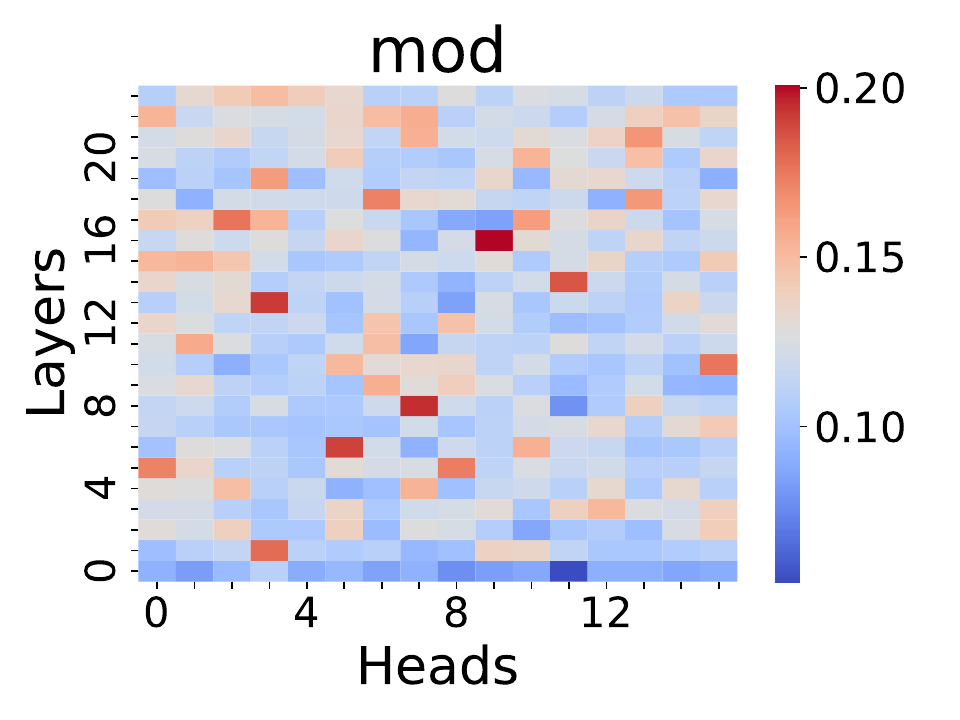}
        \hfill
        \includegraphics[width=0.245\linewidth]{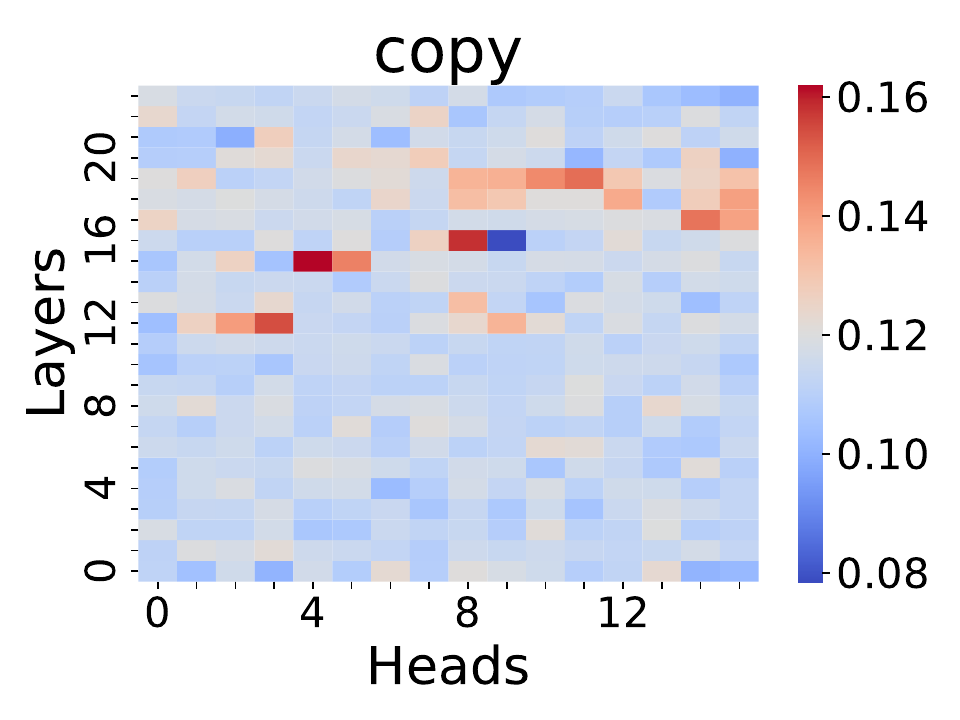}
    \end{minipage}
    \caption{Heatmaps of the average relation index of attention heads for syntactic dependency between tokens and the heatmap of the average copying score~\cite{bansal-etal-2023-rethinking} of attention heads.}
    \label{fig: vis_of_dep_score}
\end{figure*}

We focus on three frequent types of syntactic dependencies: subject-predicate (\textit{subj}), predicate-object (\textit{obj}), and modifier-noun/verb (\textit{mod}). Each relation is represented as a triplet: $T=(t_s,\textit{relation},t_o)$, $\textit{relation}\in\{\textit{subj, obj, mod}\}$. $t_s$ denotes the token in the head node, termed the \textit{head token}, where $s$ denotes its index in the input sequence. $t_o$ represents the token in the child node, termed the \textit{tail token}, where $o$ represents its index in the sequence.

To examine whether attention heads encode dependency relationships between tokens, we use the OV circuit to analyze the influence of attention heads on output logits of tail tokens when attending to head tokens.
Given an input sequence $[t_1, \ldots, t_n]$ and the triplet $T=(t_s,\textit{relation},t_o)$, we measure how much each head $h$ raises the tail token $t_o$ when attending to the head token $t_s$.

First, we look for attention heads that attend to the head token via the QK circuit. Given the current token $t_j$, let $A^h_j = \textit{softmax}(x_jW^h_{QK}\boldsymbol{x}^T)$ denote the attention probability of the $h$-th attention head over all tokens. If the head $h$ attends from $t_j$ to the head token $t_s$ with a high probability, \emph{i.e.,} $s=\arg\max_{1\le k\le j} A^h_{j,k}$ and $ A^h_{j,s}/ \max_{k\ne {s}} \{A^h_{j,k}\} > \tau$, we consider this head as a potential candidate for representing the triplet associated with the head token $t_s$. Otherwise, we skip this head on this triplet.
We set $\tau=2.2$ in experiments\footnote{Please refer to Appendix~\ref{sec: app_tau} for discussions about the setting of $\tau$.}.

Second, we examine whether these heads raise the tail token $t_o$ by computing the projection of the OV output on the vocabulary, $x_jW^h_{OV}W_u\in\mathbb{R}^{\vert \mathcal{V} \vert}$. Similar to \cite{bansal-etal-2023-rethinking}, we first compute the output probabilities of tokens as $\boldsymbol{p}^{h,j} = \textit{softmax}(x_jW^h_{OV}W_u)\in\mathbb{R}^{\vert \mathcal{V} \vert}$. Then, we extract the probability $p^{h,j}_{t_k}(k\le j)$ of each token $t_k$ before the token $t_j$ . 
Here we suppose all tokens $t_k$ before $t_j$ are unique for simplification, and if several positions share the same token (\emph{e.g.,} $t_{k_1}=t_{k_2}$), we only consider it once.
These probabilities are further transformed by subtracting their mean value and ruling out values smaller than zero, as follows.
\begin{equation}
    q^{h,j}_{t_k} = \max(0, p^{h,j}_{t_k} - \mathbb{E}_{1\le k'\le j} [p^{h,j}_{t_{k'}}])
\end{equation}
This transformation helps to focus on tokens whose output probabilities are raised. Then, we compute the following ratio $a^{h,j}_T$ to measure the significance of raising the tail token $t_o$ relative to all tokens before $t_j$. 
\begin{equation}
    a^{h,j}_T = q^{h,j}_{t_o} / {\sum}_{k=1}^j q^{h,j}_{t_k}
    \label{eq: relation score}
\end{equation}
Finally, for each head $h$, we average the relation index $a^{h,j}_T$ across all current tokens $t_j$ and across all triplets $T$. Note that we only consider current tokens after the head and tail tokens, \emph{i.e.,} $j\ge\max(s,o)$.

\paragraph{Model's ability in understanding dependencies.}
Before examining whether attention heads encode dependencies between tokens, we first test the model's overall proficiency in learning and understanding dependency relationships. We follow~\citet{clark-etal-2019-bert} to train an attention-and-words probing classifier, which takes the word embeddings and the attention weights extracted from InternLM2-1.8B as input and fits the probability of each token being the syntactic head of another token.
We train the classifier on 200 sequences from the AGENDA dataset, each with a length of less than 32. Then, we evaluate the accuracy of the predicted head positions on another 100 sequences. 
For 52\% tokens in the input sequence, the classifier can identify the position of their head tokens based on attention weights extracted from InternLM2-1.8B. This accuracy is significantly higher than the random guess, indicating that InternLM2-1.8B can well understand syntactic dependencies.
Therefore, we are motivated to further study dependencies in its attention heads.

\paragraph{The \textit{subj} and \textit{obj} dependencies are encoded in attention heads.}
Figure~\ref{fig: vis_of_dep_score} shows heatmaps of the relation index of each attention head \emph{w.r.t.} three types of dependency relationships.
As a baseline, we also compute the relation index when setting tail tokens in all triplets to the 10th token.
Such triplets do not represent any relationships, and the relation indexes of attention heads are lower than 0.1.
In comparison, for the \textit{subj} and \textit{obj} dependencies, the attention heads exhibit relation indexes close to 0.3, and the relation index \emph{w.r.t.} the \textit{mod} dependency is a bit lower.
This suggests the model may have better learned the \textit{subj} and \textit{obj} dependencies than \textit{mod}.
In Section~\ref{subsec: ICL}, it is also observed that the model exhibits better performance on the justification of the \textit{subj} and \textit{obj} dependencies than the \textit{mod} dependency, coinciding with this discovery.
Furthermore, relation indexes for the \textit{obj} dependency are more sparsely distributed than those for \textit{subj} dependency. There are about ten attention heads that exhibit salient values for the \textit{obj} dependency. 
This indicates that the model may store the \textit{obj} dependency using a few attention heads, while the \textit{subj} dependency is widely encoded in more attention heads.
These observations highlight the varying degrees of the model's understanding and representation of different dependency relationships.

\subsection{Semantic Relationship in Attention Heads}
\label{subsec: semantic_relationship}

\begin{figure}[t]
    \setlength{\abovecaptionskip}{0pt}
    \setlength{\belowcaptionskip}{-5pt}
    \centering    
    \includegraphics[width=\linewidth]{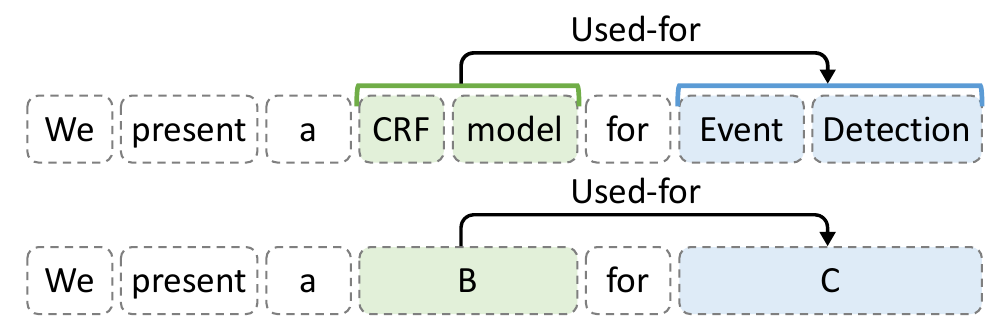}
    \caption{We replace the entities in the sentence (the first row) with capital English letters (the second row).}
    \label{fig: replacement}
\end{figure}

\begin{figure*}
    \setlength{\abovecaptionskip}{-2pt}
    \centering
    \includegraphics[width=0.245\linewidth]{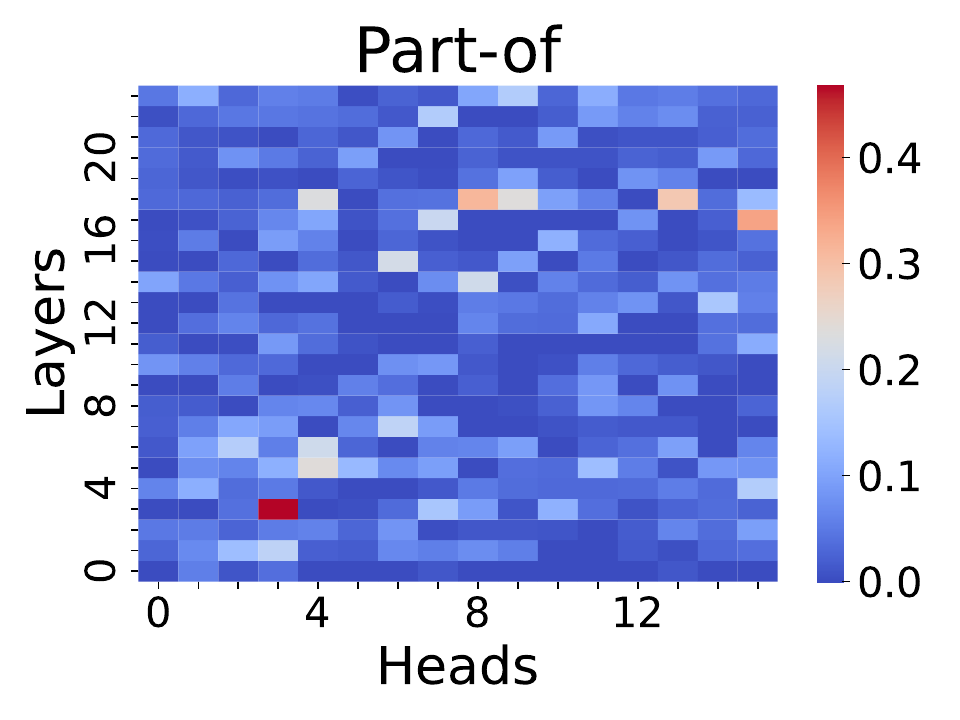}
    \includegraphics[width=0.245\linewidth]{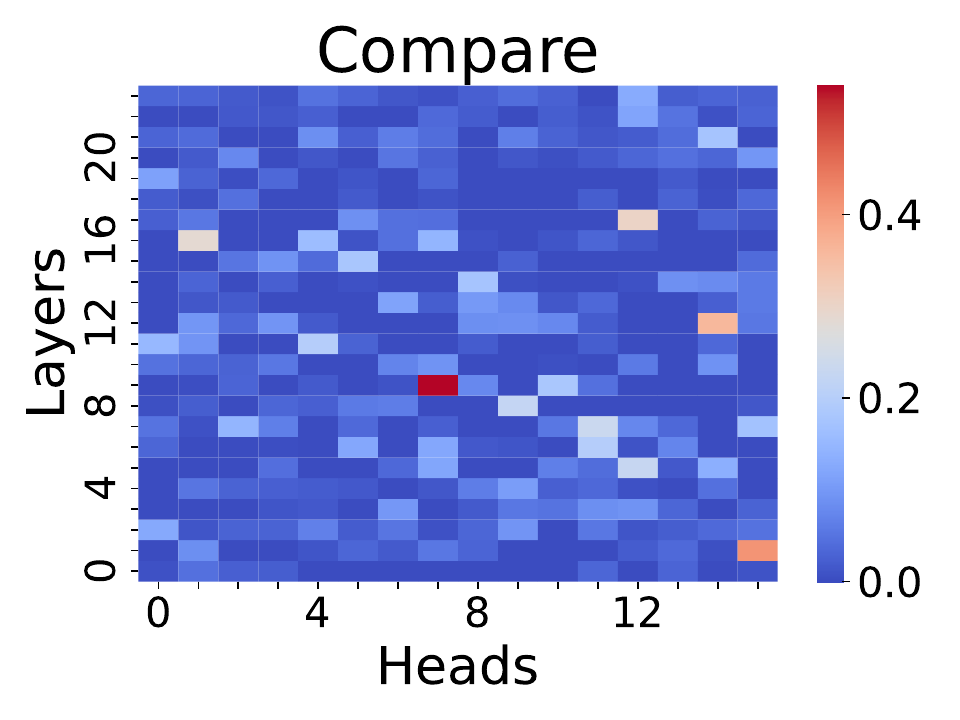}
    \includegraphics[width=0.245\linewidth]{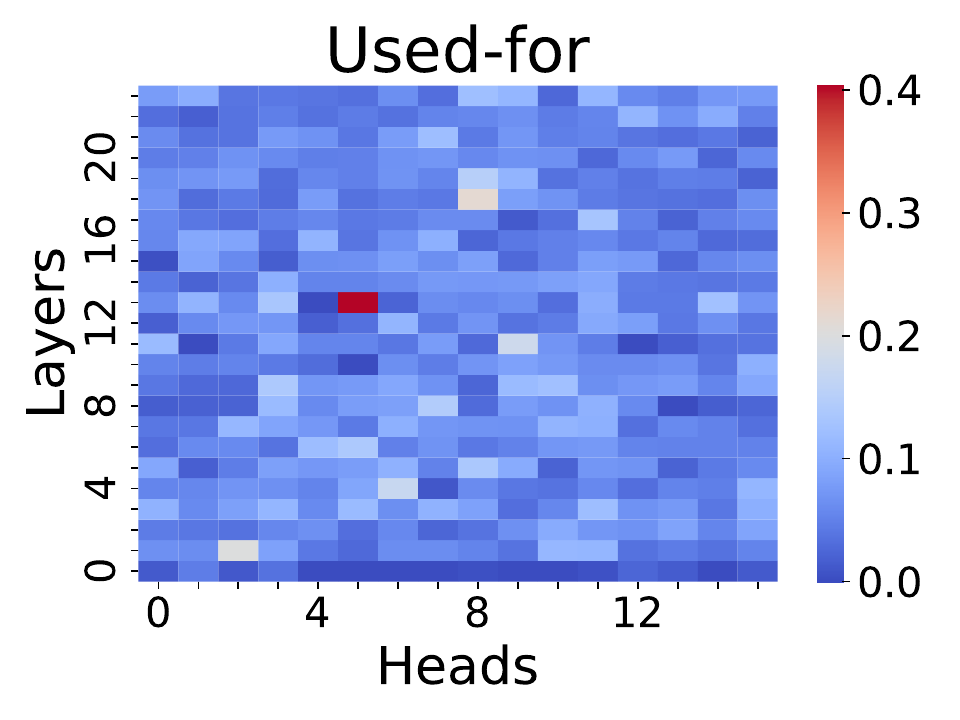}
    \includegraphics[width=0.245\linewidth]{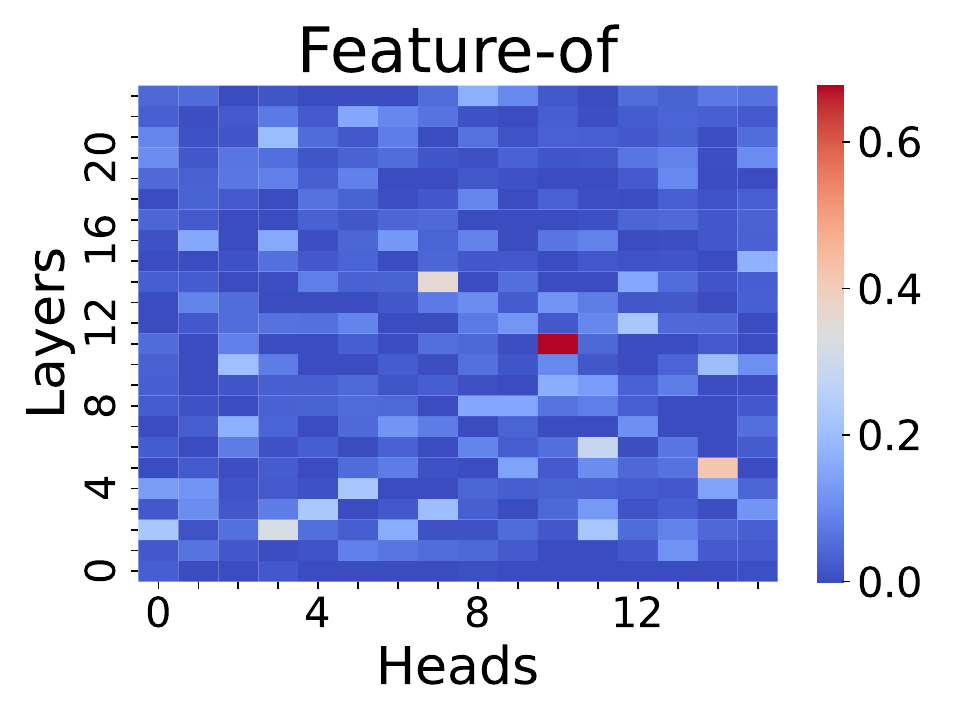}
    \includegraphics[width=0.245\linewidth]{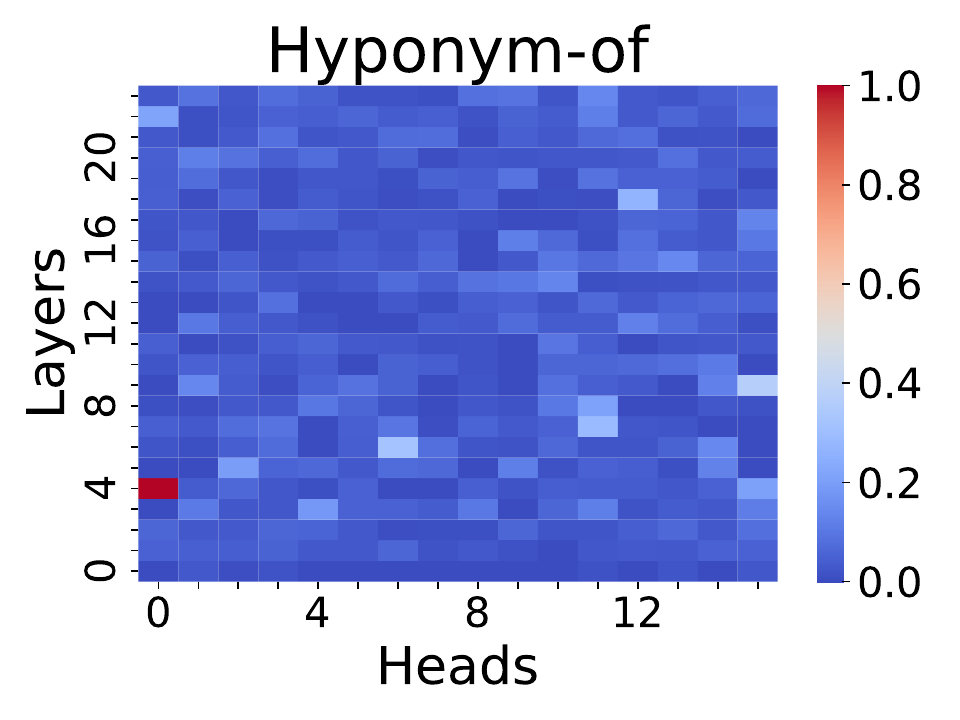}
    \includegraphics[width=0.245\linewidth]{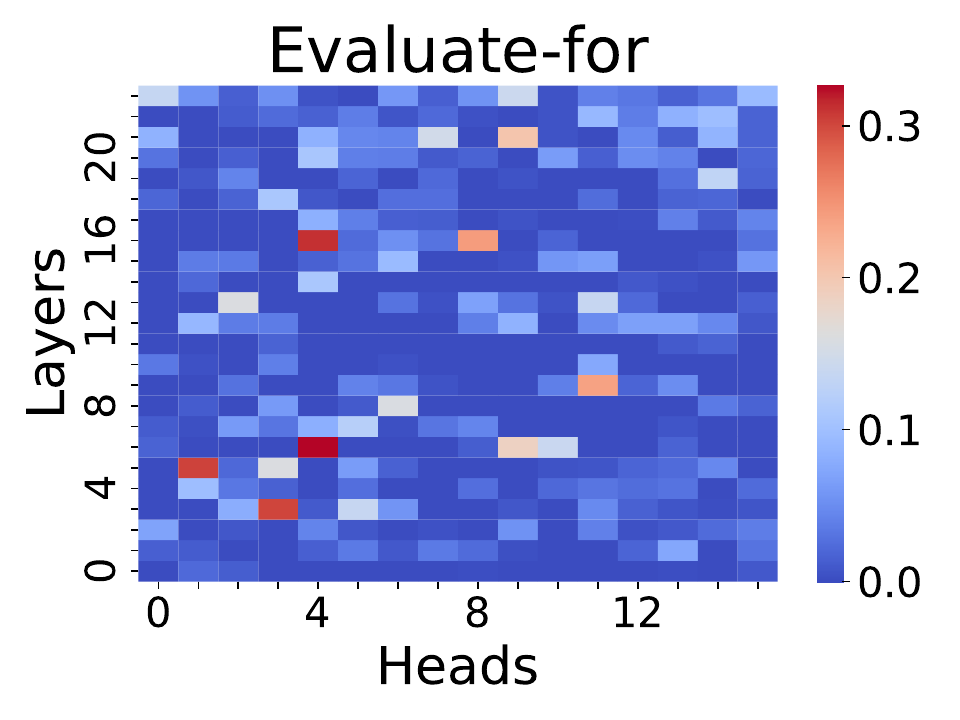}
    \includegraphics[width=0.245\linewidth]{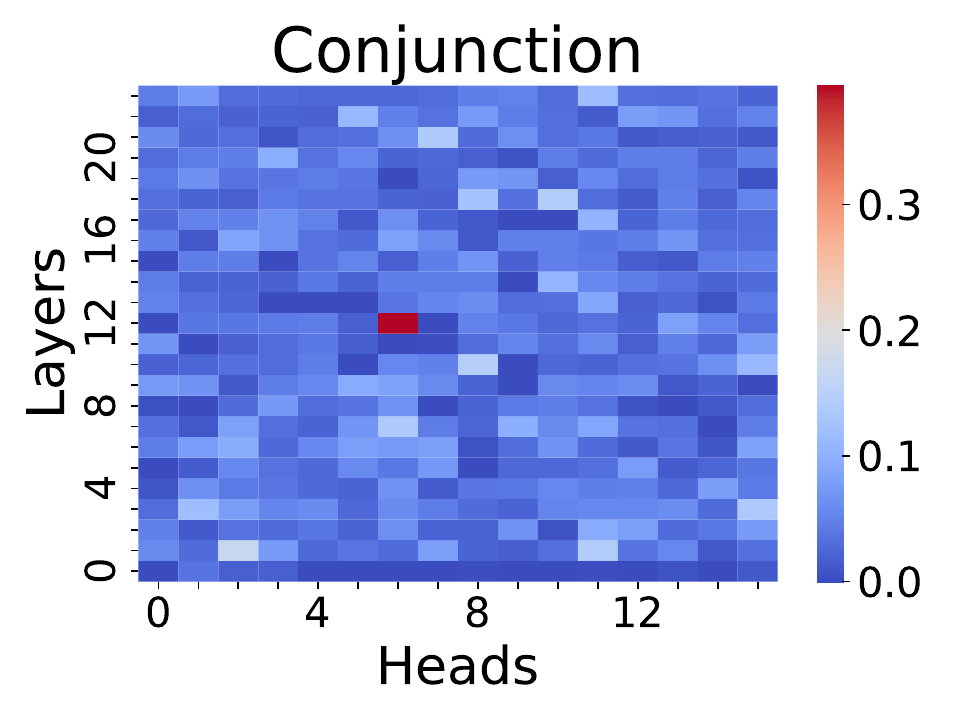}
    \caption{Heatmaps of the average relation index of attention heads for semantic relationships in knowledge graphs.}
    \label{fig: vis_of_relation_score}
\end{figure*}

In addition to syntactic dependencies, we also study semantic relationships between entities in knowledge graphs.  
There are seven types of relations between entities in the AGENDA dataset: Part-of, Compare, Used-for, Feature-of, Hyponym-of, Evaluate-for, and Conjunction.

Similar to dependencies in Section~\ref{subsec: syntactic_dependency}, we also expect to represent each relation between entities as a triplet $T=(t_s,\textit{relation},t_o)$.
Because the entities in sentences often consist of multiple tokens, we replace them with capital English letters\footnote{We exclude special letters like A,I,N,S,W, and E, which often appear alone and are meaningful alone.} as shown in Figure~\ref{fig: replacement}.
By representing each entity with a single letter, we can directly adopt the metric in Equation~\eqref{eq: relation score}. 
Besides, we remove the most frequent function words annotated by spaCy in the sentence. This step helps to reduce noise and focus on the more informative content words.

\paragraph{Various semantic relationships are encoded in attention heads.}
Figure~\ref{fig: vis_of_relation_score} illustrates relation indexes of attention heads \emph{w.r.t.} seven types of semantic relationships.
In contrast to syntactic dependencies, semantic relationships exhibit clearer patterns within attention heads.
Each type of relationship is represented by a range of 5 to 15 attention heads.
Interestingly, certain relationships, such as "Used-for," "Hyponym-of," and "Conjunction," appear to be more clustered in specific attention heads.
These findings suggest that the model possesses a capacity to represent various semantic relationships in attention heads.
Furthermore, considering these semantic relationships are bidirectional, we also analyze the reverse relation triplet ($\tilde{T}=(t_o, \textit{relation}, t_s)$) in attention heads in Appendix~\ref{sec: app_reverse_relation}.
Results in Figure~\ref{fig: vis_of_reverse_relation_score} show that some attention heads store both directions of the relationship, reflecting the model's ability to understand the reciprocal nature of these semantic relationships.

\section{In-Context Learning and Semantic Induction Heads}
\label{sec: correlation}
In this section, we investigate the correlation between in-context learning and semantic induction heads. We first categorize the ICL ability into three levels and observe the gradual emergence of different levels of ICL.
Then, we investigate the formation of semantic induction heads in the training process to understand the emergence of ICL.

\subsection{In-Context Learning of Different Levels}
\label{subsec: ICL}

\begin{table*}[t]
    \setlength{\abovecaptionskip}{3pt}
    \setlength{\belowcaptionskip}{-8pt}
    \centering
    \resizebox{\linewidth}{!}{
    \begin{tabular}{ c | c | c }
        \toprule
         Task & Entity set & Template  \\
         \midrule
         Binary classification & (fruit, month), (furniture, profession) & <E1, E2>: 0; <E2, E1>: 1 \\
         \midrule
         Four-class classification & (fruit, month) & <E1, E1>: 0; <E1, E2>: 1; <E2, E1>: 2; <E2, E2>: 3 \\
         \midrule
         Nine-class classification & (fruit, animal, month) & <E1, E1>: 0; <E1, E2>: 1; <E1, E3>: 2; <E2, E1>: 3; $\cdots$ \\
         \midrule
         Relation justification & (\textit{subj}, \textit{verb}), (\textit{verb}, \textit{obj}), (\textit{mod}, \textit{obj}), (part, whole) &  <E1, E2>: true ; <animal, month>: false \\
         \bottomrule
         \end{tabular}
    }
    \caption{We construct toy tasks for evaluating the ICL ability of models. E1, E2, and E3 in the template refer to instances belonging to the 1st, 2nd, and 3rd categories in each pair of entity sets. For example, binary classification contains inputs like ``apple, January: 0'' and ``April, orange:1''.}
    \label{tab: ICL_tasks}
\end{table*}

In-context learning refers to the ability to learn from the context to perform an unseen task. However, there is no standard measurement for the ICL ability of LLMs. \citet{kaplan2020scaling,olsson2022context} consider ICL as the ability to better predict later tokens in the context than earlier tokens.
Another more widely adopted definition of ICL follows a few-shot setting. In this setting, language models are provided with a few examples within the context of the prompt, and the model can better perform the task with more examples given. This definition emphasizes the model's ability to extract and generalize the information in the context. 


We rethink the ICL ability from the perspective of what the model has learned from the context.
The loss reduction considered in~\citep{olsson2022context} only demonstrates that the context does help models make predictions, but it is unclear what the model has learned. 
\citet{Chomsky1957} has proposed that when humans learn new languages, they initially grasp the surface structure of the language before delving into the deep structure. Inspired by this, we hypothesize that LLMs also first learn the surface format of the context, and then gradually comprehend the deep patterns or rules within the context.
Based on this hypothesis, we categorize the ICL ability into three levels: 

\textbullet~ Loss reduction: This level of ICL is characterized by a reduction in the loss of tokens as the model predicts later tokens in the context. \citet{olsson2022context} demonstrates ICL at this initial level. 

\textbullet~ Format compliance (few-shot): At this level, the model learns the format of examples in the prompt (\emph{e.g.,} numbers and symbols), and generates outputs following the same format. Although the outputs have the correct format, the predictions may be incorrect.

\textbullet~  Pattern discovery (few-shot): This level expects the model to recognize and comprehend the underlying pattern within the examples, and apply it consistently to generate the correct prediction.

By categorizing ICL into these levels, we can systematically assess the progression and development of the model's ICL abilities. 

\begin{figure}
    \setlength{\abovecaptionskip}{-1pt}
    \setlength{\belowcaptionskip}{-5pt}
    \centering  
    \includegraphics[width=\linewidth]{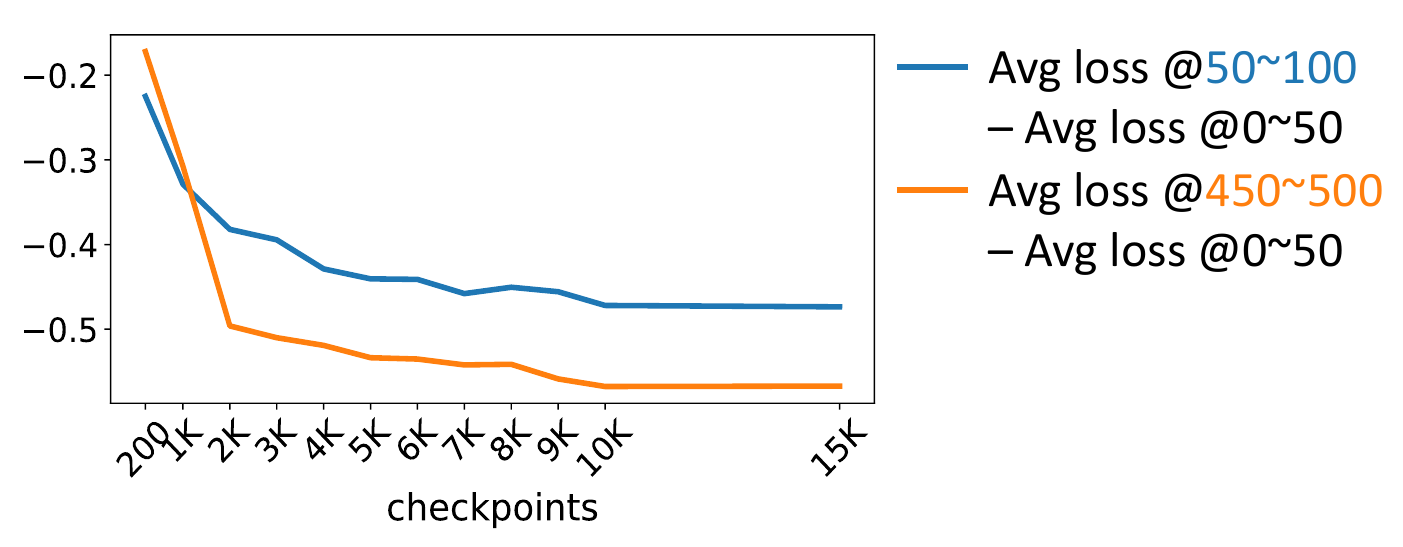}
    \caption{The average difference in loss between later tokens and early tokens decreases from the very beginning of the training.}
    \label{fig: loss_reduction}
\end{figure}

\textbf{Model.}
To study the ICL ability of the model during the training process, we train a model from scratch using the InternLM framework~\citep{2023internlm}. The model contains 20 transformer layers, and each layer consists of $H\!=\!16$ attention heads. The hidden size of the model is $d=2048$, thus each head has a dimension of $d_h=128$. 
The model is trained using the SlimPajama dataset~\citep{cerebras2023slimpajama} on 32 GPUs, and the batch size on each GPU is 128K tokens.
We train the model for 40k steps, with checkpoints saved every 200 steps to monitor the model's progress in relationship representing and ICL during training.

\begin{figure*}
    \setlength{\abovecaptionskip}{0pt}
    \setlength{\belowcaptionskip}{-5pt}
    \centering
    \includegraphics[width=0.245\linewidth]{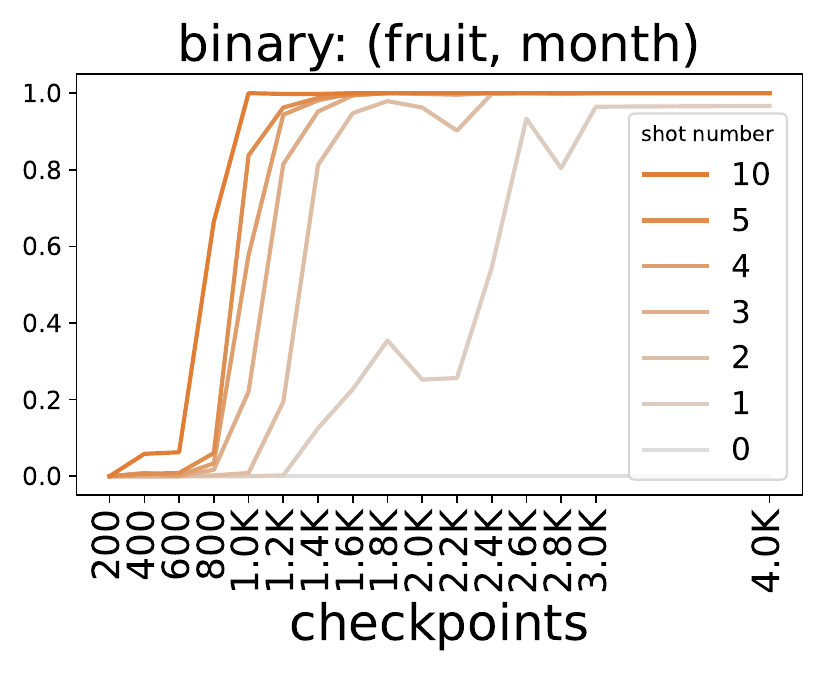}
    \hfill
    \includegraphics[width=0.245\linewidth]{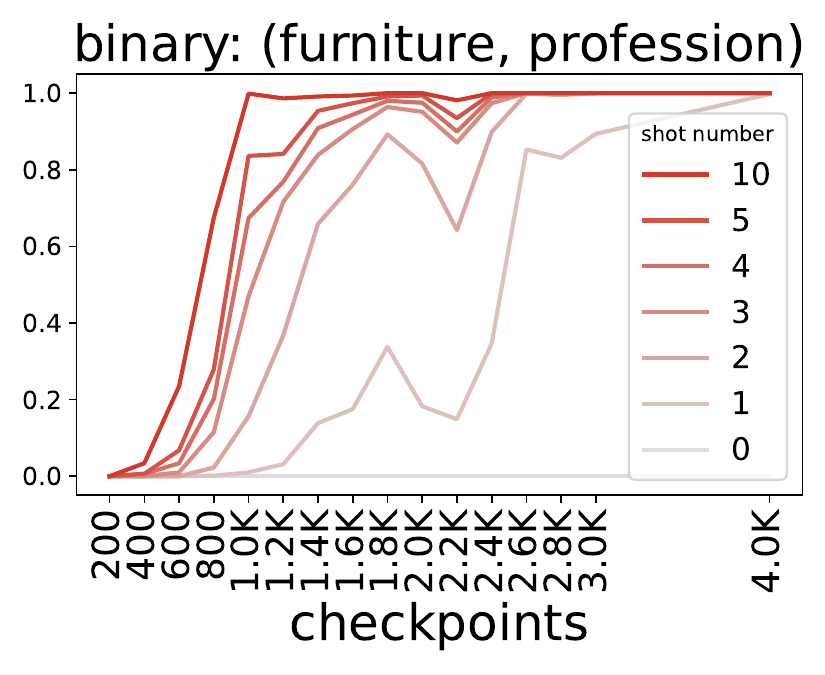}
    \hfill
    \includegraphics[width=0.245\linewidth]{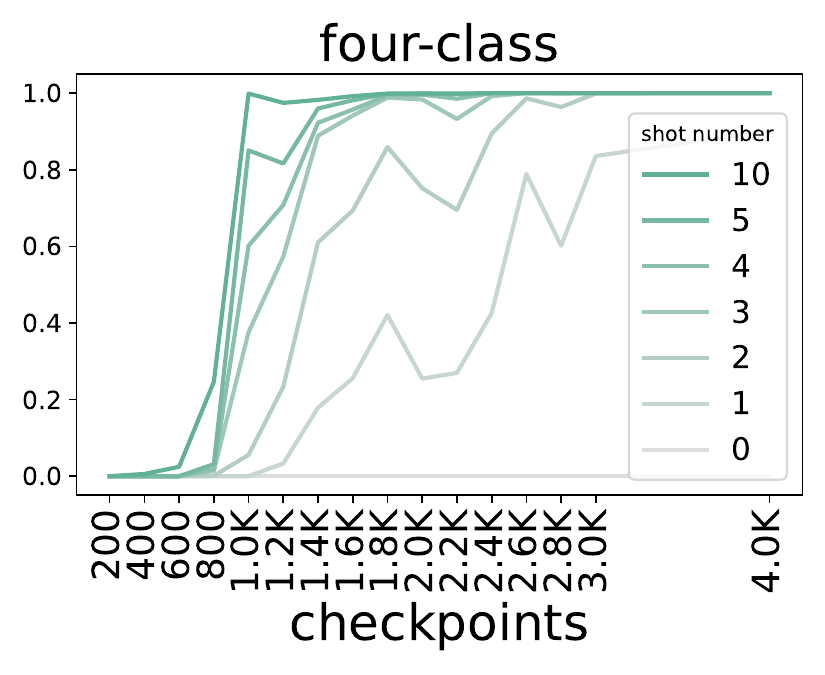}
    \hfill
    \includegraphics[width=0.245\linewidth]{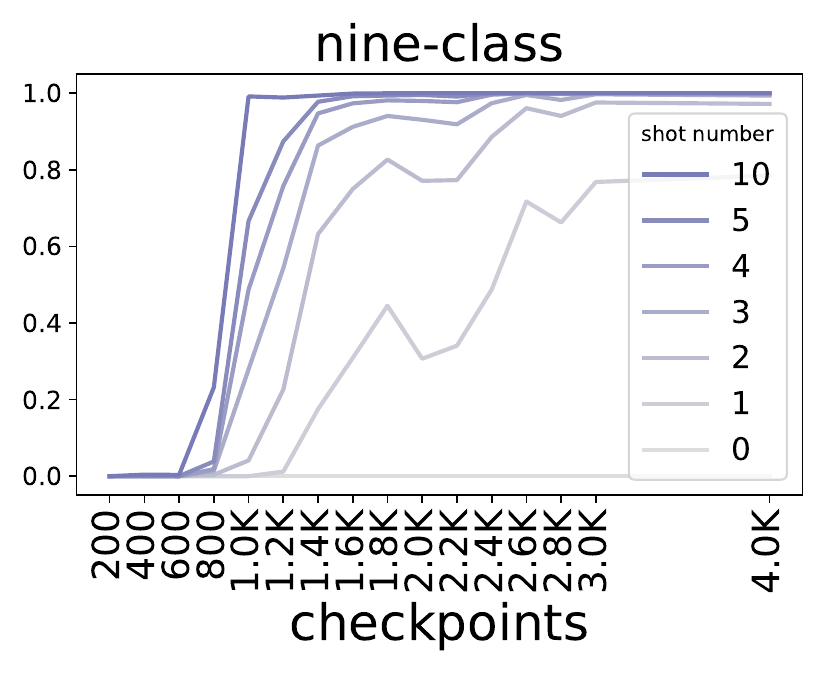}
    \caption{Format accuracy of different tasks at different checkpoints. Each line represents the format accuracy with different numbers of shots (examples) in the prompt.}
    \label{fig: format_acc}
\end{figure*}

\begin{figure*}
    \setlength{\abovecaptionskip}{0pt}
    \setlength{\belowcaptionskip}{-5pt}
    \centering
    \includegraphics[width=0.245\linewidth]{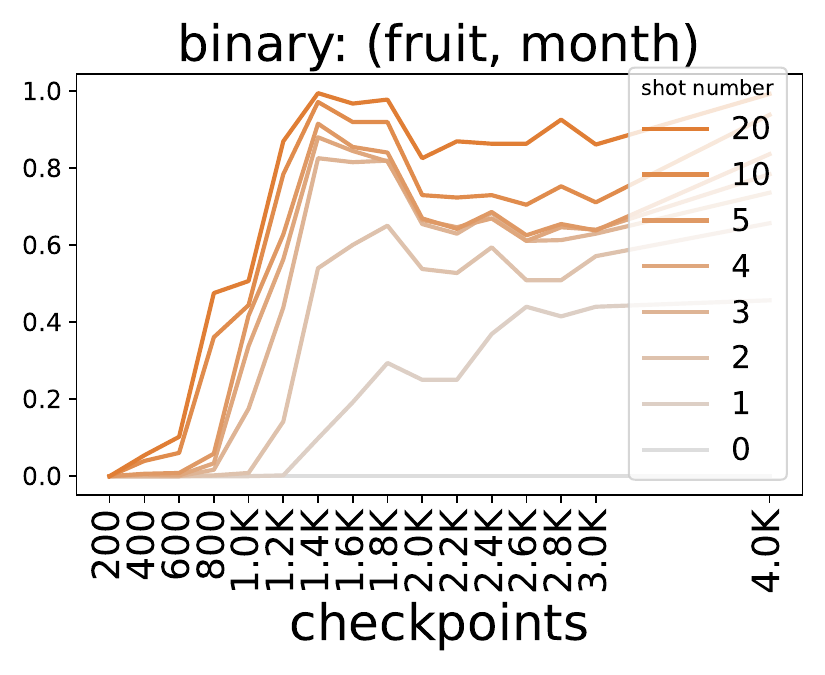}
    \hfill
    \includegraphics[width=0.245\linewidth]{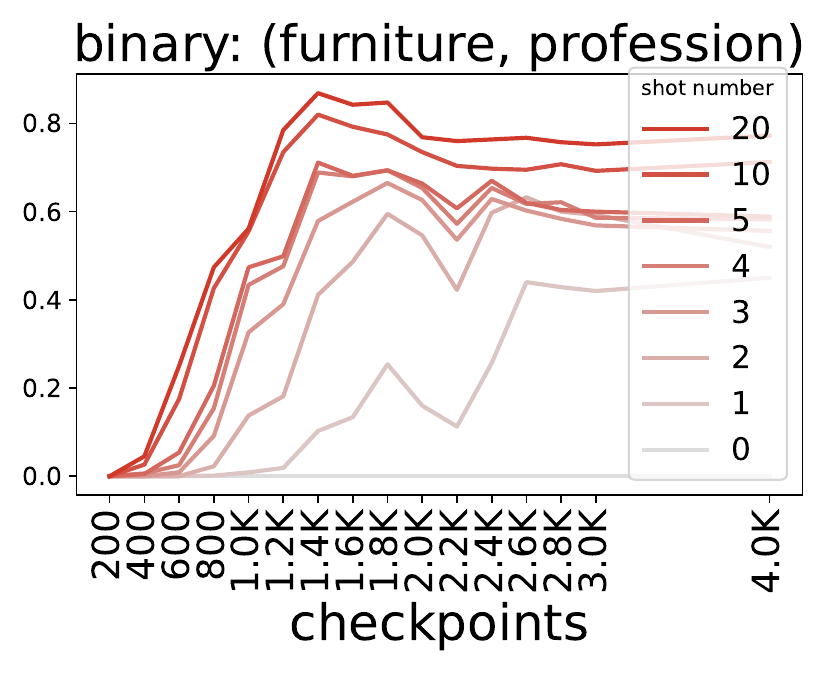}
    \hfill
    \includegraphics[width=0.245\linewidth]{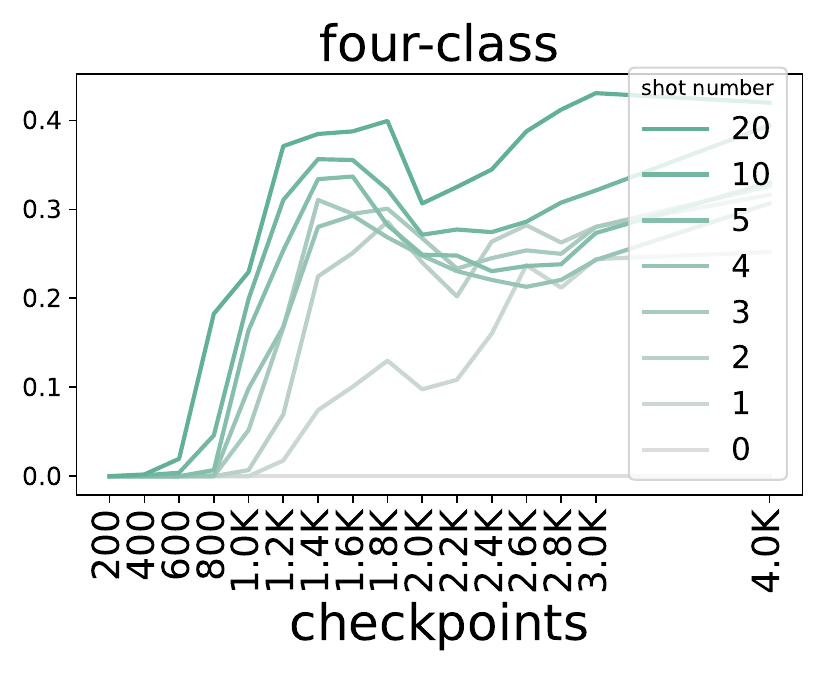}
    \hfill
    \includegraphics[width=0.245\linewidth]{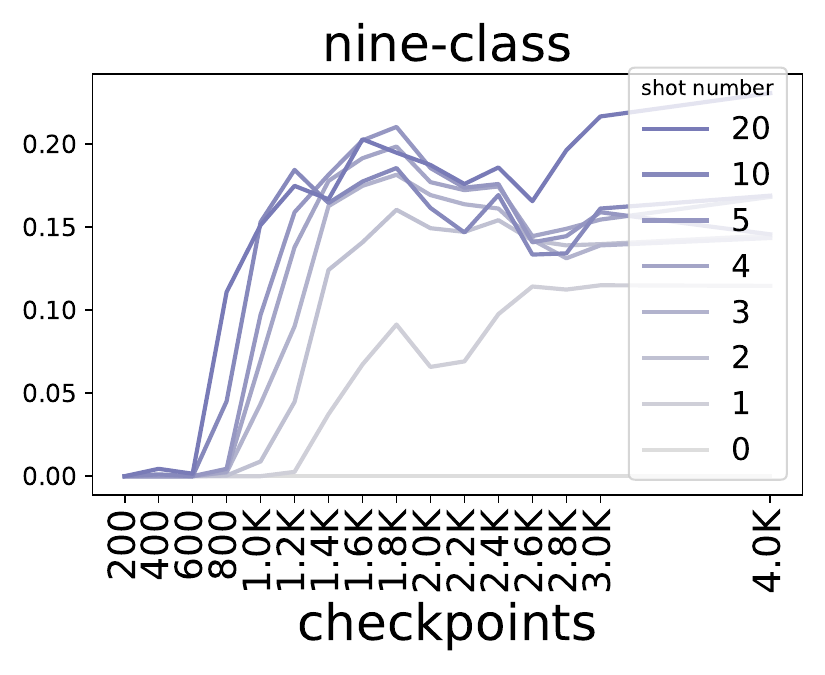}
    \includegraphics[width=0.245\linewidth]{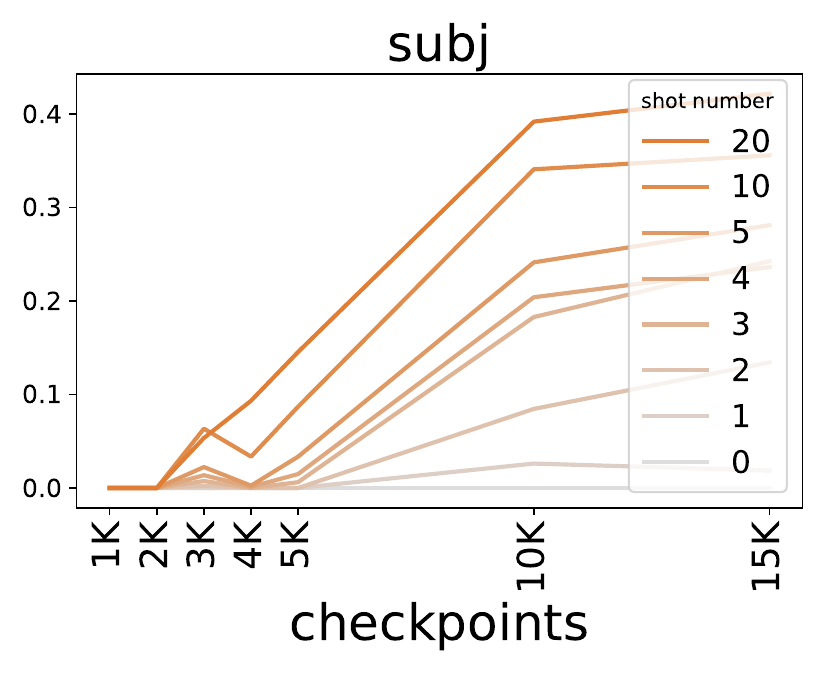}
    \hfill
    \includegraphics[width=0.245\linewidth]{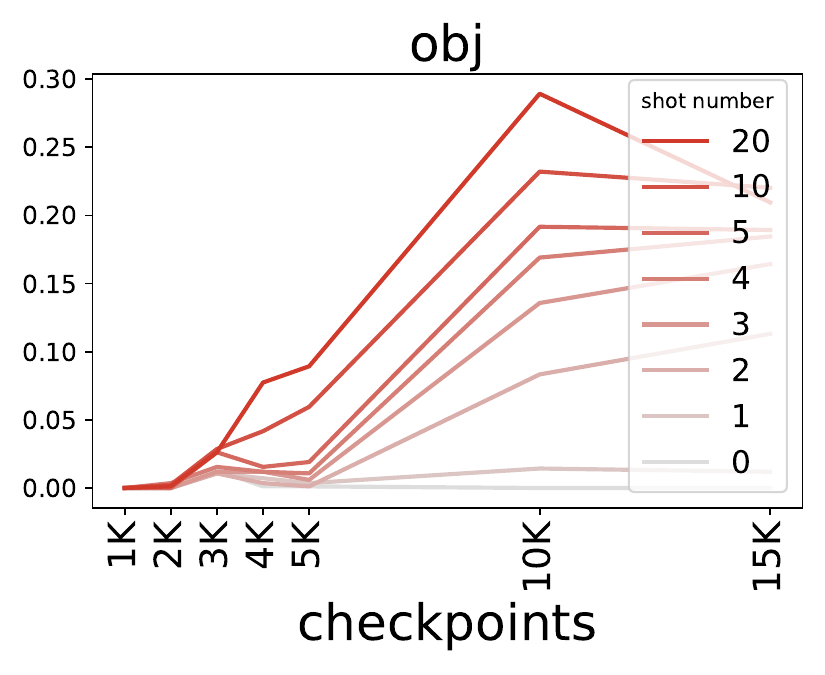}
    \hfill
    \includegraphics[width=0.245\linewidth]{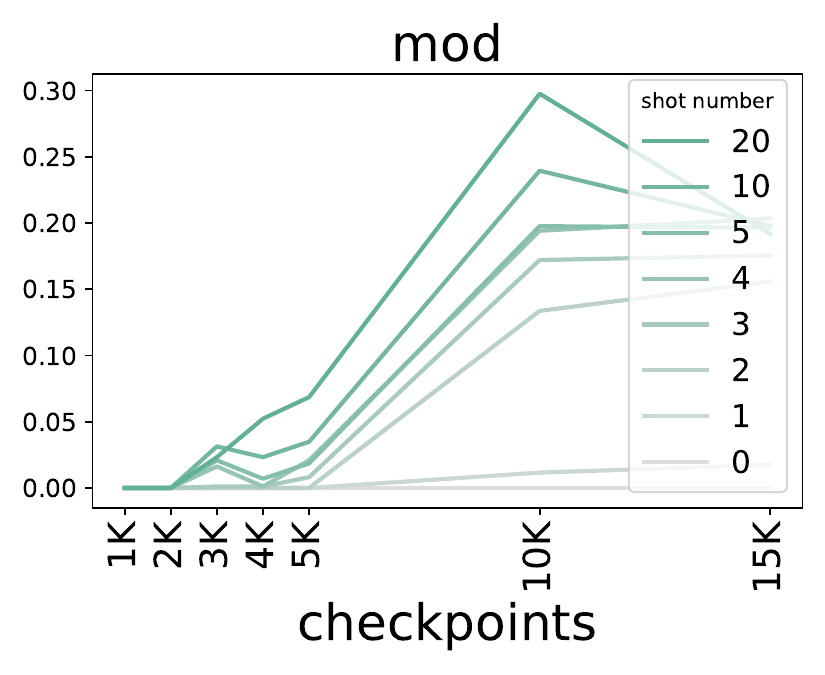}
    \hfill
    \includegraphics[width=0.245\linewidth]{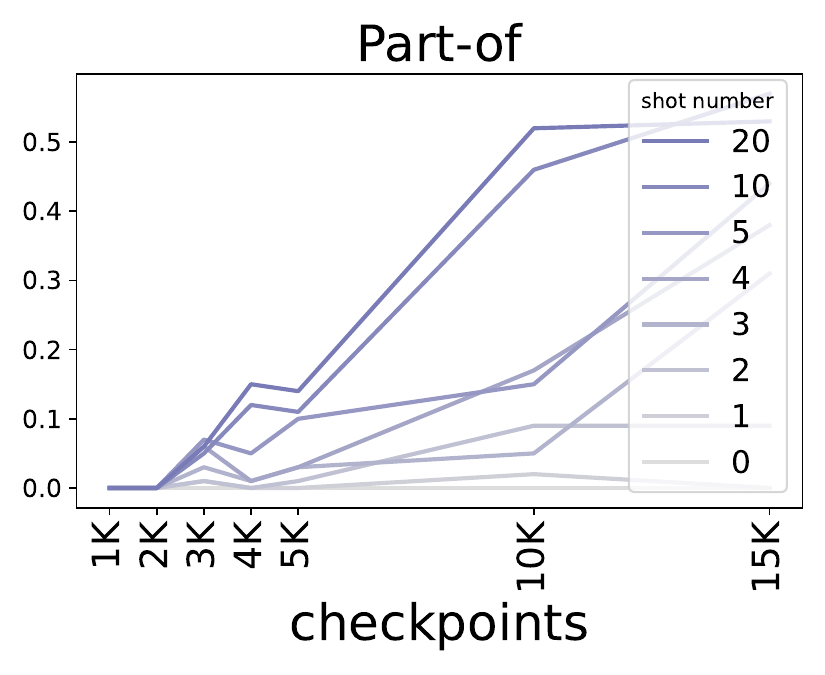}
    \caption{Prediction accuracy of different tasks at different checkpoints. Each line represents the prediction accuracy with different numbers of shots (examples) in the prompt. 
    It is worth noting that these tasks cannot be simply considered as binary classification tasks, which is discussed in Appendix~\ref{sec: app_pattern_tasks}.}
    \label{fig: pattern_acc}
\end{figure*}

\textbf{Measurements and results.}
We assess the ICL ability at each level using the following methods. 

\textit{For loss reduction},
we follow~\citet{olsson2022context} but with a minor modification that improves the statistical stability. Specifically, we adjust the formula from the loss at the $j$-th token minus the loss at the $i$-th token ($i\le j$), to the averaged loss over the interval from the $j\sim (i+j)$-th tokens minus the averaged loss over the $0\sim i$-th tokens.

We sample 100 sentences from the SlimPajama dataset and we set $(i,j)\in\{(50,50), (50,450)\}$ to measure the loss reduction at different training checkpoints. Figure~\ref{fig: loss_reduction} shows that the loss difference between later and early tokens decreases quickly from the very beginning of the training.
This suggests that from the beginning of the training, the model progressively improves its ability to leverage longer contexts for better predictions.

\textit{For format compliance}, we construct classification tasks in Table~\ref{tab: ICL_tasks}. 
We adopt the few-shot setting of ICL, and the prompt is designed to include several examples followed by a query.
Here we ensure that when the number of examples exceeds the number of classes, there is at least one example of each class in the prompt. To test the format compliance ability of the model, we force the model to generate only one token. If the generated token is also a number, matching the format presented in the examples, we consider it to have a correct format. We compute the accuracy of the format to measure the format compliance ability of models.

Figure~\ref{fig: format_acc} reports the format accuracy given different numbers of shots at different training checkpoints.
For two binary classification tasks, we observe that the model's format accuracy progressively improves as the number of shots increases, starting from the 400th step.
This suggests that the model's format compliance ability emerges at the early training stage, and it is independent of the entities involved in the task.
For the four-class and nine-class classifications, the model gains improvement with an increasing number of shots at later stages of training (the 600th step and the 800th step, respectively).
This indicates that the format compliance to more difficult tasks tends to appear at later training stages.
Despite that, the model consistently achieves around 100\% format accuracy when using 20 shots at the 1k-th step, and achieves a high accuracy with only one shot after 3k steps.
We also measure the format compliance ability from the perspective of the minimum number of shots required to achieve a format accuracy of over 80\%. Please refer to Appendix~\ref{sec: app_format_shot_num} for details.

\begin{figure*}
    \setlength{\abovecaptionskip}{-1pt}
    \setlength{\belowcaptionskip}{-8pt}
    \centering
    \begin{minipage}{\linewidth}
        \centering
        \includegraphics[width=0.32\linewidth]{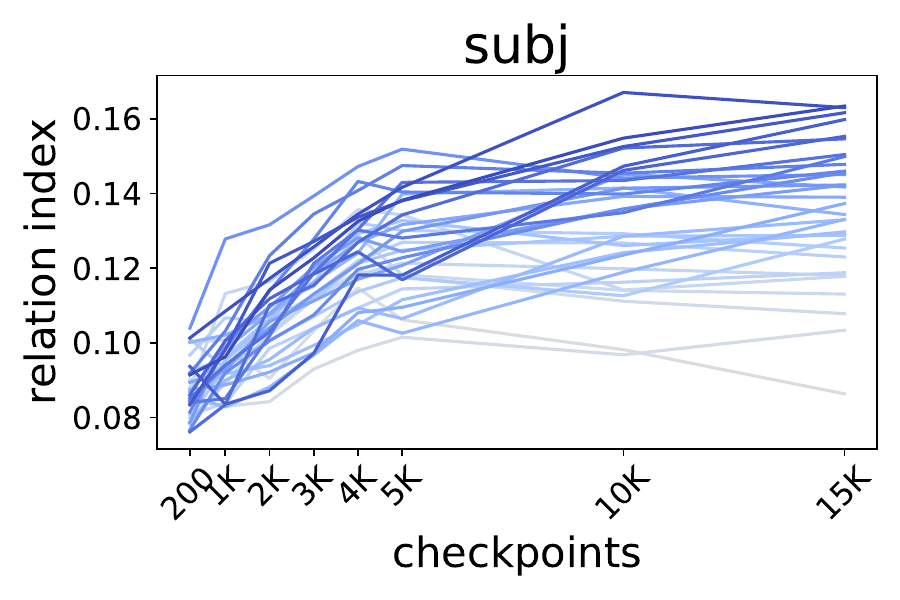}
        \hfill
        \includegraphics[width=0.32\linewidth]{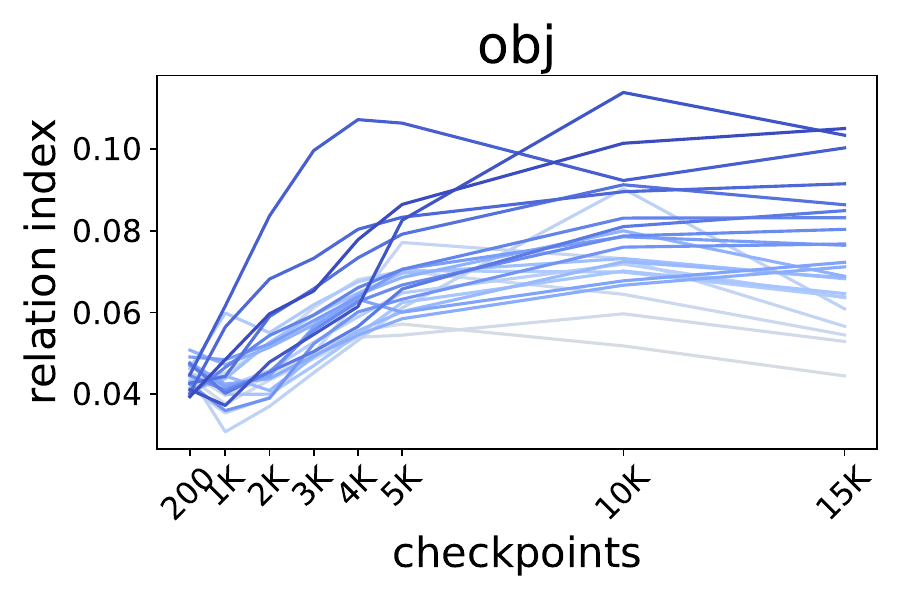}
        \hfill
        \includegraphics[width=0.32\linewidth]{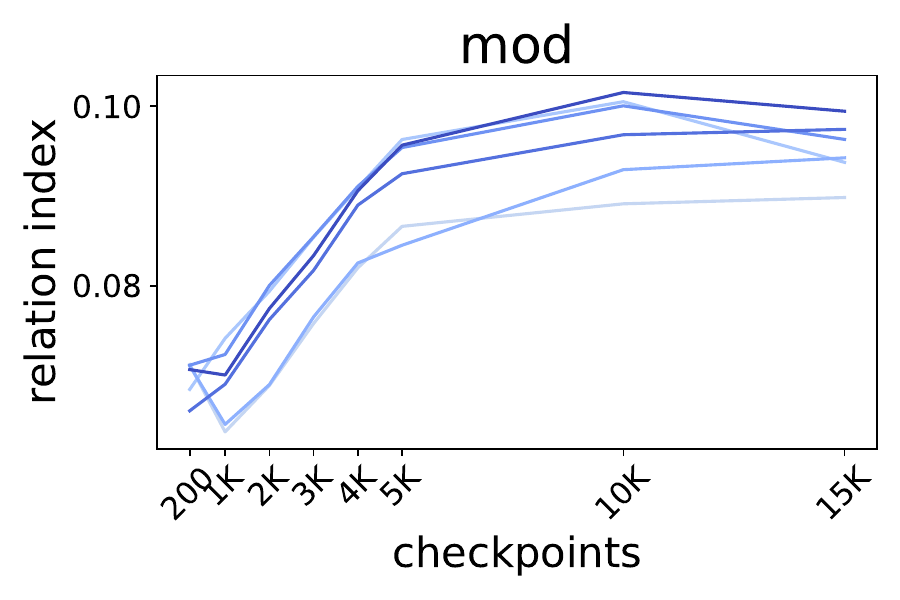}
    \end{minipage}
    \caption{The change curve of average relation indexes of attention heads for syntactic dependency. Each line in the figure represents the relation index of an attention head over training time, and lines are colored according to the value at the 15k-th step.}
    \label{fig: dep_curve}
\end{figure*}

\begin{figure}
    \setlength{\abovecaptionskip}{-1pt}
    \setlength{\belowcaptionskip}{-8pt}
    \centering
    \includegraphics[width=0.75\linewidth]{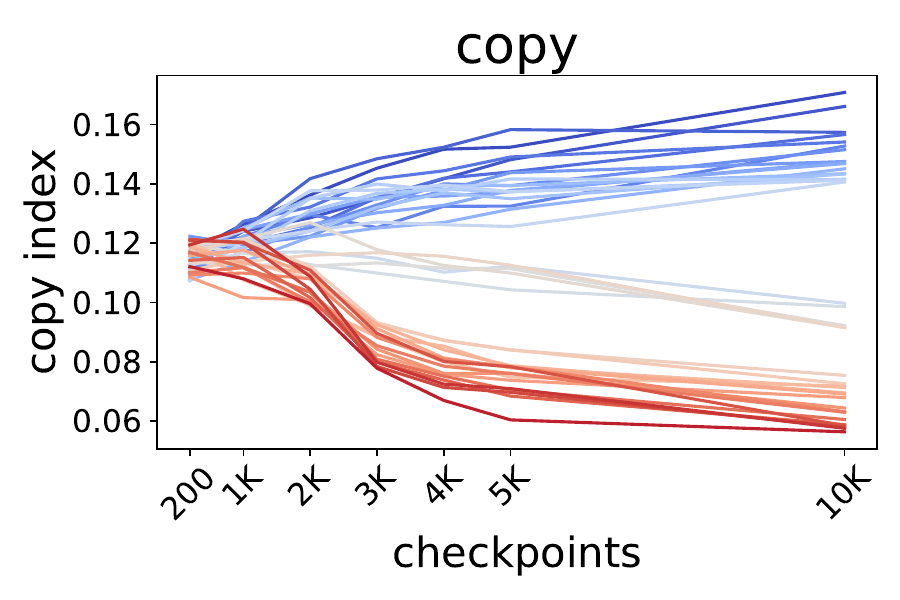}
    \caption{The change curve of copying scores of attention heads. Each line in the figure represents the copying score of an attention head over training time, and lines are colored according to the value at the 10k-th step.}
    \label{fig: copy_curve}
\end{figure}

\textit{For pattern discovery}, we still use the above classification tasks, but the difference is that we compute the accuracy of the predicted label. If the model can generate correct labels, we consider it to have successfully discovered and applied the underlying pattern in the prompt.
Besides, we also construct four relation justification tasks in Table~\ref{tab: ICL_tasks}, which are related to the syntactic dependency and semantic relationship studied in this paper.

Figure~\ref{fig: pattern_acc} reports the prediction accuracy of the model at different checkpoints.
For classification tasks in the first row, the model achieves considerable accuracy with 20 shots at around 1400 steps, indicating that the pattern discovery ability is mastered later than the format compliance.
Furthermore, simple binary classification tasks are learned earlier than complex four-class and nine-class classification tasks.
The relation justification tasks in the second row, which are more difficult than classification tasks, are learned at later stages, typically starting from around the 2k-th step.
After approximately 10k steps, the prediction accuracy tends to saturate.
Figure~\ref{fig: luyou_icl_class_acc} in Appendix~\ref{sec: app_luyou_icl} shows that till the end of the training, even with 100 examples, the model cannot fully learn the pattern in the prompt.

\textbf{Progressive learning of ICL of different levels.}
From the above results, we can observe a progressive learning process for the different levels of ICL.
The loss reduction happens from the beginning of the training, followed by the emergence of format compliance (after 400 steps), and pattern discovery is mastered in the last (after 1k or 2k steps).
This discovery aligns with our hypothesis that three levels of ICL have increasing difficulties.
Moreover, within the format compliance and pattern discovery, we observe that the model typically learns more challenging tasks at later training stages.

\subsection{Correlation Between Semantic Induction Heads and ICL}
\label{subsec: correlation}

In this section, we investigate the formation of semantic induction heads during the training process and discover their correlation with ICL.
We compute the average relation index of attention heads over all triplets for each syntactic dependency and each semantic relationship in different checkpoints.
Here we only ensure $s=\arg\max_{1\le k\le N} A^h_{j,k}$, and do not require $A^h_{j,s}/ \max_{k\ne {s}} \{A^h_{j,k}\} > \tau$ any more, because in the early stage of the training, it is too challenging to find attention heads having an extremely high attention probability on head tokens.

We find that the relation index of some attention heads increases during the same stage as the emergence of the ICL ability.
Specifically, Figure~\ref{fig: dep_curve} shows the change in the relation index for syntactic dependencies of attention heads. We sampled attention heads with an increasing relation index for visualization.
It can be observed that relation indexes of some attention heads begin increasing from the beginning of the training, aligning with the emergence of loss reduction.
On the other hand, relation indexes of other attention heads begin to increase after around 1k or 2k steps, which coincides with the emergence of the pattern discovery ability.
Thus, we infer that the formation of semantic induction heads plays a crucial role in the development of the ICL ability. These semantic induction heads likely contribute to capturing and representing relationships between tokens, which are essential for the ICL ability.

\begin{figure*}[t]
    \begin{minipage}{0.27\linewidth}
        \caption{Occurrence of each attention head having the largest value of $\mathbb{E}_j[a^{h,j}_T]$ with $\mathbb{E}_j[a^{h,j}_T]>0$ over all triplets $T$ of each type of dependency.}
    \label{fig: vis_of_dep_group}
    \end{minipage}
    \hfill
    \begin{minipage}{0.71\linewidth}
    \centering
    \includegraphics[width=0.32\linewidth]{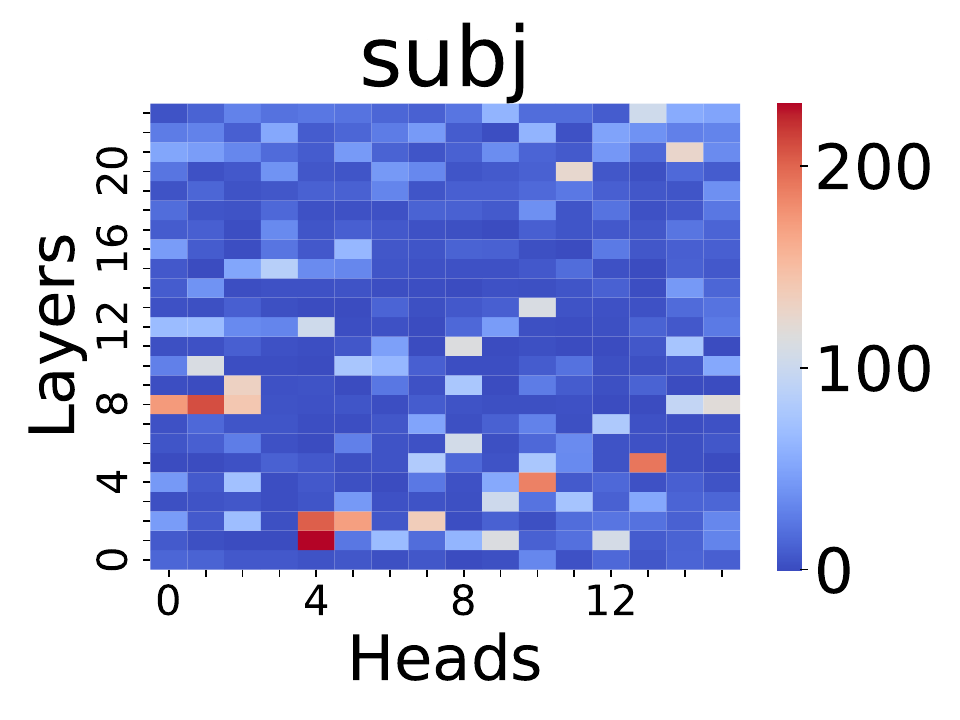}
    \includegraphics[width=0.32\linewidth]{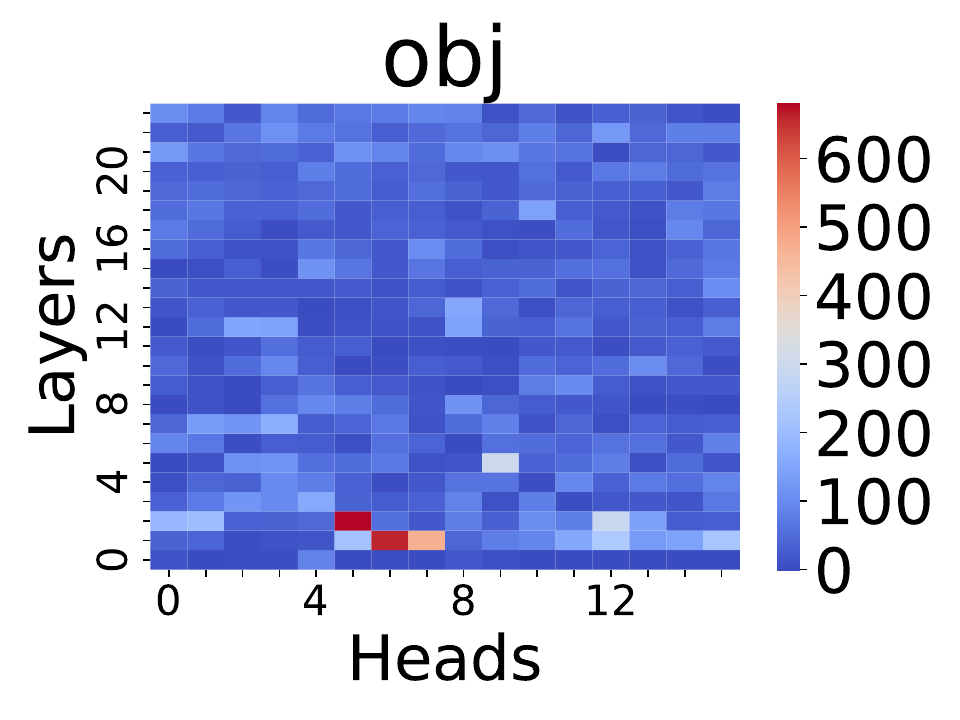}
    \includegraphics[width=0.32\linewidth]{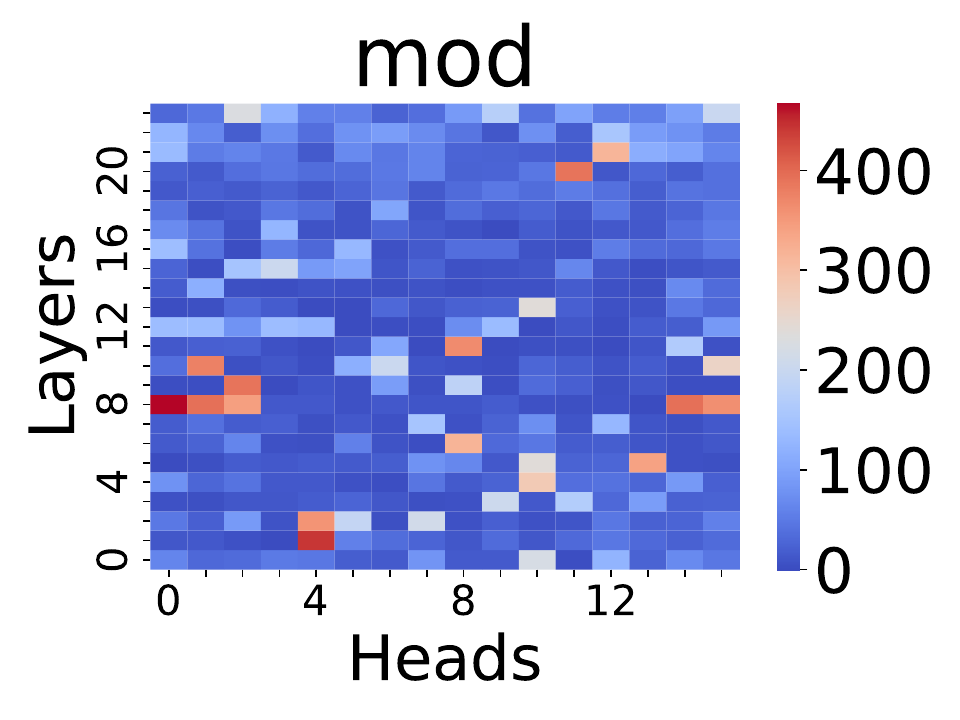}
    \end{minipage}
    \vspace{-10pt}
\end{figure*}

On the other hand, Figure~\ref{fig: copy_curve} shows the change in copying scores~\citep{olsson2022context, bansal-etal-2023-rethinking} of attention heads.
Copying scores of some attention heads start increasing from 200 steps, so we infer the copying mechanism is responsible for the loss reduction and the format compliance. It is reasonable because the task used for evaluating the format compliance can be simply achieved by copying the token (``0'' or ``1'') after the colon in the preceding context to the output.
More interestingly, copying scores of other heads begin to drop from the 1K step, where the pattern discovery ability emerges. Therefore, we hypothesize that the copying behavior is not always good for ICL, because sometimes direct copying may cause incorrect predictions.

\section{More Discussions about Relationships Encoded in Attention Heads}
\label{sec: grouping}
A notable distinction between copying and relationship is that the relationship between tokens depends on the input context, while copying is context-agnostic. Therefore, different inputs may utilize different attention heads. Thus, we propose to find a common group of attention heads that represent specific relationships in different inputs. For each triplet $T$, we identify the attention head that has the largest value of $\mathbb{E}_j[a^{h,j}_T]$  (larger than 0). Then, we count the occurrence of each head having the largest value among all triplets $T$ for each dependency relationship.

Figure~\ref{fig: vis_of_dep_group} shows the number of occurrences of each attention head having the largest relation index for syntactic dependencies. Please refer to Appendix~\ref{sec: app_relation_group} for results on semantic relationships. We find that for each type of dependency/relationship, there are around 5$\sim$15 attention heads frequently activated by different inputs.
Besides, different relationships tend to share some common attention heads (\emph{e.g.,} layer2, head4 for syntactic dependencies). This may indicate that these human-defined relationships are not mutually exclusive from the LLMs' point of view. In other words, there exists a many-to-many mapping between human-defined relationships and attention heads in LLMs.

\section{Conclusion}
Previous studies~\citep{elhage2021mathematical,olsson2022context} in mechanistic interpretability only studied the simple functions in very specific tasks~\citep{wang2023interpretability,lieberum2023does}.
In this study, we extended the conventional induction heads to analyze high-level relationships between words/entities in natural languages.
Our experiments revealed that specific attention heads encode syntactic dependencies and semantic relationships in natural languages.
Furthermore, we identified three levels of the in-context learning ability of LLMs, and experimental results showed they are progressively learned during the training process.
Finally, we observed a close correlation between the formation of semantic induction heads and in-context learning ability, strengthening our understanding of in-context learning.

\section*{Limitations}
Limitations of this paper lie in the following three perspectives.  (1) While the proposed relation index has the potential to be adapted to different relationships in various languages, this paper only focuses on syntactic dependency and semantic relations in English. We think it is a promising direction to examine the representation of relationships in different languages and leave it to future work.
(2) Although we have extended the simple copying operation to complex semantic relationships, the proposed method is limited to relationships between two tokens/entities. 
(3) Due to limitations in computational resources, we only conduct experiments on $\sim$1B models.

\section*{Acknowledgement}
This work is supported by Shanghai Artificial Intelligence Laboratory. 

\bibliography{custom,anthology}

\appendix
\include{appendix}

\end{document}

%% file: appendix.tex
\section{Setting of $\tau$ in the relation index.}
\label{sec: app_tau}
\begin{figure}[h]
    \centering
    \includegraphics[width=0.85\linewidth]{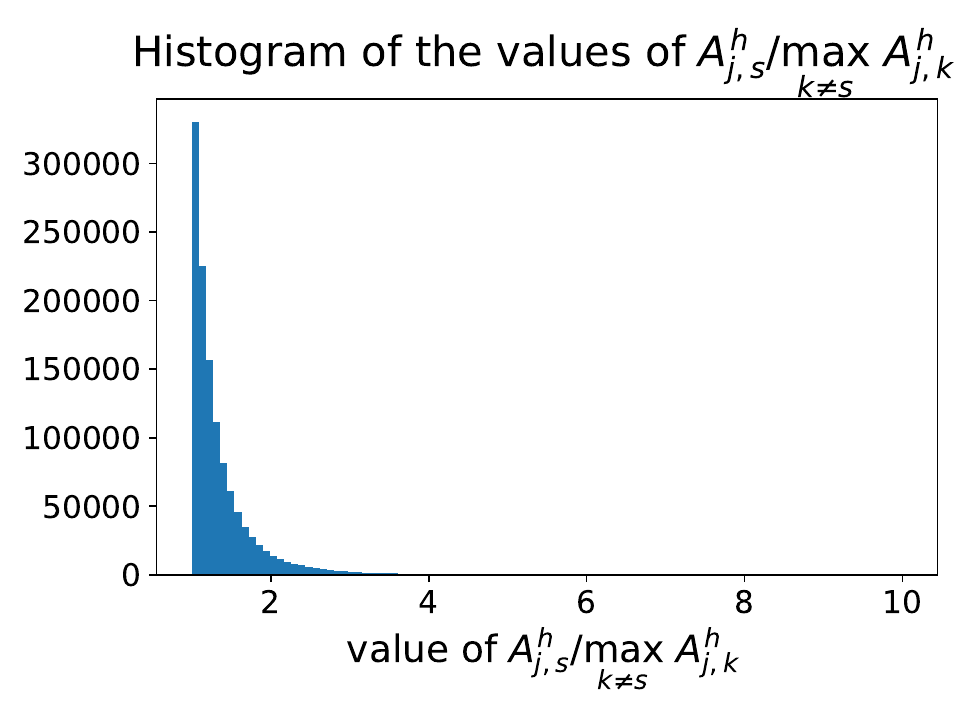}
    \vspace{-10pt}
    \caption{Distribution of $\frac{A^h_{j,s}}{\max_{k\ne s}\{A^h_{j,k}\}}$ in InternLM2-1.8B. For better visualization, we just show the distribution of $\frac{A^h_{j,s}}{\max_{k\ne s}\{A^h_{j,k}\}}$ within the range of 0~10.}
    \vspace{-5pt}
    \label{fig: distribution_of_tau}
\end{figure}

The setting of the threshold $\tau=2.2$ for the value of $\frac{A^h_{j,s}}{\max_{k\ne s}\{A^h_{j,k}\}}$ is based on our observation in the distribution of values of $\frac{A^h_{j,s}}{\max_{k\ne s}\{A^h_{j,k}\}}$. Using input sentences and corresponding triplets in the AGENDA test set, we computed the value of $\frac{A^h_{j,s}}{\max_{k\ne s}\{A^h_{j,k}\}}$ at all heads $h$ and all current tokens $t_j$ that satisfy $s=\arg\max_k \{A^h_{j,k}\}$. The distribution of this value is shown in Figure~\ref{fig: distribution_of_tau}. The frequency of values larger than 2.2 dropped significantly to less than 5\%. Thus, we empirically set the threshold $\tau=2.2$ to only focus on attention heads that \textit{exclusively attended to the head token}.

Moreover, we also conduct ablation experiments with different values of $\tau$. Specifically, we compute the relation index for syntactic dependencies on InternLM2-1.8B with a smaller value $\tau=2.0$ and a larger value $\tau=2.5$, respectively. Heatmaps in Figure~\ref{fig: different_tau} show that the setting of $\tau$ does not significantly affect the distribution of the relation index. Different settings of $\tau$ yield a similar set of semantic induction heads that have a high relation index.

\section{Relation index for the reverse semantic relationships}
\label{sec: app_reverse_relation}
The semantic relationships in knowledge graphs are bidirectional, thus we also compute the relation index of attention heads for the reverse semantic relationships. Figure~\ref{fig: vis_of_reverse_relation_score} shows the results in InternLM2-1.8B.
Comparing Figure~\ref{fig: vis_of_reverse_relation_score} and Figure~\ref{fig: vis_of_relation_score}, we find that some attention heads represent both directions of the relation.

\section{Format compliance ability}
\label{sec: app_format_shot_num}
\begin{figure}[h]
    \centering
    \includegraphics[width=0.8\linewidth]{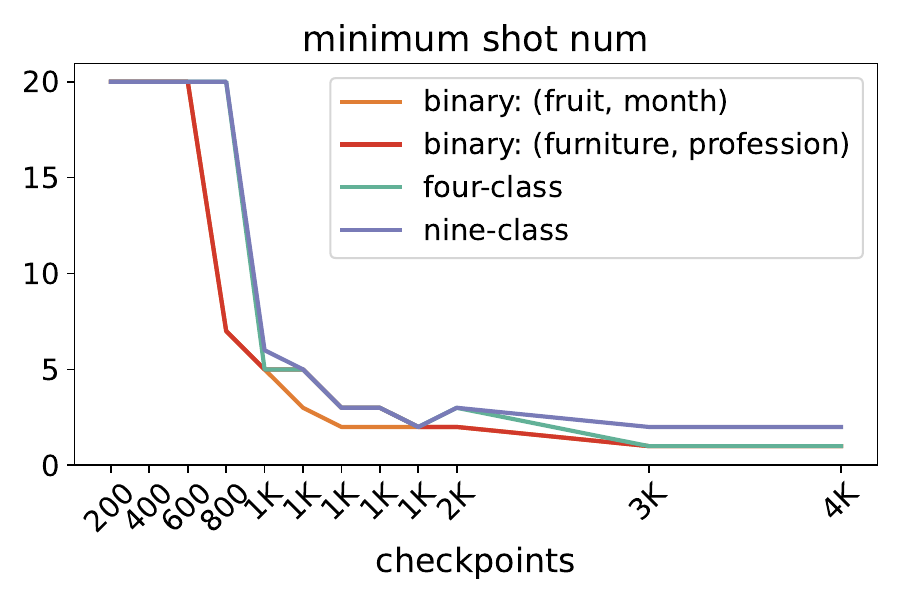}
    \vspace{-5pt}
    \caption{The minimum number of shots (examples in the prompt) required to achieve over 80\% format accuracy. }
    \label{fig: format_minimum_number}
\end{figure}

Besides the format accuracy in Figure~\ref{fig: format_acc}, we also measure the format compliance ability from another perspective: the minimum number of shots required to achieve a format accuracy of over 80\%. A lower minimum number of shots indicates a better format compliance ability.
We set a maximum limit of 20 shots. If the model fails to achieve an accuracy of 80\% even with 20 shots, we record the result as 20.
Figure~\ref{fig: format_minimum_number} consistently shows that the model learns format compliance on simple tasks earlier than on complex tasks, but all achieve a good performance at 1000 steps.

Moreover, considering that the format of generating "true" or "false" in relation justification tasks is different from the simpler format in classification tasks in Figure~\ref{fig: format_acc}, we additionally examine the format accuracy on relation justification tasks in Figure~\ref{fig: format_rel_acc}. The format compliance in these tasks emerges from about 2K steps, later than that in simpler tasks in Figure~\ref{fig: format_acc}, and achieves about 50\%-70\% at 15K steps.

\section{Pattern discovery tasks are not binary classification tasks}
\label{sec: app_pattern_tasks}

Unlike binary classification tasks, pattern discovery tasks are actually more difficult for generative models. 

First, the model generates the next token as the prediction, which is different from the classic binary task. When generating the next token, there are a total of $\vert\mathcal{V}\vert=92544$ candidates in the vocabulary $\mathcal{V}$. If the model could perfectly comply with the format, the problem is simplified as a binary classification task. However, models in the early stages of training do not have such ideal capabilities yet. For example, Figure~\ref{fig: format_rel_acc} shows that the format compliance in relation justification tasks emerges from about 2K steps, later than that in simpler tasks in Figure 6, and achieves about 50\%-70\% at 15K steps. Thus, the exact prediction accuracy of the model will be lower.
Second, the model might inherit certain biases from the training data. Thus, the probability of generating either the token ``true'' or ``false'' is not 0.5 vs 0.5.

\section{Pattern discovery ability of a well-trained model}
\label{sec: app_luyou_icl}
\begin{figure}[h]
    \centering
    \includegraphics[width=0.8\linewidth]{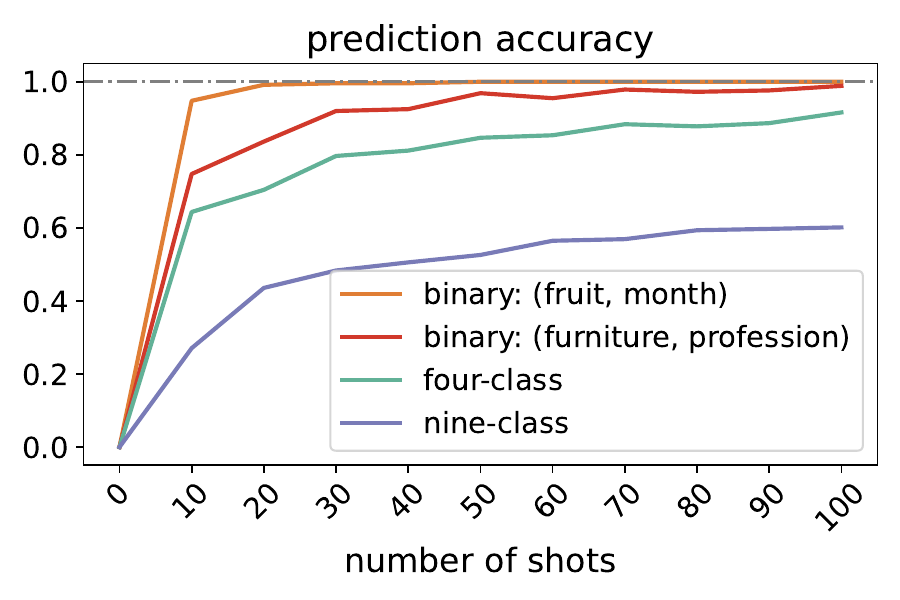}
    \vspace{-10pt}
    \caption{Prediction accuracy of InternLM2-1.8B on classification tasks.}
    \vspace{-10pt}
    \label{fig: luyou_icl_class_acc}
\end{figure}

\begin{figure}[h]
    \centering
    \includegraphics[width=0.8\linewidth]{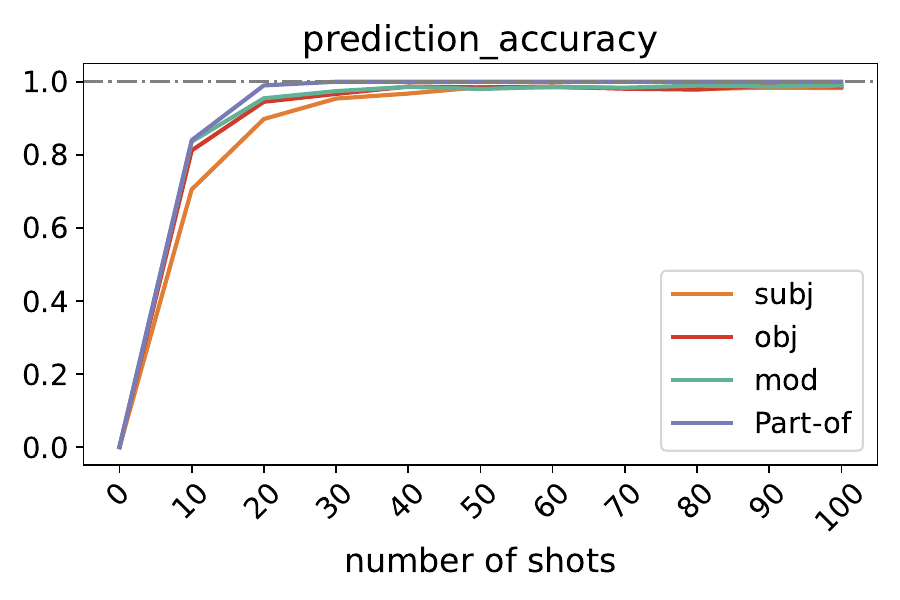}
    \vspace{-10pt}
    \caption{Prediction accuracy of InternLM2-1.8B on relation justification tasks.}
    \vspace{-10pt}
    \label{fig: luyou_icl_rel_acc}
\end{figure}

\begin{figure*}
    \setlength{\abovecaptionskip}{4pt}
    \setlength{\belowcaptionskip}{-5pt}
    \centering
    \includegraphics[width=0.8\linewidth]{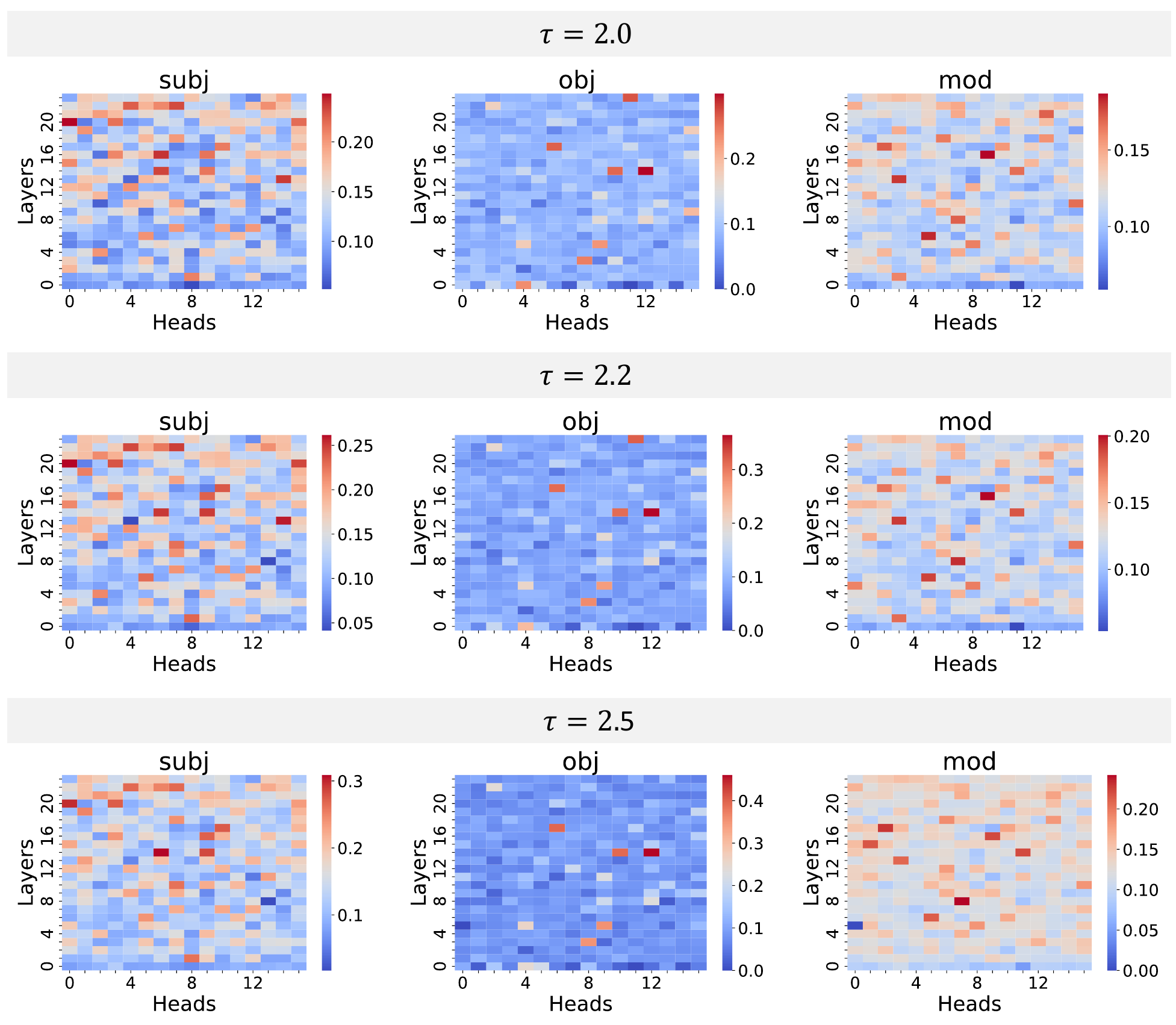}
    \caption{Heatmaps of relation indexes with different settings of $\tau$. Different settings of $\tau$ yield a similar set of semantic induction heads that have a high relation index.}
    \label{fig: different_tau}
\end{figure*}

\begin{figure*}
    \setlength{\abovecaptionskip}{4pt}
    \setlength{\belowcaptionskip}{-5pt}
    \centering
    \includegraphics[width=0.245\linewidth]{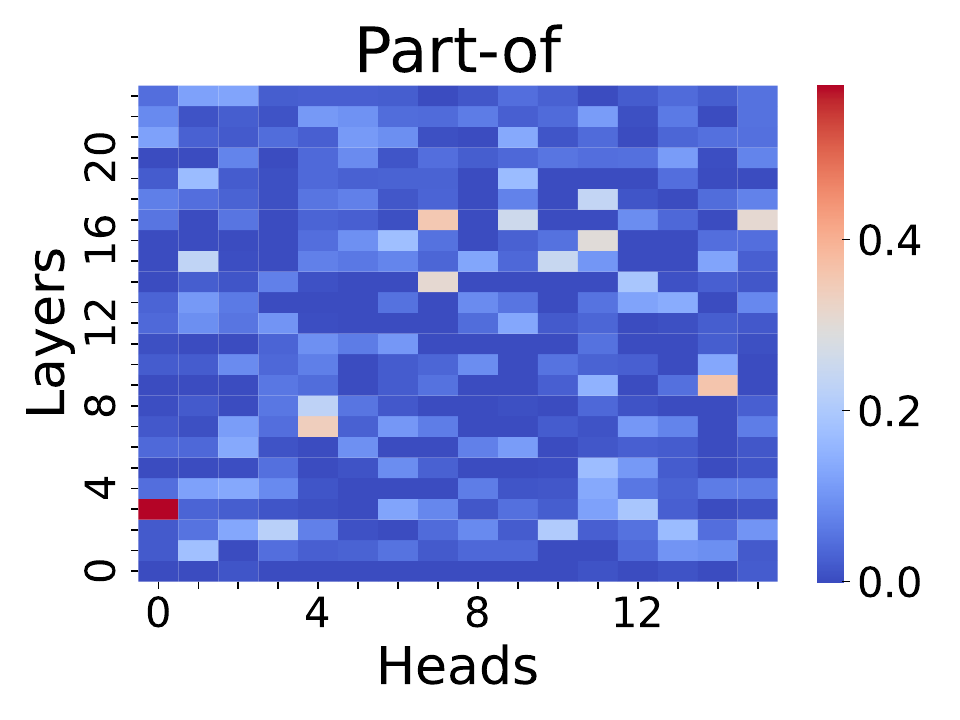}
    \includegraphics[width=0.245\linewidth]{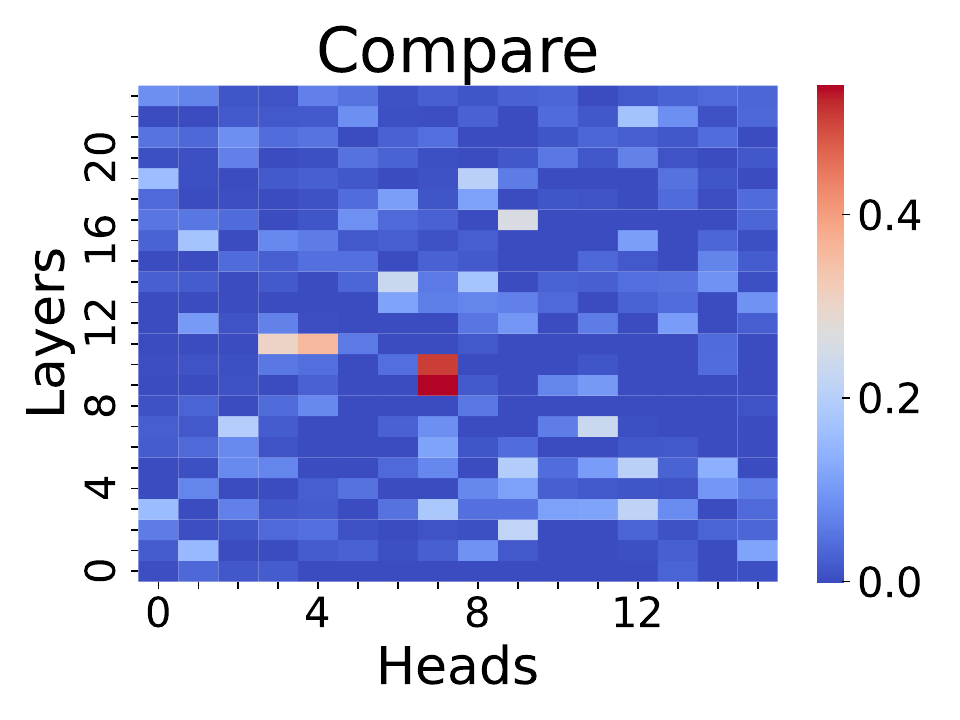}
    \includegraphics[width=0.245\linewidth]{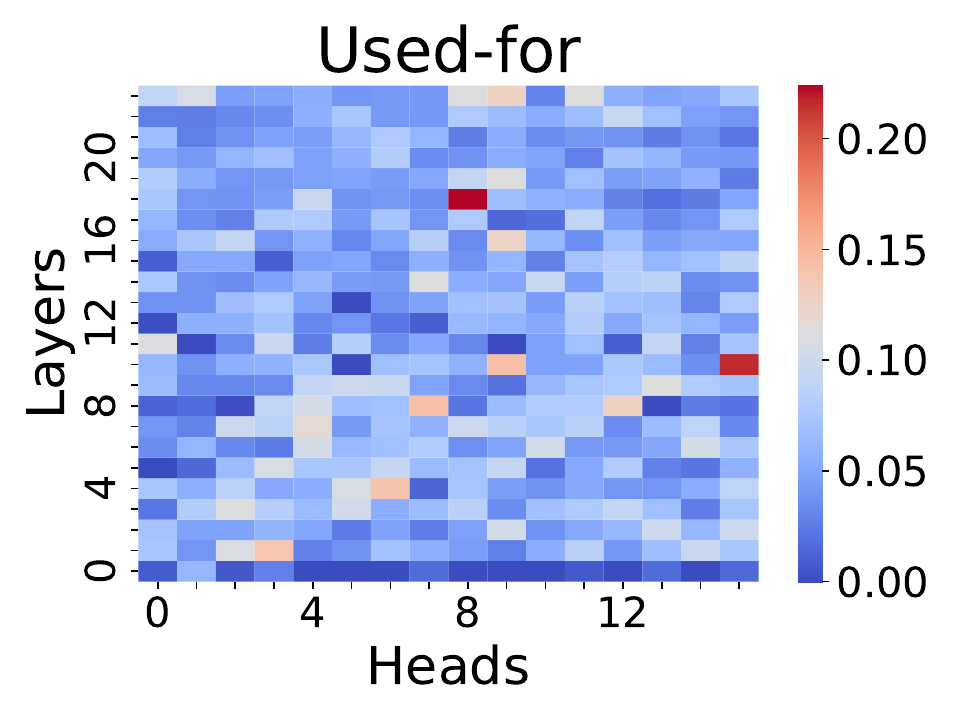}
    \includegraphics[width=0.245\linewidth]{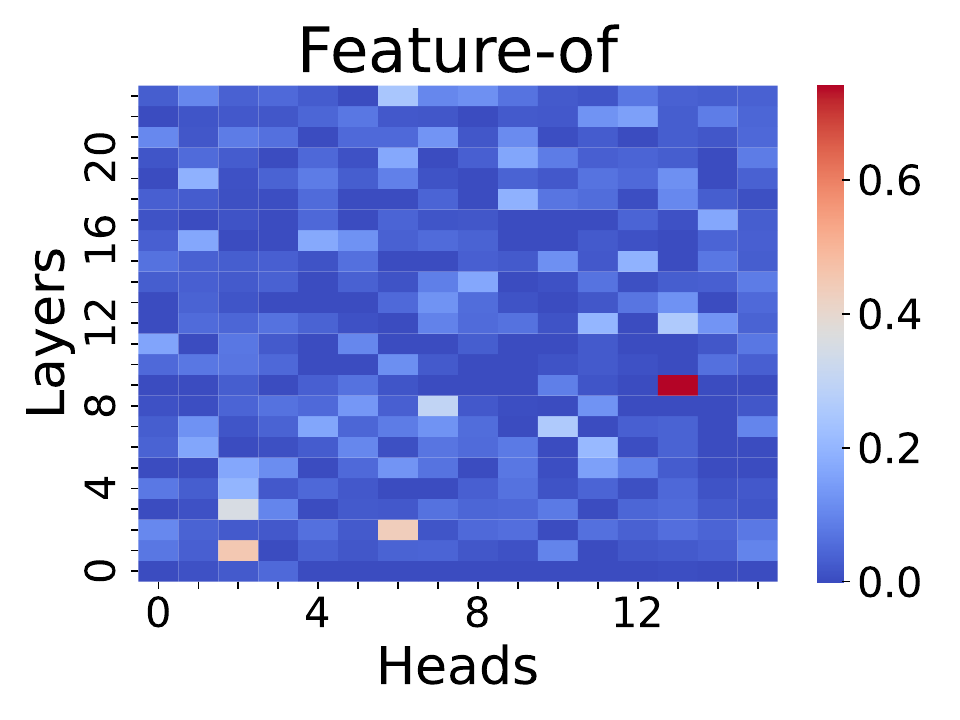}
    \includegraphics[width=0.245\linewidth]{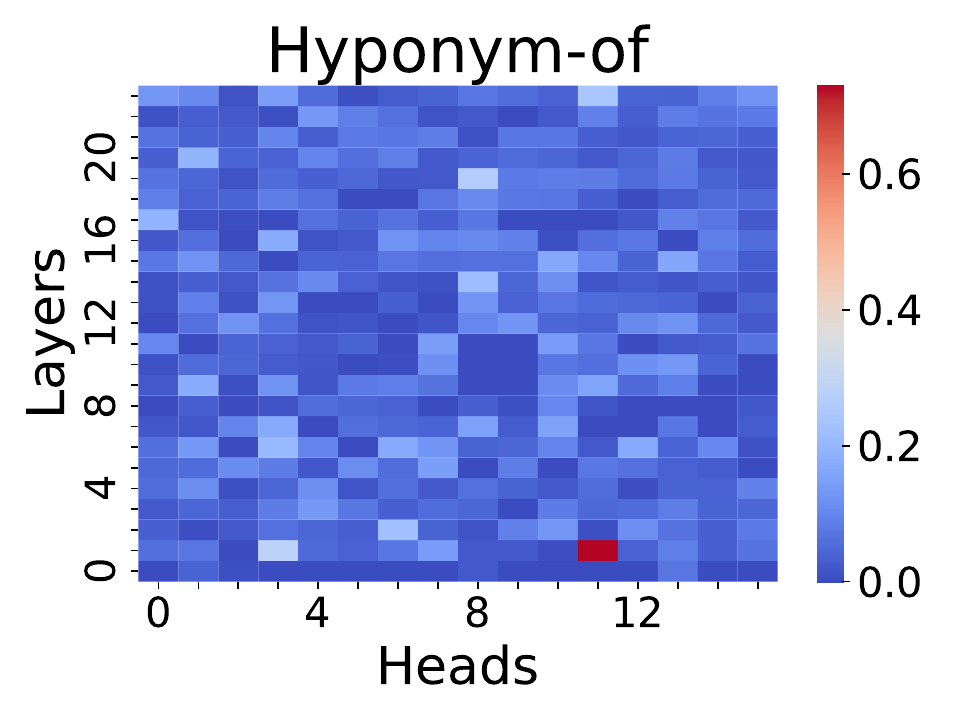}
    \includegraphics[width=0.245\linewidth]{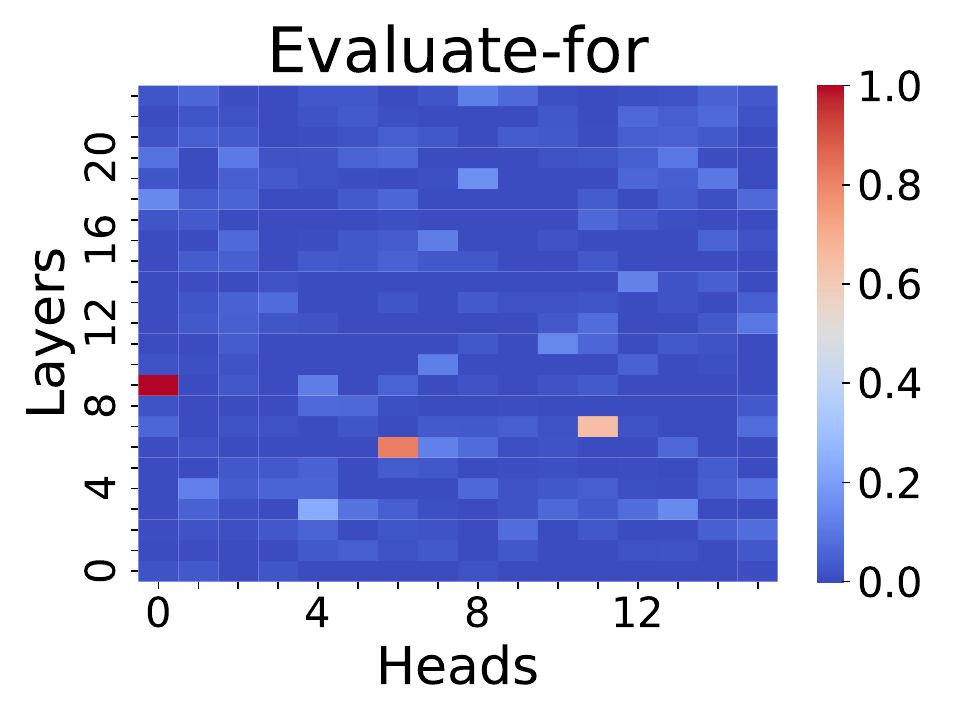}
    \includegraphics[width=0.245\linewidth]{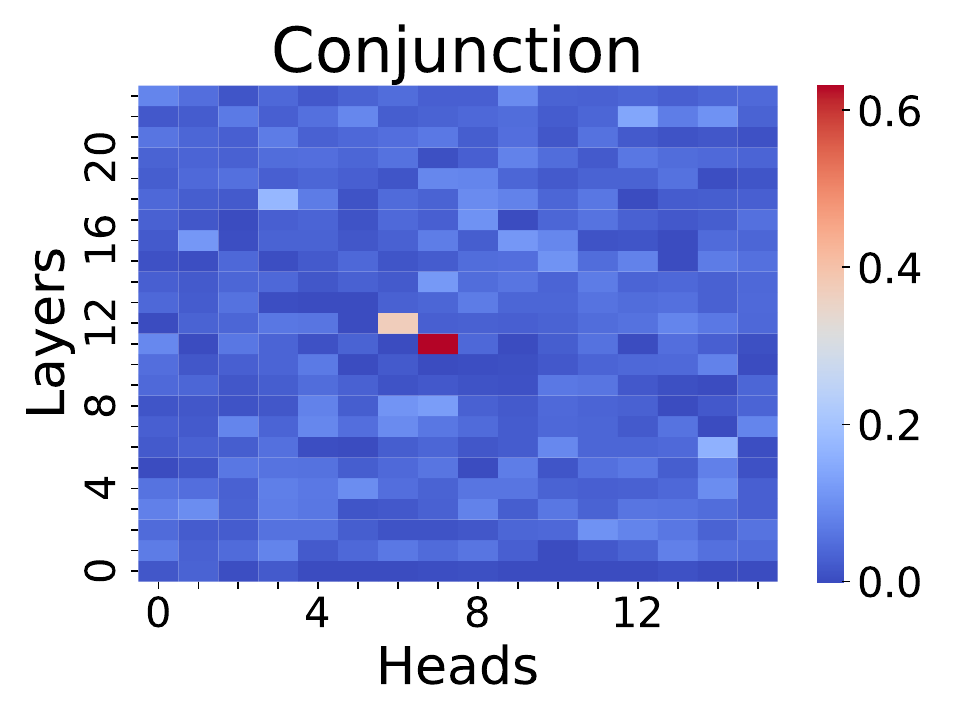}
    \caption{Heatmaps of the average relation index of attention heads for the reverse semantic relationships in knowledge graphs.}
    \label{fig: vis_of_reverse_relation_score}
\end{figure*}

\begin{figure*}
    \setlength{\abovecaptionskip}{4pt}
    \setlength{\belowcaptionskip}{-5pt}
    \centering
    \includegraphics[width=0.245\linewidth]{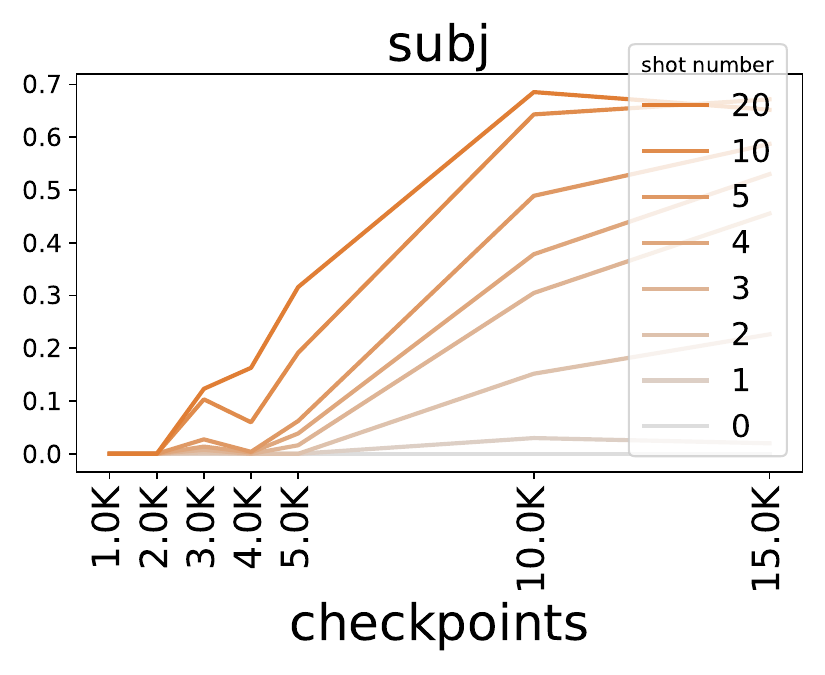}
    \hfill
    \includegraphics[width=0.245\linewidth]{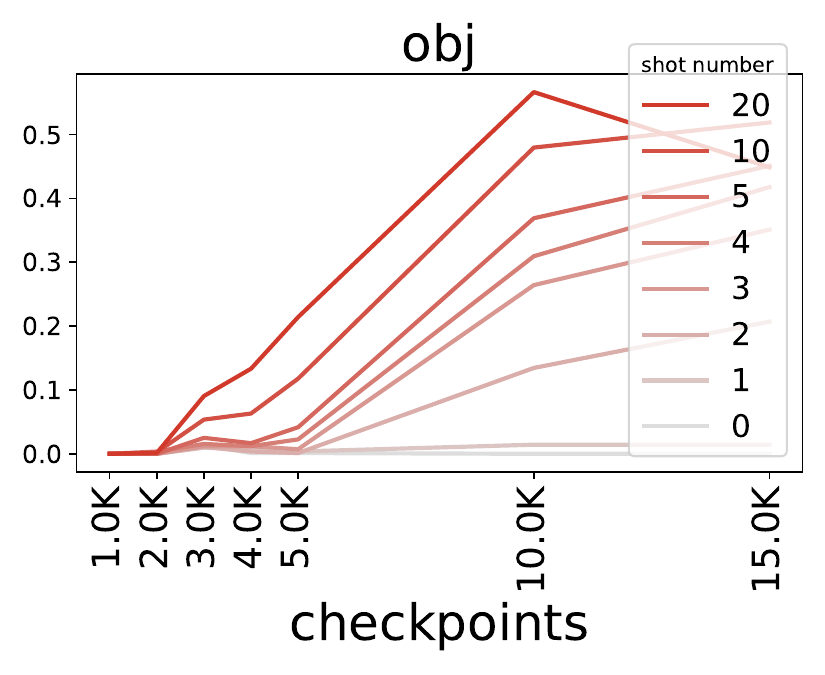}
    \hfill
    \includegraphics[width=0.245\linewidth]{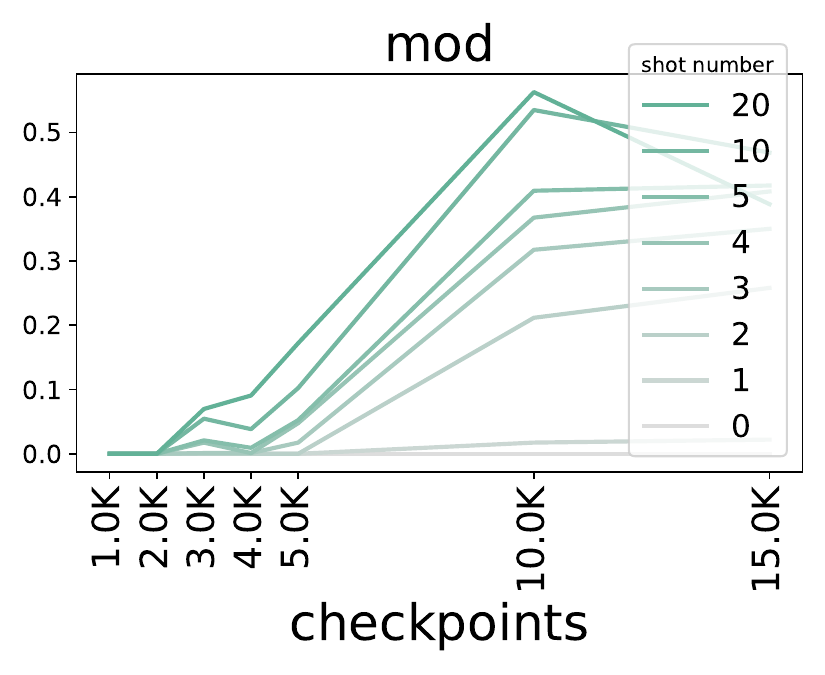}
    \hfill
    \includegraphics[width=0.245\linewidth]{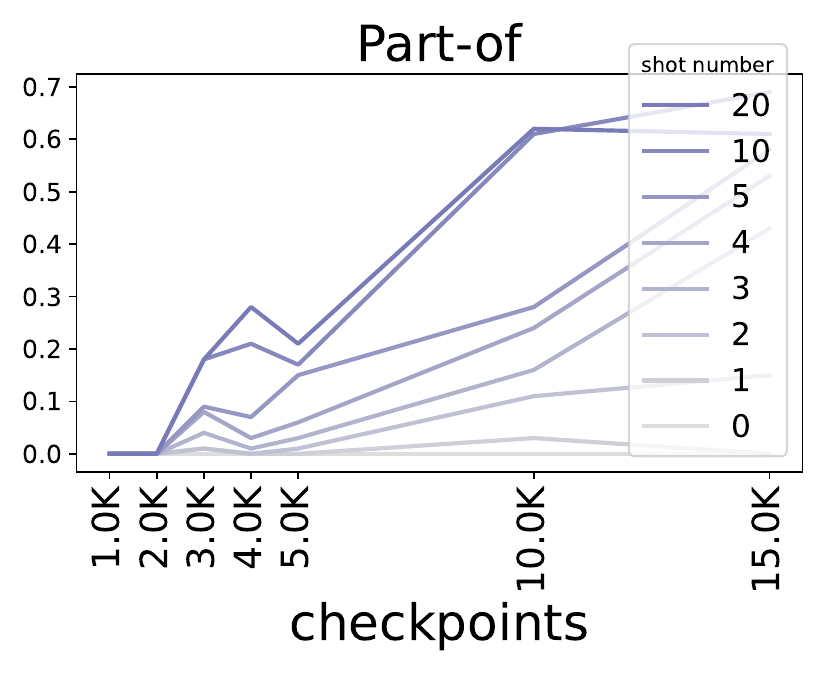}
    \caption{Format accuracy of different relation justification tasks at different checkpoints. Each line represents the format accuracy with different numbers of shots (examples) in the prompt.}
    \label{fig: format_rel_acc}
\end{figure*}

Although we have observed the development of the pattern discovery ability in Figure~\ref{fig: pattern_acc}, we find that it is hard to be fully mastered by the model.
Figure~\ref{fig: luyou_icl_class_acc} reports the prediction accuracy of the well-trained InternLM2-1.8B on classification tasks.
Even with 100 examples in the prompt, the model still cannot perfectly recognize and utilize the pattern in the prompt.
On the other hand, Figure~\ref{fig: luyou_icl_rel_acc} shows that InternLM2-1.8B achieves higher accuracy on relation justification tasks.


\section{Change of relation index for semantic relationships over training time}
\label{sec: app_relation_curve}
\begin{figure*}
    \setlength{\abovecaptionskip}{4pt}
    \setlength{\belowcaptionskip}{-5pt}
    \centering
    \includegraphics[width=0.245\linewidth]{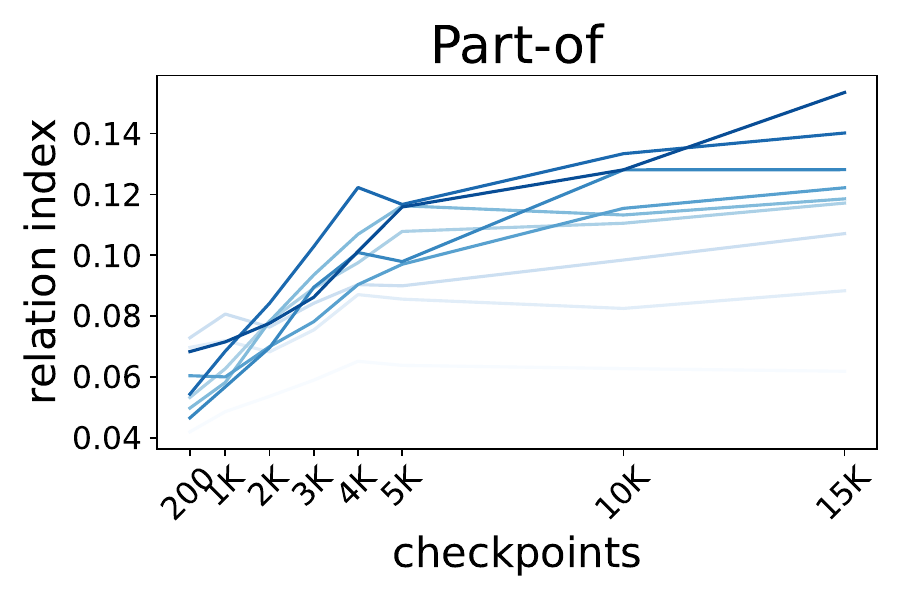}
    \includegraphics[width=0.245\linewidth]{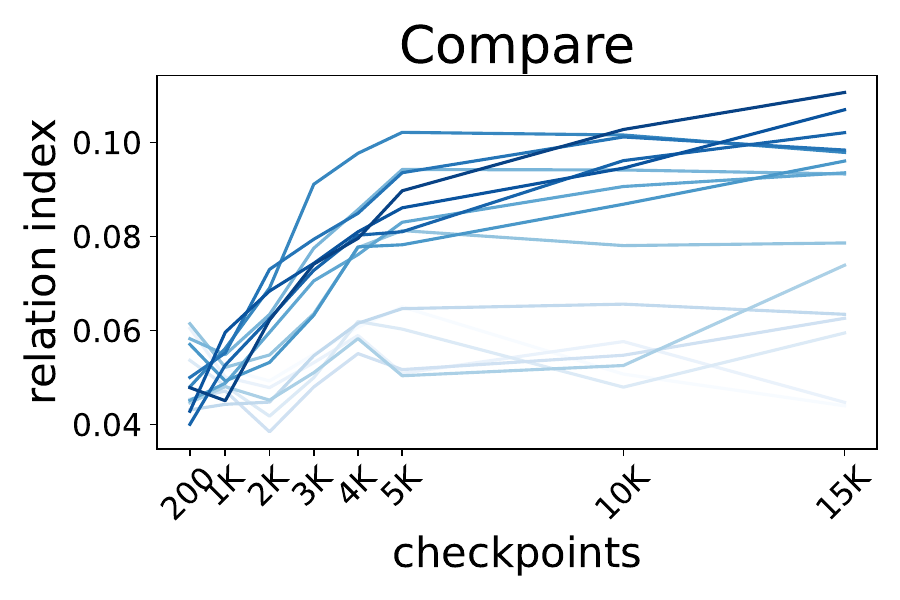}
    \includegraphics[width=0.245\linewidth]{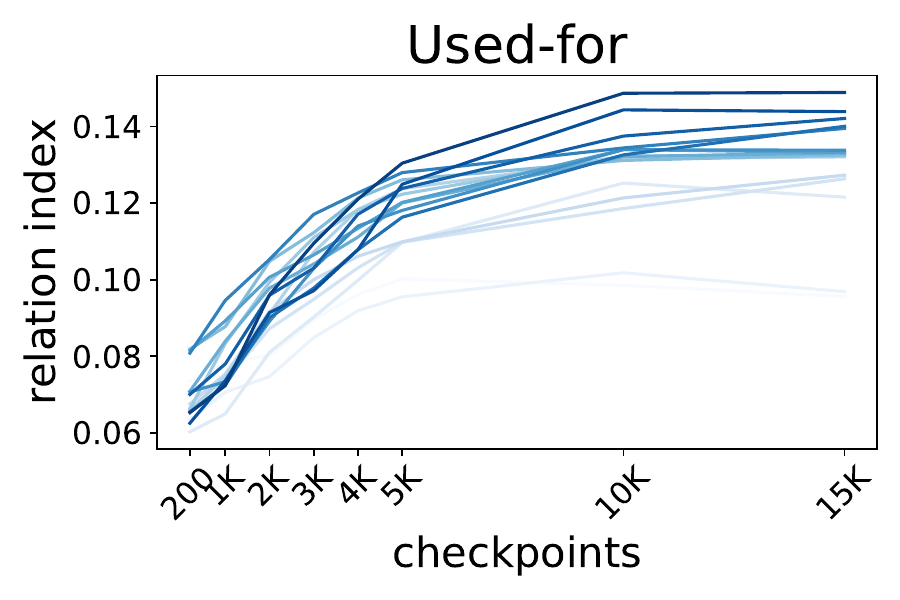}
    \includegraphics[width=0.245\linewidth]{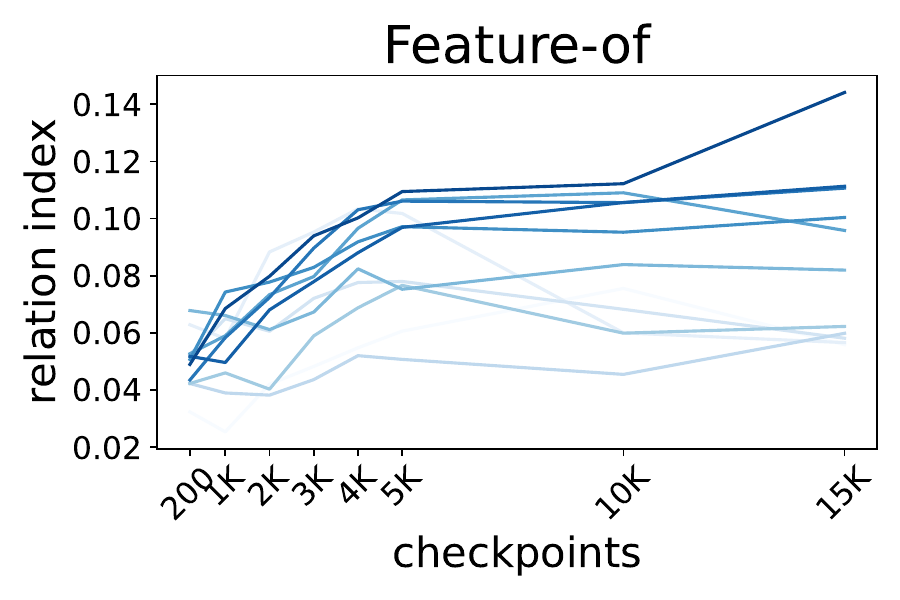}
    \includegraphics[width=0.245\linewidth]{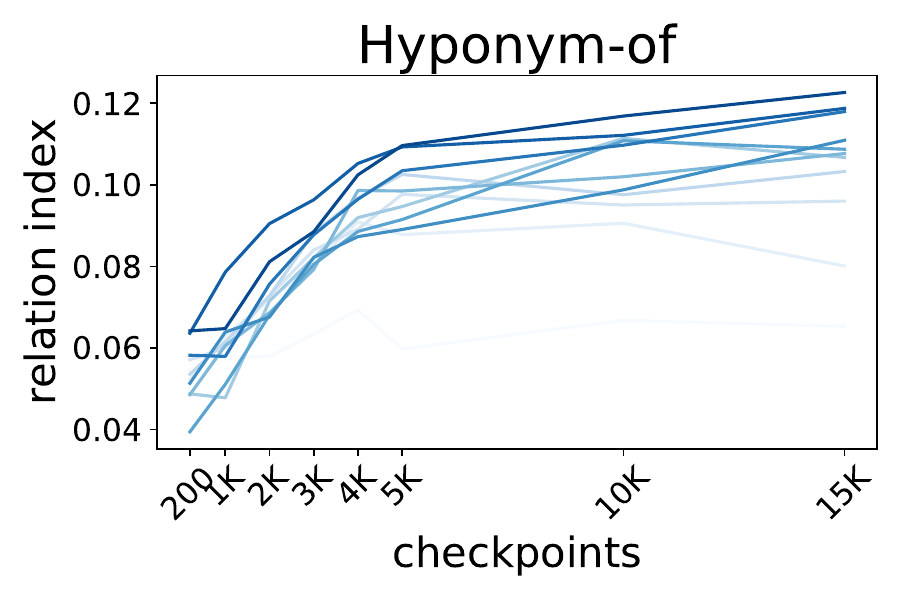}
    \includegraphics[width=0.245\linewidth]{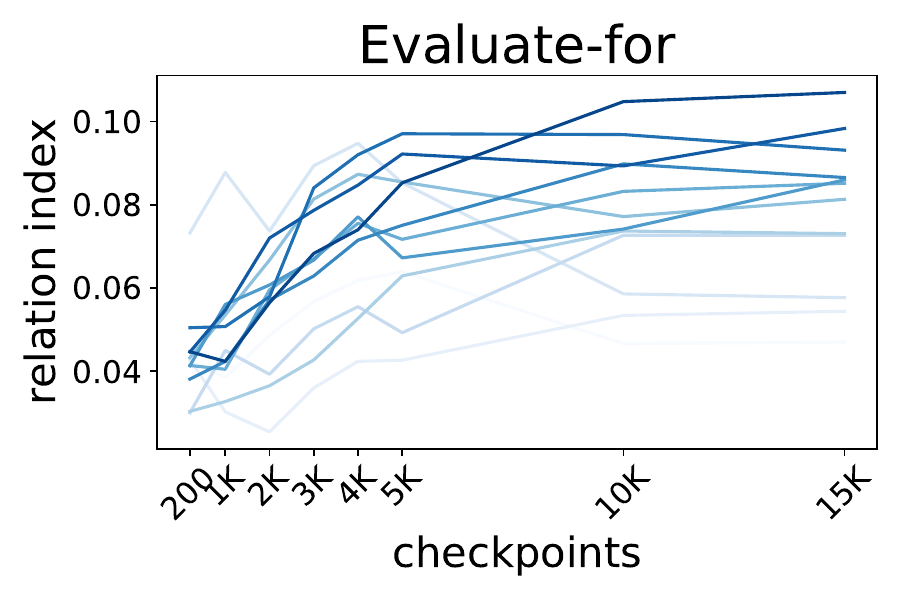}
    \includegraphics[width=0.245\linewidth]{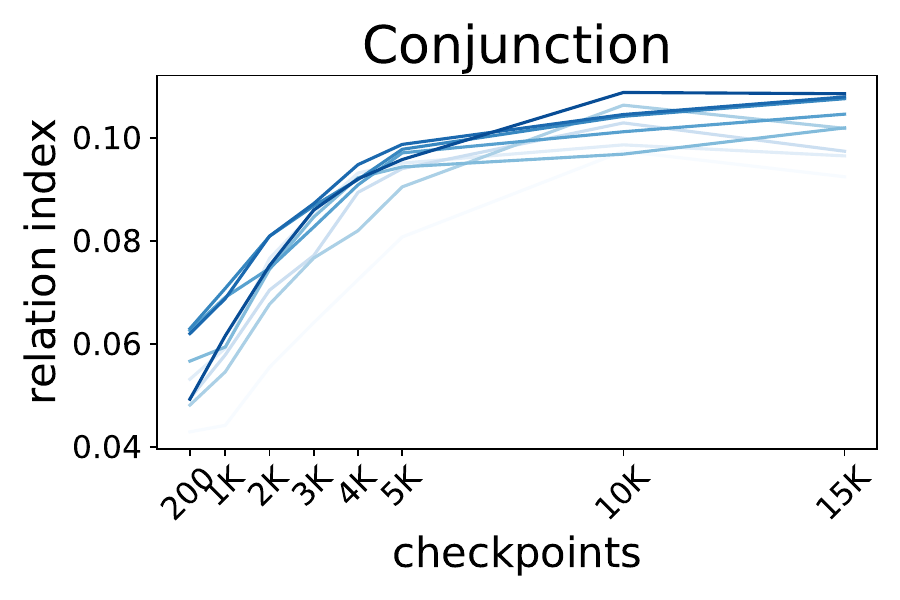}
    \caption{The change curve of relation indexes of attention heads for semantic relationships.}
    \label{fig: curve_of_relation}
\end{figure*}

\begin{figure*}
    \setlength{\abovecaptionskip}{4pt}
    \setlength{\belowcaptionskip}{-5pt}
    \centering
    \includegraphics[width=0.245\linewidth]{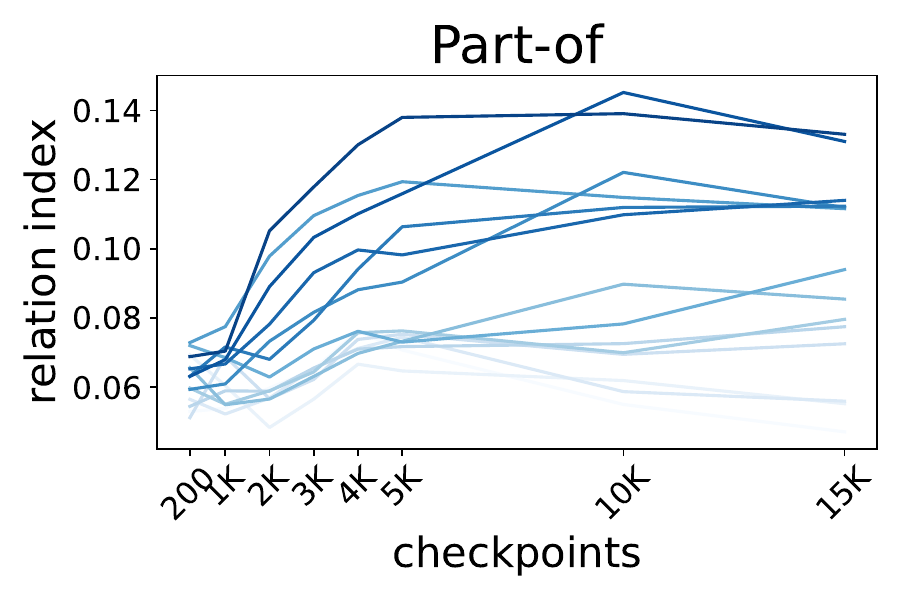}
    \includegraphics[width=0.245\linewidth]{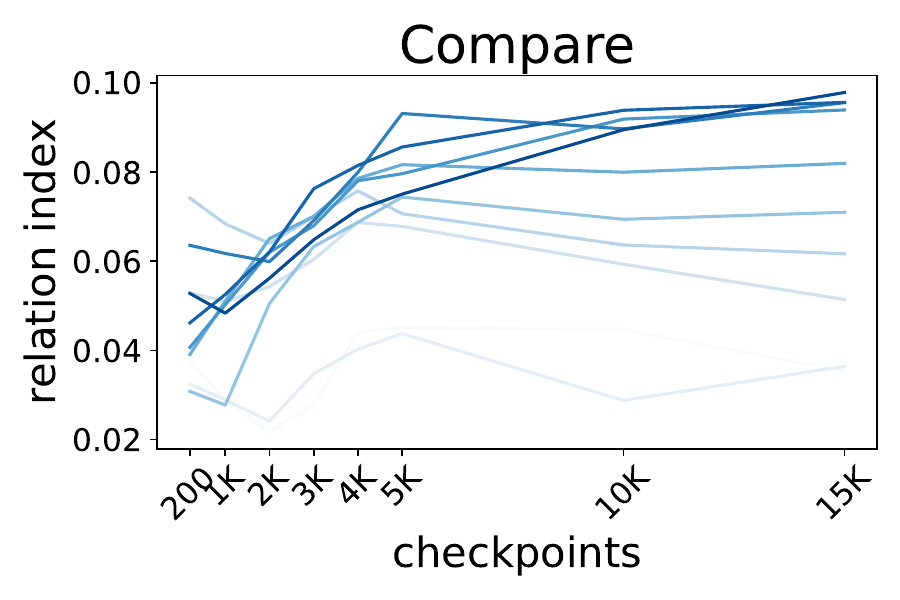}
    \includegraphics[width=0.245\linewidth]{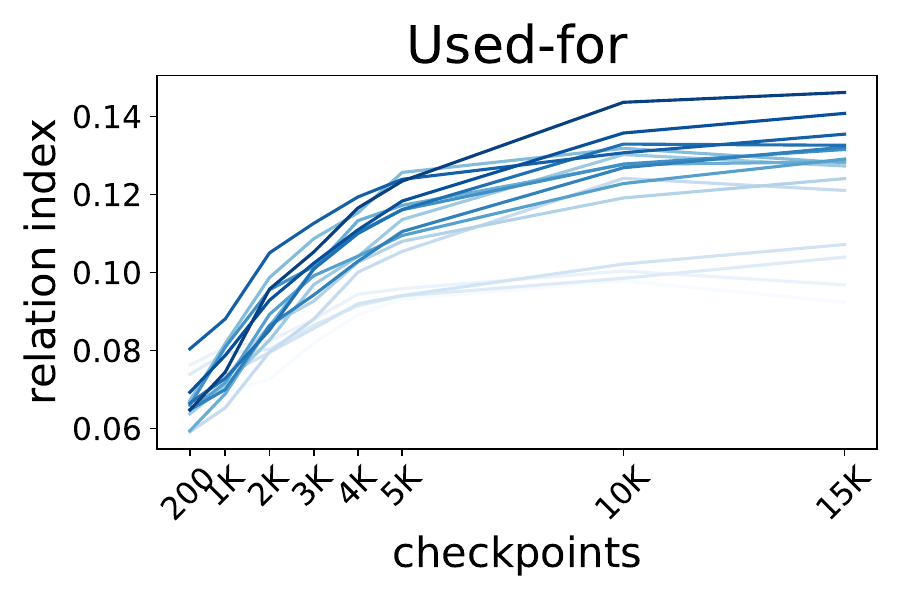}
    \includegraphics[width=0.245\linewidth]{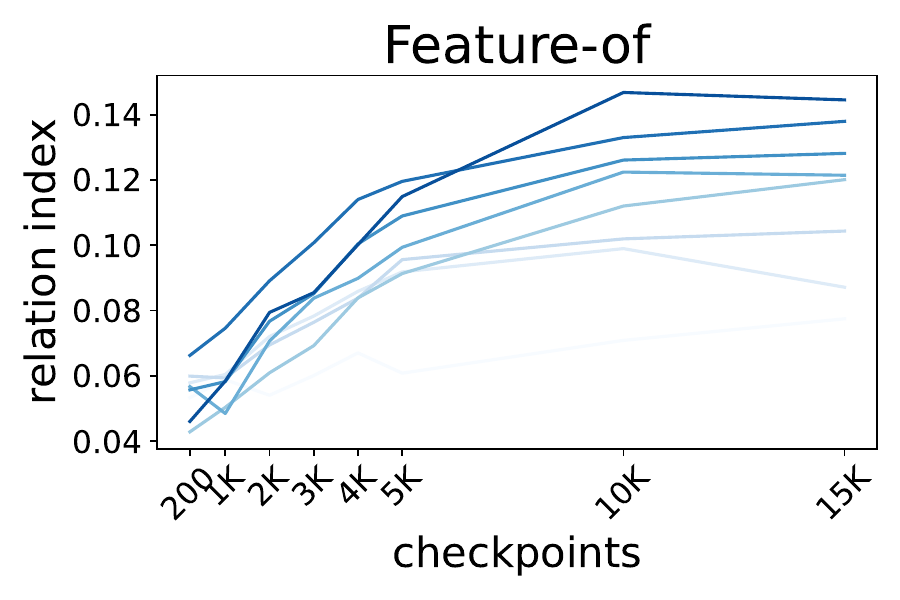}
    \includegraphics[width=0.245\linewidth]{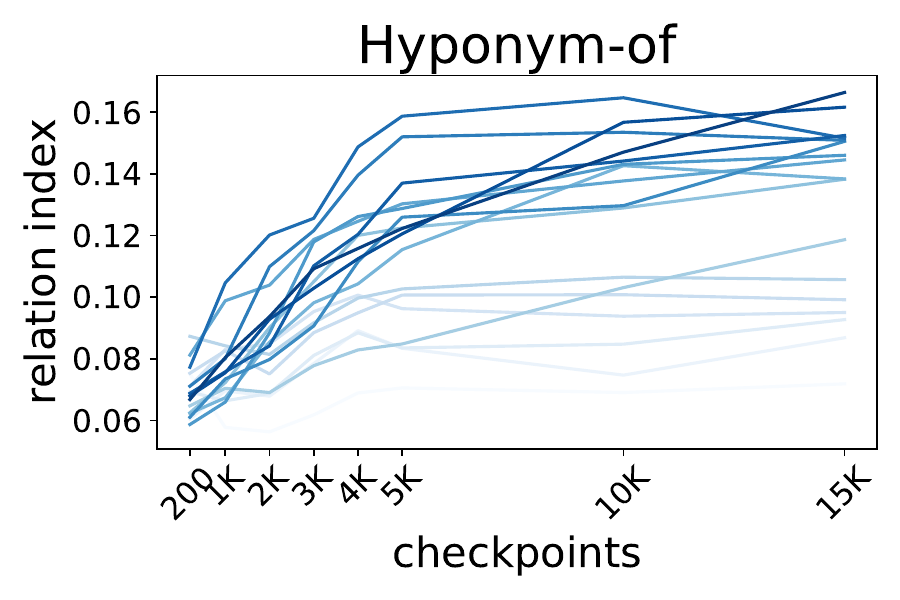}
    \includegraphics[width=0.245\linewidth]{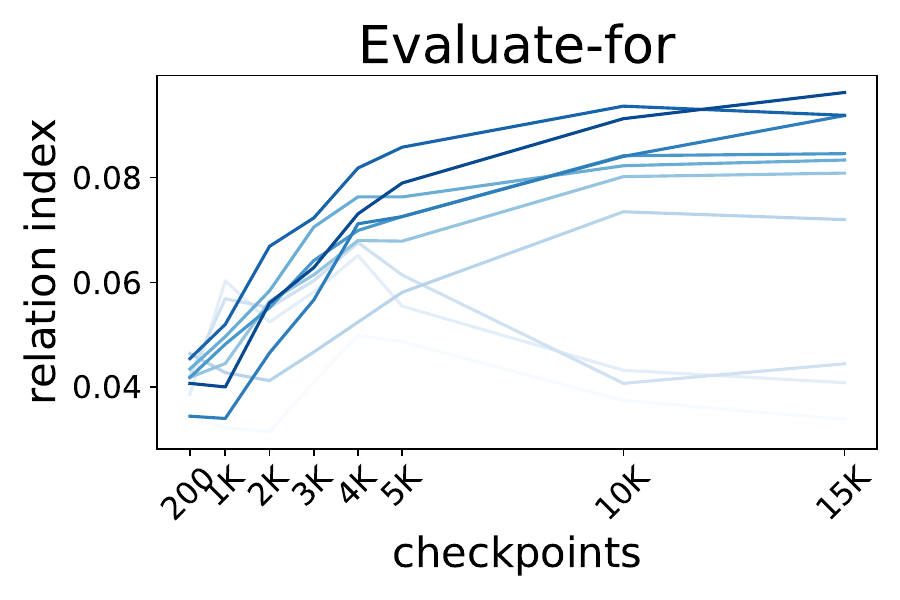}
    \includegraphics[width=0.245\linewidth]{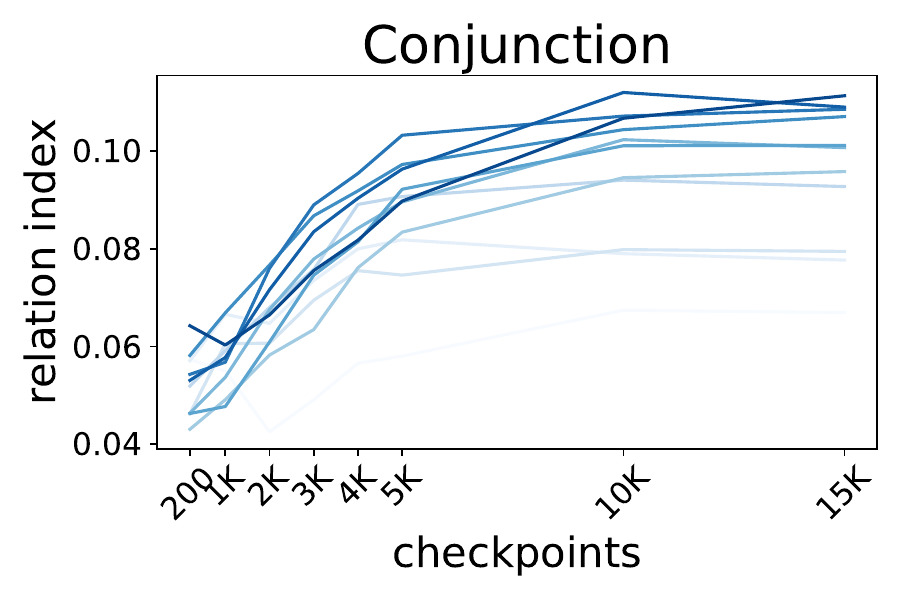}
    \caption{The change curve of relation indexes of attention heads for reverse semantic relationships.}
    \label{fig: curve_of_reverse_relation}
\end{figure*}

This section provides results of the change of relation indexes of attention heads for semantic relationships in knowledge graphs.
We consider both directions of the semantic relationship, and results are shown in Figure~\ref{fig: curve_of_relation} and Figure~\ref{fig: curve_of_reverse_relation}.

\section{Grouping attention heads for relations}
\label{sec: app_relation_group}
As discussed in Section~\ref{sec: grouping}, the semantic relationships are dependent on the context, so they may be stored in different attention heads given different contexts. In this section, we perform the grouping analysis on semantic relations in knowledge graphs.

\begin{figure*}
    \centering
    \includegraphics[width=0.245\linewidth]{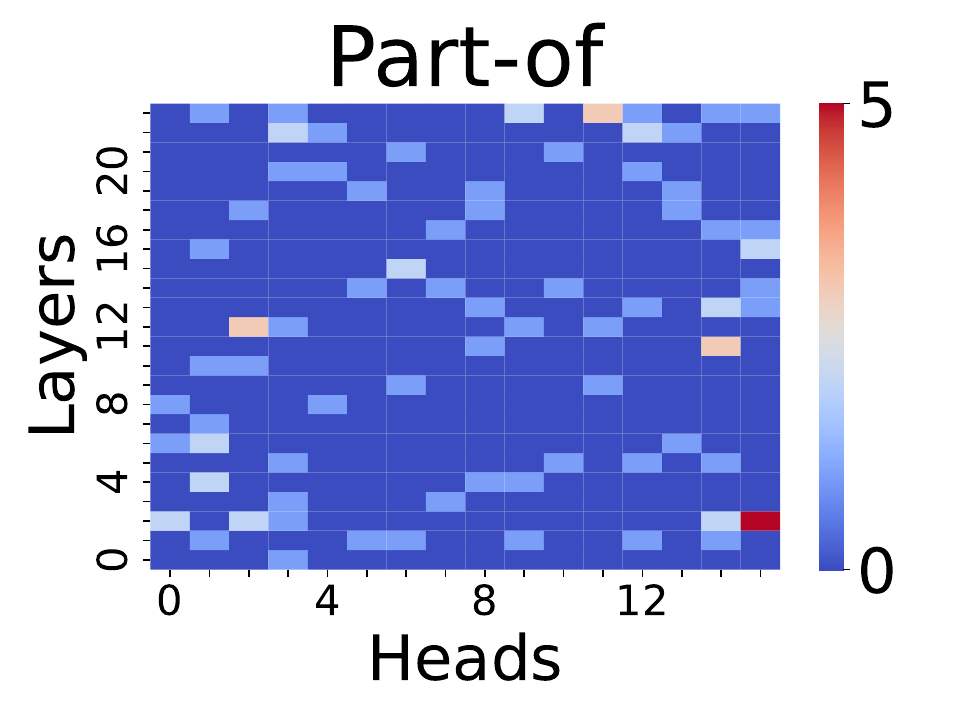}
    \includegraphics[width=0.245\linewidth]{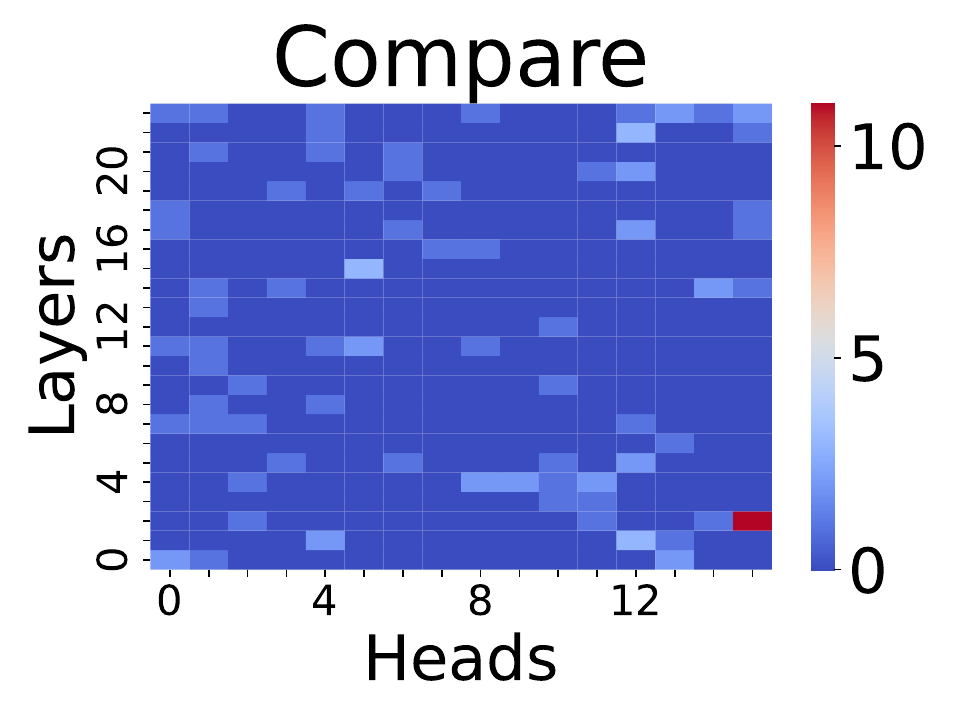}
    \includegraphics[width=0.245\linewidth]{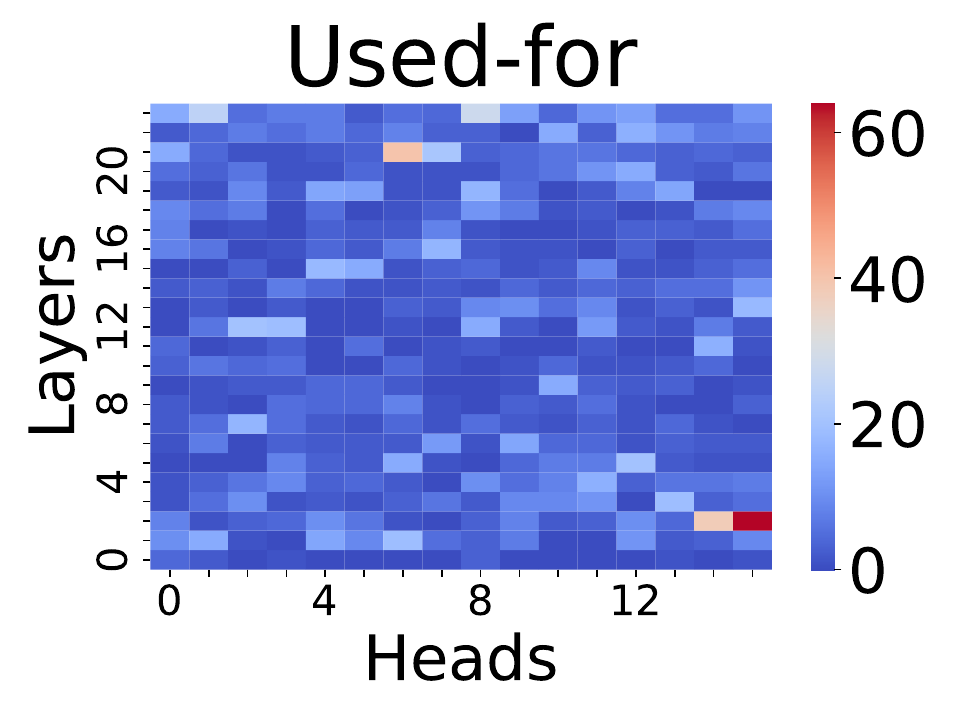}
    \includegraphics[width=0.245\linewidth]{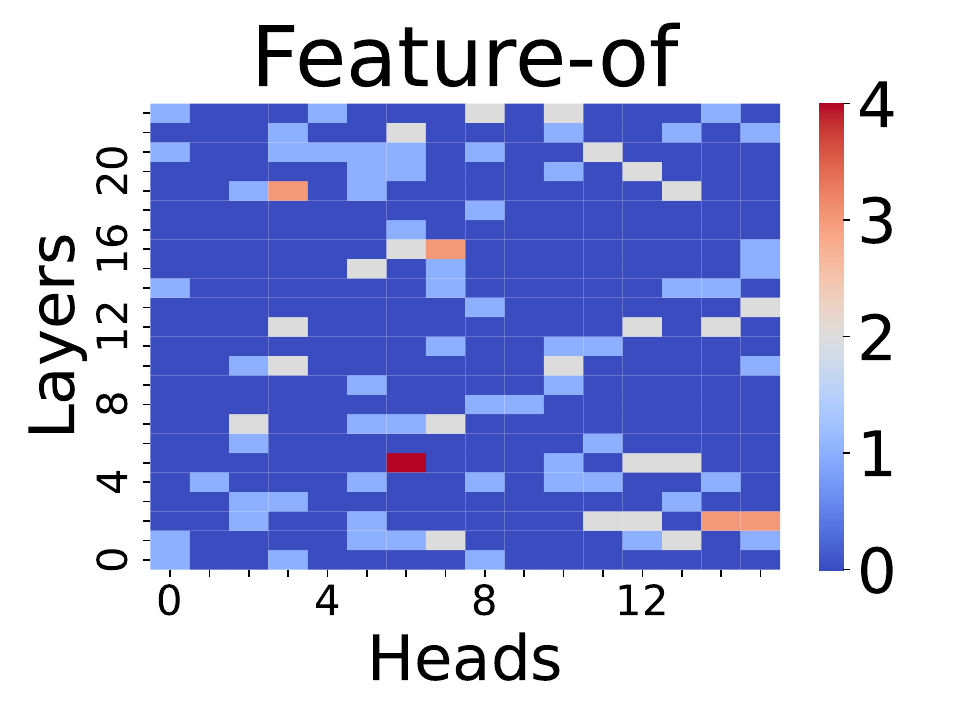}
    \includegraphics[width=0.245\linewidth]{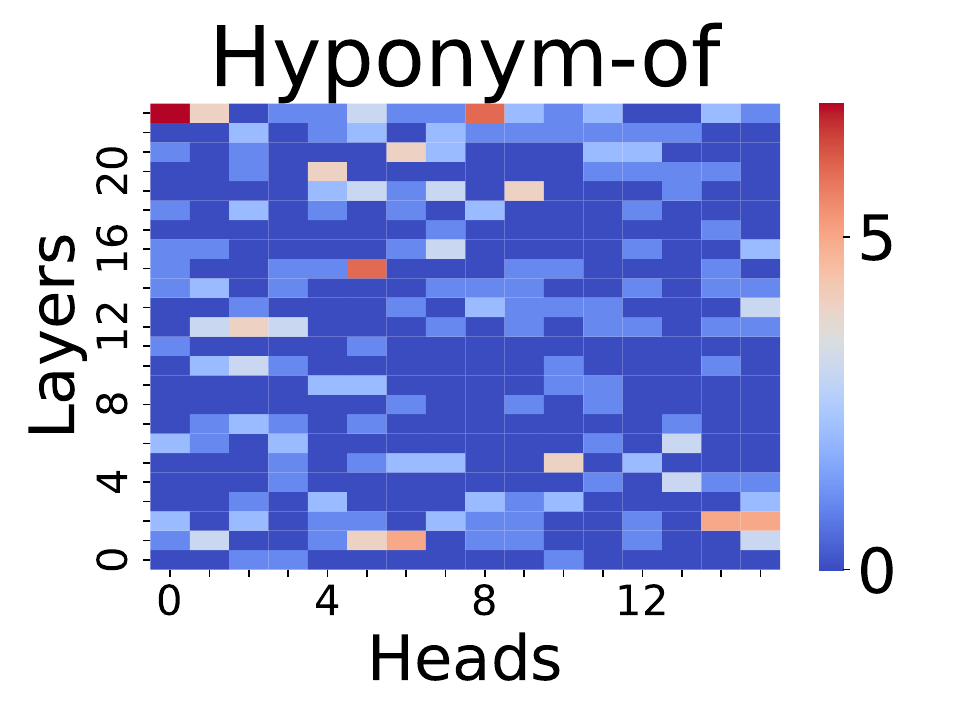}
    \includegraphics[width=0.245\linewidth]{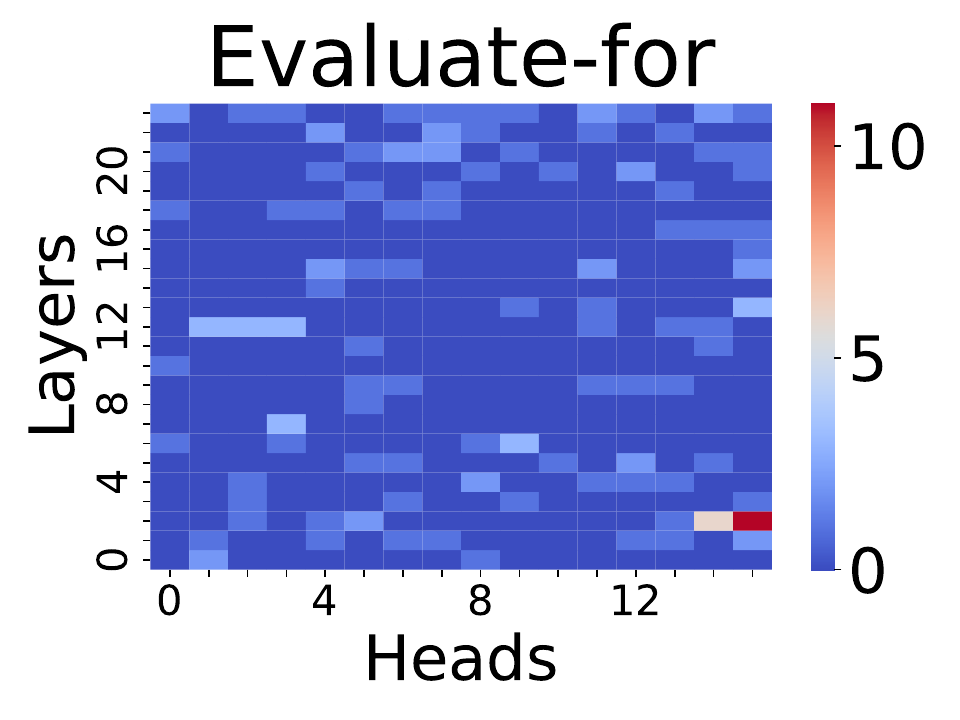}
    \includegraphics[width=0.245\linewidth]{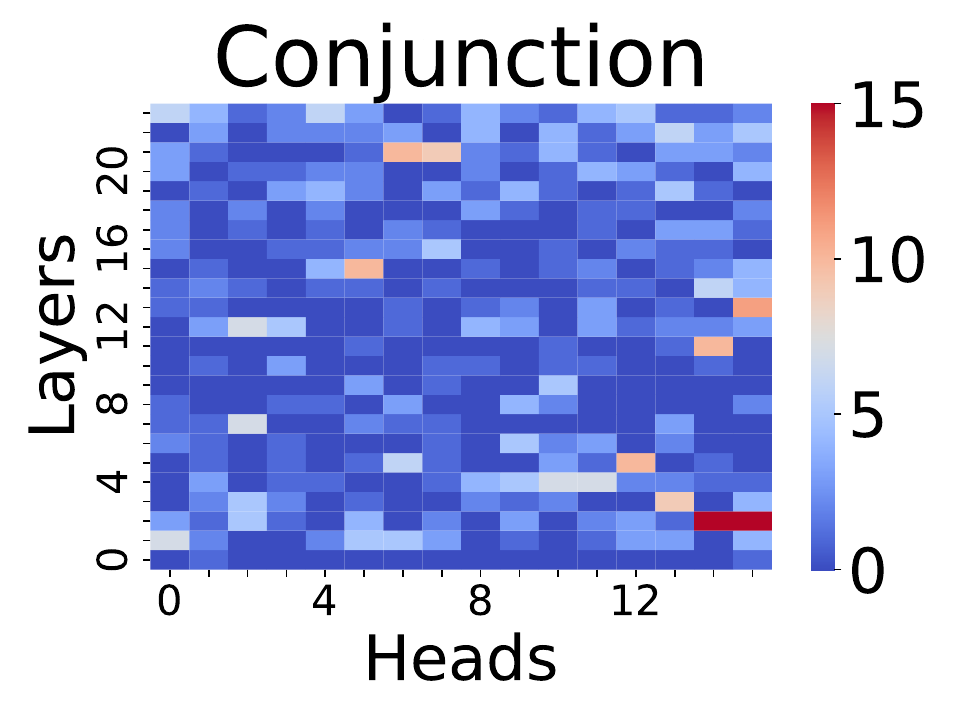}
    \caption{Heatmaps of occurrence of attention heads having the largest value of $\mathbb{E}_j[a^{h,j}_T]$ for each triplet in semantic relations.}
    \label{fig: vis_of_relation_group}
\end{figure*}

\begin{figure*}
    \centering
    \includegraphics[width=0.245\linewidth]{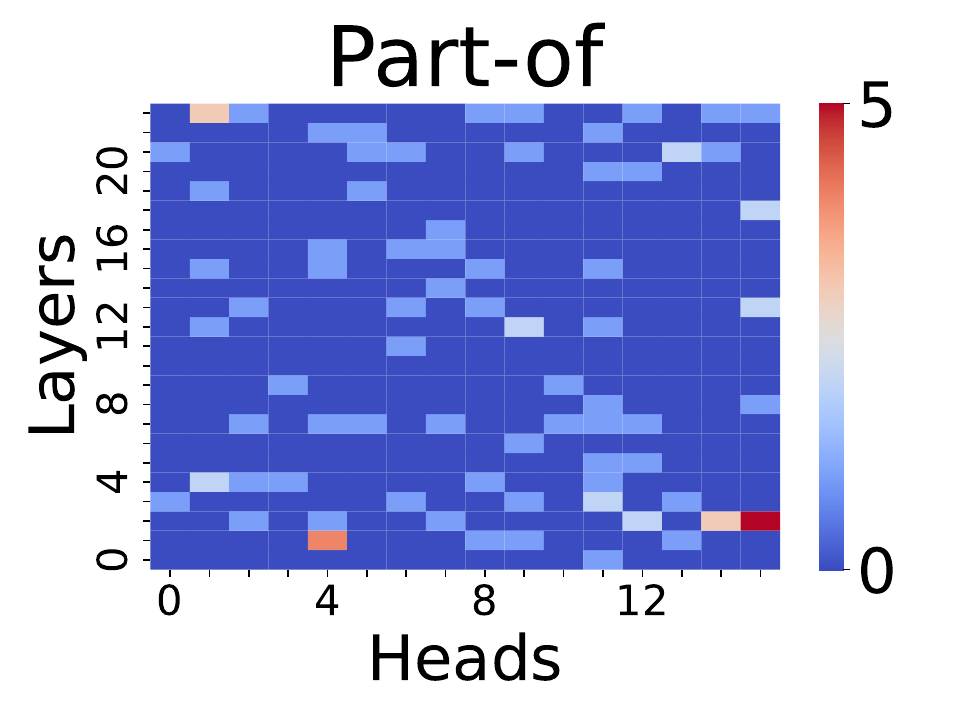}
    \includegraphics[width=0.245\linewidth]{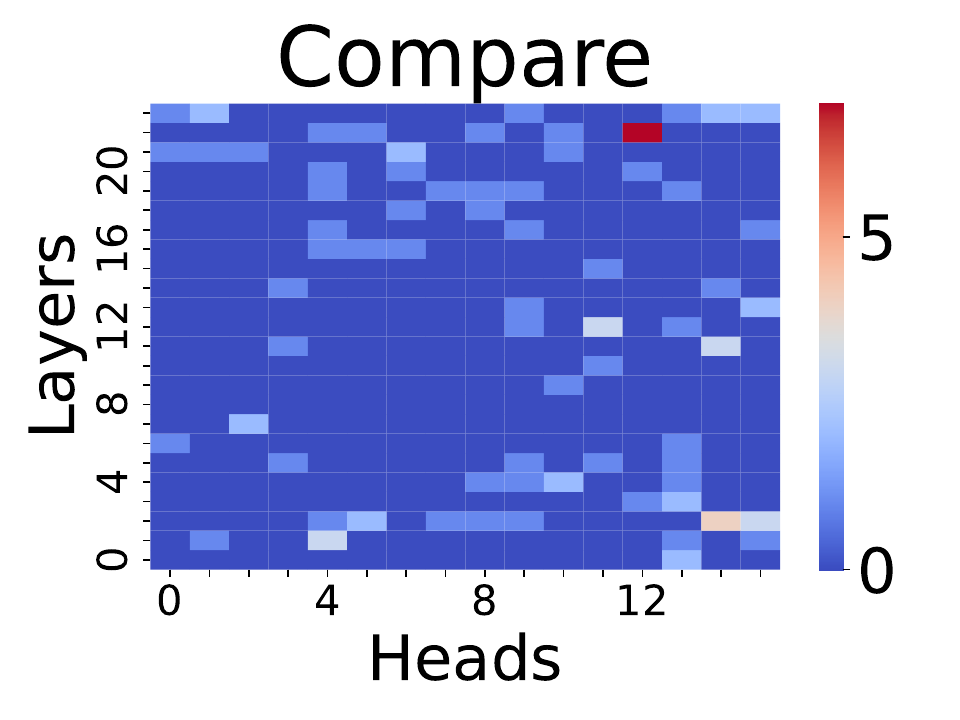}
    \includegraphics[width=0.245\linewidth]{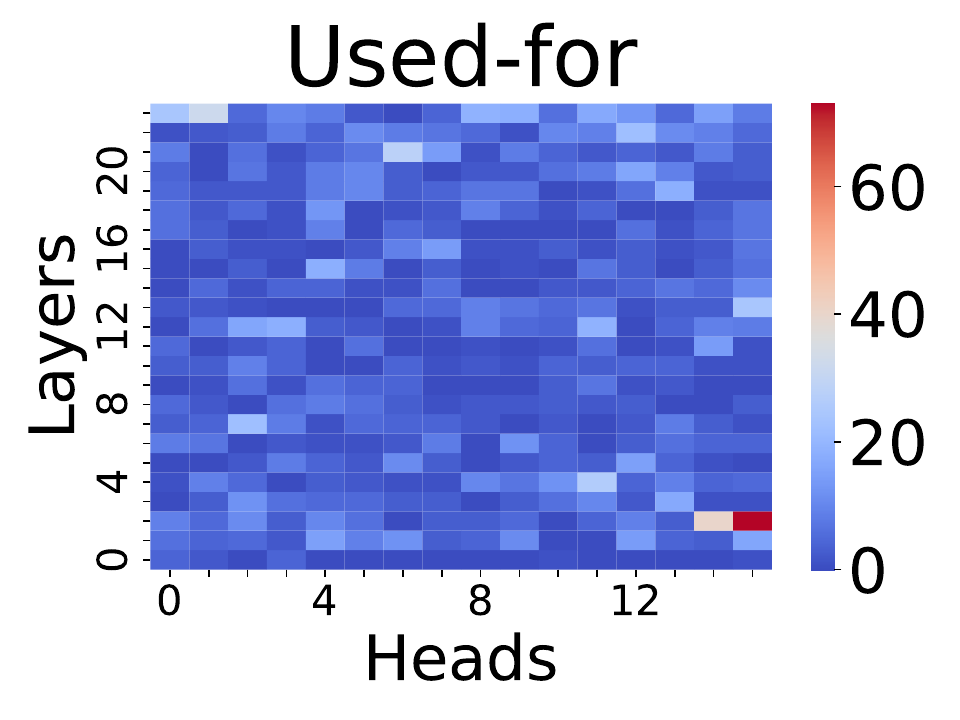}
    \includegraphics[width=0.245\linewidth]{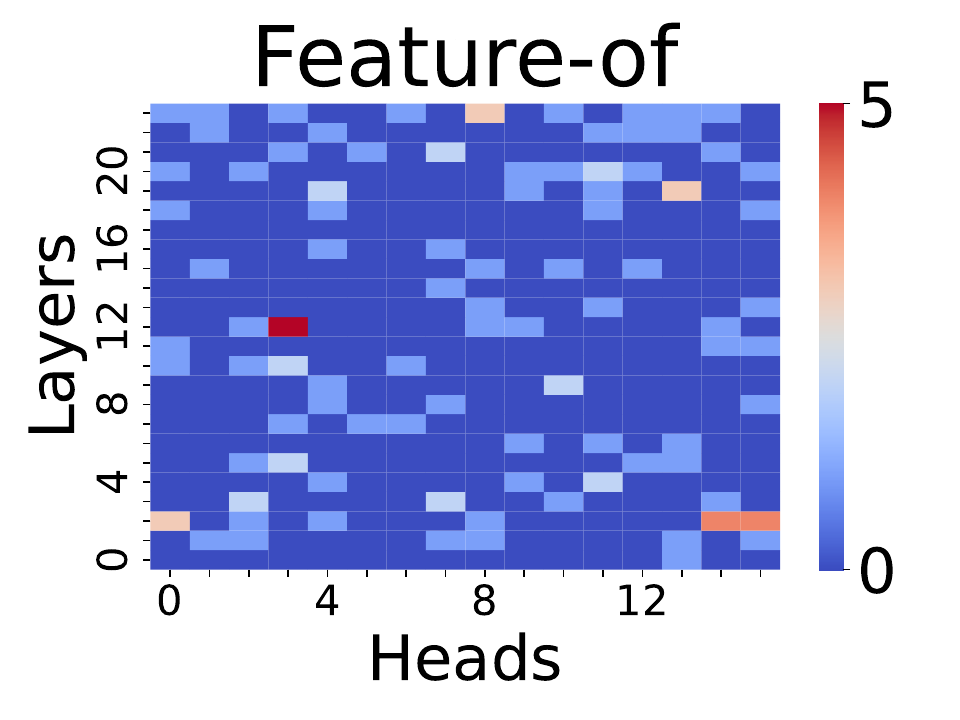}
    \includegraphics[width=0.245\linewidth]{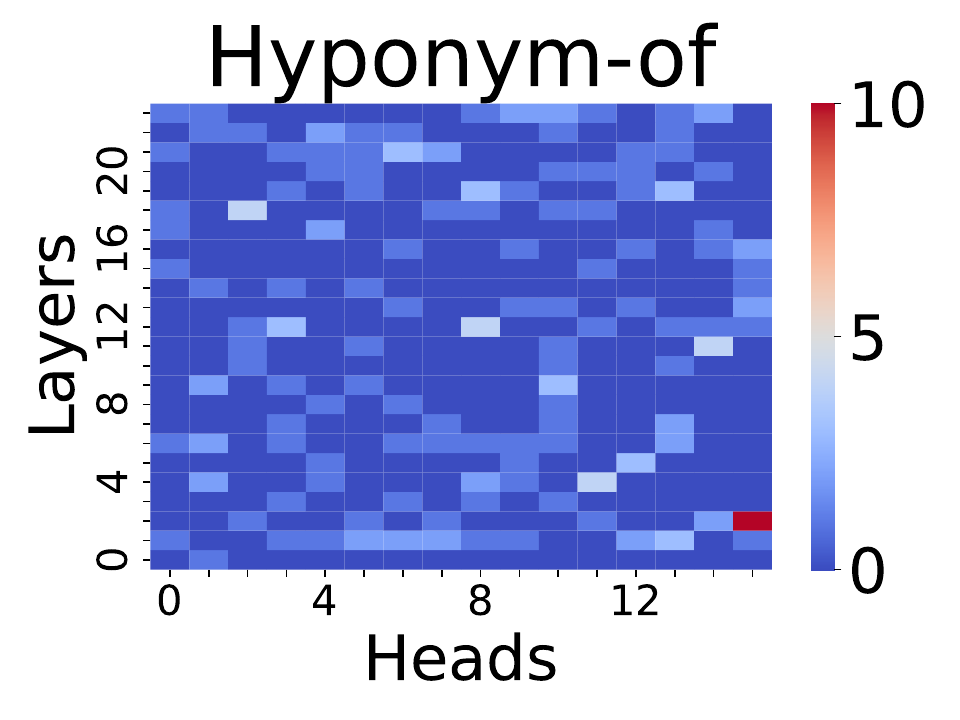}
    \includegraphics[width=0.245\linewidth]{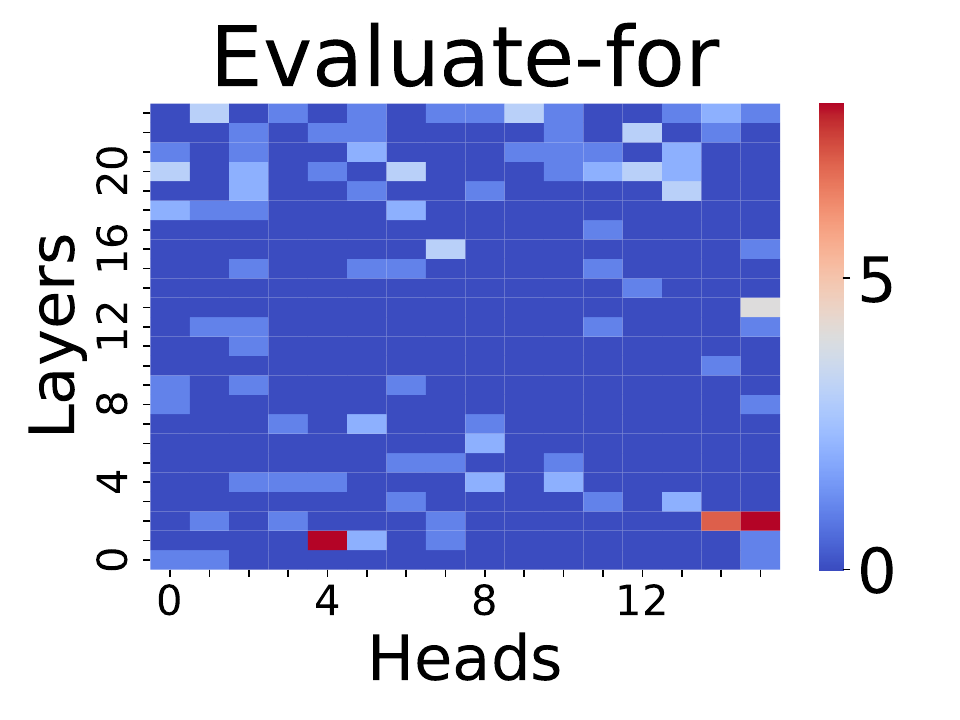}
    \includegraphics[width=0.245\linewidth]{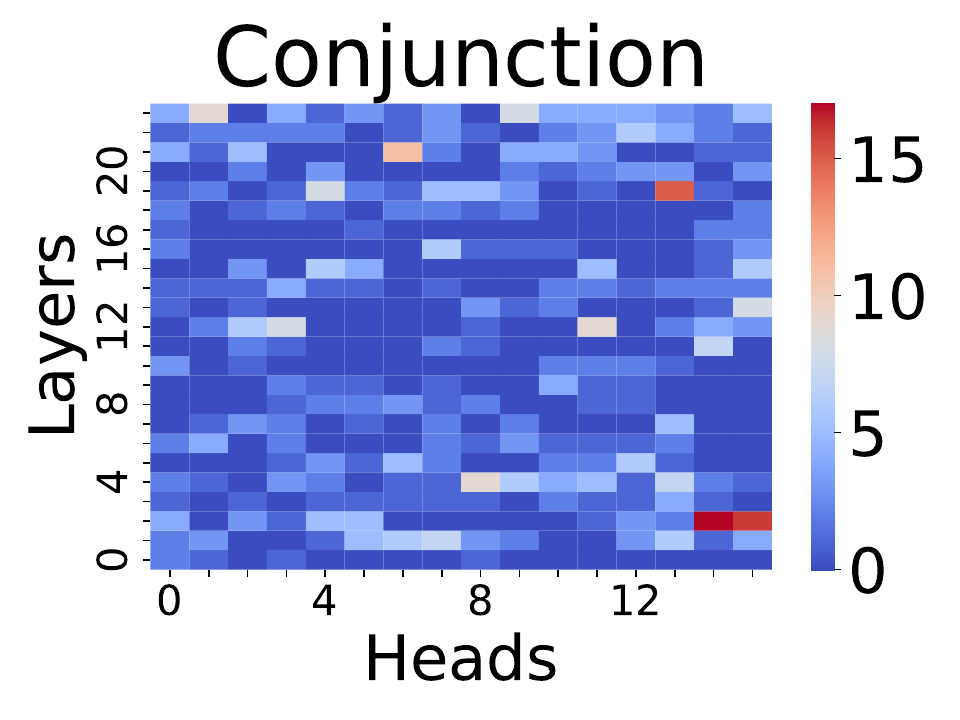}
    \caption{Heatmaps of occurrence of attention heads having the largest value of $\mathbb{E}_j[a^{h,j}_T]$ for each triplet in reverse semantic relations.}
    \label{fig: vis_of_reverse_relation_group}
\end{figure*}

Figure~\ref{fig: vis_of_relation_group} and Figure~\ref{fig: vis_of_reverse_relation_group} show the occurrence times of each attention head having the largest value of $\mathbb{E}_j[a^{h,j}_T]$ for triplets in semantic relations and the reverse triplets.
We observe that some attention heads are commonly highlighted in different types of relationships.

%% file: acl_latex.bbl
\begin{thebibliography}{43}
\expandafter\ifx\csname natexlab\endcsname\relax\def\natexlab#1{#1}\fi

\bibitem[{Bansal et~al.(2023)Bansal, Gopalakrishnan, Dingliwal, Bodapati, Kirchhoff, and Roth}]{bansal-etal-2023-rethinking}
Hritik Bansal, Karthik Gopalakrishnan, Saket Dingliwal, Sravan Bodapati, Katrin Kirchhoff, and Dan Roth. 2023.
\newblock \href {https://doi.org/10.18653/v1/2023.acl-long.660} {Rethinking the role of scale for in-context learning: An interpretability-based case study at 66 billion scale}.
\newblock In \emph{Proceedings of the 61st Annual Meeting of the Association for Computational Linguistics (Volume 1: Long Papers)}, pages 11833--11856, Toronto, Canada. Association for Computational Linguistics.

\bibitem[{Brown et~al.(2020)Brown, Mann, Ryder, Subbiah, Kaplan, Dhariwal, Neelakantan, Shyam, Sastry, Askell et~al.}]{brown2020language}
Tom Brown, Benjamin Mann, Nick Ryder, Melanie Subbiah, Jared~D Kaplan, Prafulla Dhariwal, Arvind Neelakantan, Pranav Shyam, Girish Sastry, Amanda Askell, et~al. 2020.
\newblock Language models are few-shot learners.
\newblock \emph{Advances in neural information processing systems}, 33:1877--1901.

\bibitem[{Bubeck et~al.(2023)Bubeck, Chandrasekaran, Eldan, Gehrke, Horvitz, Kamar, Lee, Lee, Li, Lundberg et~al.}]{bubeck2023sparks}
S{\'e}bastien Bubeck, Varun Chandrasekaran, Ronen Eldan, Johannes Gehrke, Eric Horvitz, Ece Kamar, Peter Lee, Yin~Tat Lee, Yuanzhi Li, Scott Lundberg, et~al. 2023.
\newblock Sparks of artificial general intelligence: Early experiments with gpt-4.
\newblock \emph{arXiv preprint arXiv:2303.12712}.

\bibitem[{Cammarata et~al.(2020)Cammarata, Carter, Goh, Olah, Petrov, Schubert, Voss, Egan, and Lim}]{cammarata2020thread}
Nick Cammarata, Shan Carter, Gabriel Goh, Chris Olah, Michael Petrov, Ludwig Schubert, Chelsea Voss, Ben Egan, and Swee~Kiat Lim. 2020.
\newblock \href {https://doi.org/10.23915/distill.00024} {Thread: Circuits}.
\newblock \emph{Distill}.
\newblock Https://distill.pub/2020/circuits.

\bibitem[{Carlini et~al.(2021)Carlini, Tram{\`{e}}r, Wallace, Jagielski, Herbert{-}Voss, Lee, Roberts, Brown, Song, Erlingsson, Oprea, and Raffel}]{carlini2021extracting}
Nicholas Carlini, Florian Tram{\`{e}}r, Eric Wallace, Matthew Jagielski, Ariel Herbert{-}Voss, Katherine Lee, Adam Roberts, Tom~B. Brown, Dawn Song, {\'{U}}lfar Erlingsson, Alina Oprea, and Colin Raffel. 2021.
\newblock Extracting training data from large language models.
\newblock In \emph{{USENIX} Security Symposium}, pages 2633--2650. {USENIX} Association.

\bibitem[{Chomsky(1957)}]{Chomsky1957}
Noam Chomsky. 1957.
\newblock \href {https://doi.org/doi:10.1515/9783112316009} {\emph{Syntactic Structures}}.
\newblock De Gruyter Mouton, Berlin, Boston.

\bibitem[{Clark et~al.(2019)Clark, Khandelwal, Levy, and Manning}]{clark-etal-2019-bert}
Kevin Clark, Urvashi Khandelwal, Omer Levy, and Christopher~D. Manning. 2019.
\newblock \href {https://doi.org/10.18653/v1/W19-4828} {What does {BERT} look at? an analysis of {BERT}{'}s attention}.
\newblock In \emph{Proceedings of the 2019 ACL Workshop BlackboxNLP: Analyzing and Interpreting Neural Networks for NLP}, pages 276--286, Florence, Italy. Association for Computational Linguistics.

\bibitem[{Das et~al.(2022)Das, Mangrulkar, Manchanda, Kapadnis, and Patnaik}]{das-etal-2022-enolp}
Millon Das, Archit Mangrulkar, Ishan Manchanda, Manav Kapadnis, and Sohan Patnaik. 2022.
\newblock \href {https://aclanthology.org/2022.smm4h-1.42} {Enolp musk@{SMM}4{H}{'}22 : Leveraging pre-trained language models for stance and premise classification}.
\newblock In \emph{Proceedings of The Seventh Workshop on Social Media Mining for Health Applications, Workshop {\&} Shared Task}, pages 156--159, Gyeongju, Republic of Korea. Association for Computational Linguistics.

\bibitem[{Elhage et~al.(2021)Elhage, Nanda, Olsson, Henighan, Joseph, Mann, Askell, Bai, Chen, Conerly, DasSarma, Drain, Ganguli, Hatfield-Dodds, Hernandez, Jones, Kernion, Lovitt, Ndousse, Amodei, Brown, Clark, Kaplan, McCandlish, and Olah}]{elhage2021mathematical}
Nelson Elhage, Neel Nanda, Catherine Olsson, Tom Henighan, Nicholas Joseph, Ben Mann, Amanda Askell, Yuntao Bai, Anna Chen, Tom Conerly, Nova DasSarma, Dawn Drain, Deep Ganguli, Zac Hatfield-Dodds, Danny Hernandez, Andy Jones, Jackson Kernion, Liane Lovitt, Kamal Ndousse, Dario Amodei, Tom Brown, Jack Clark, Jared Kaplan, Sam McCandlish, and Chris Olah. 2021.
\newblock A mathematical framework for transformer circuits.
\newblock \emph{Transformer Circuits Thread}.

\bibitem[{Geva et~al.(2023)Geva, Bastings, Filippova, and Globerson}]{geva-etal-2023-dissecting}
Mor Geva, Jasmijn Bastings, Katja Filippova, and Amir Globerson. 2023.
\newblock \href {https://doi.org/10.18653/v1/2023.emnlp-main.751} {Dissecting recall of factual associations in auto-regressive language models}.
\newblock In \emph{Proceedings of the 2023 Conference on Empirical Methods in Natural Language Processing}, pages 12216--12235, Singapore. Association for Computational Linguistics.

\bibitem[{Honnibal et~al.(2020)Honnibal, Montani, Van~Landeghem, and Boyd}]{Honnibal2020spaCy}
Matthew Honnibal, Ines Montani, Sofie Van~Landeghem, and Adriane Boyd. 2020.
\newblock \href {https://doi.org/10.5281/zenodo.1212303} {{spaCy: Industrial-strength Natural Language Processing in Python}}.

\bibitem[{Kaplan et~al.(2020)Kaplan, McCandlish, Henighan, Brown, Chess, Child, Gray, Radford, Wu, and Amodei}]{kaplan2020scaling}
Jared Kaplan, Sam McCandlish, Tom Henighan, Tom~B Brown, Benjamin Chess, Rewon Child, Scott Gray, Alec Radford, Jeffrey Wu, and Dario Amodei. 2020.
\newblock Scaling laws for neural language models.
\newblock \emph{arXiv preprint arXiv:2001.08361}.

\bibitem[{Kim et~al.(2018)Kim, Wattenberg, Gilmer, Cai, Wexler, Viegas, and sayres}]{kim2018tcav}
Been Kim, Martin Wattenberg, Justin Gilmer, Carrie Cai, James Wexler, Fernanda Viegas, and Rory sayres. 2018.
\newblock Interpretability beyond feature attribution: Quantitative testing with concept activation vectors ({TCAV}).
\newblock In \emph{Proceedings of the 35th International Conference on Machine Learning}, volume~80 of \emph{Proceedings of Machine Learning Research}, pages 2668--2677. PMLR.

\bibitem[{Koncel-Kedziorski et~al.(2019)Koncel-Kedziorski, Bekal, Luan, Lapata, and Hajishirzi}]{koncel-kedziorski-etal-2019-text}
Rik Koncel-Kedziorski, Dhanush Bekal, Yi~Luan, Mirella Lapata, and Hannaneh Hajishirzi. 2019.
\newblock \href {https://doi.org/10.18653/v1/N19-1238} {{T}ext {G}eneration from {K}nowledge {G}raphs with {G}raph {T}ransformers}.
\newblock In \emph{Proceedings of the 2019 Conference of the North {A}merican Chapter of the Association for Computational Linguistics: Human Language Technologies, Volume 1 (Long and Short Papers)}, pages 2284--2293, Minneapolis, Minnesota. Association for Computational Linguistics.

\bibitem[{Li et~al.(2015)Li, Chen, Hovy, and Jurafsky}]{li2015visualizing}
Jiwei Li, Xinlei Chen, Eduard Hovy, and Dan Jurafsky. 2015.
\newblock Visualizing and understanding neural models in nlp.
\newblock \emph{arXiv preprint arXiv:1506.01066}.

\bibitem[{Li et~al.(2018)Li, Liu, Chen, and Rudin}]{oscar2018deep}
Oscar Li, Hao Liu, Chaofan Chen, and Cynthia Rudin. 2018.
\newblock Deep learning for case-based reasoning through prototypes: a neural network that explains its predictions.
\newblock AAAI'18/IAAI'18/EAAI'18. AAAI Press.

\bibitem[{Lieberum et~al.(2023)Lieberum, Rahtz, Kram{\'a}r, Irving, Shah, and Mikulik}]{lieberum2023does}
Tom Lieberum, Matthew Rahtz, J{\'a}nos Kram{\'a}r, Geoffrey Irving, Rohin Shah, and Vladimir Mikulik. 2023.
\newblock Does circuit analysis interpretability scale? evidence from multiple choice capabilities in chinchilla.
\newblock \emph{arXiv preprint arXiv:2307.09458}.

\bibitem[{Liu et~al.(2024)Liu, Deng, Cheng, Ren, Wang, and Zhang}]{liu2024towards}
Dongrui Liu, Huiqi Deng, Xu~Cheng, Qihan Ren, Kangrui Wang, and Quanshi Zhang. 2024.
\newblock Towards the difficulty for a deep neural network to learn concepts of different complexities.
\newblock \emph{Advances in Neural Information Processing Systems}, 36.

\bibitem[{Lundberg and Lee(2017)}]{lundberg2017unified}
Scott~M. Lundberg and Su{-}In Lee. 2017.
\newblock A unified approach to interpreting model predictions.
\newblock In \emph{Advances in Neural Information Processing}, pages 4765--4774.

\bibitem[{Manakul et~al.(2023)Manakul, Liusie, and Gales}]{manakul-etal-2023-selfcheckgpt}
Potsawee Manakul, Adian Liusie, and Mark Gales. 2023.
\newblock \href {https://doi.org/10.18653/v1/2023.emnlp-main.557} {{S}elf{C}heck{GPT}: Zero-resource black-box hallucination detection for generative large language models}.
\newblock In \emph{Proceedings of the 2023 Conference on Empirical Methods in Natural Language Processing}, pages 9004--9017, Singapore. Association for Computational Linguistics.

\bibitem[{Modarressi et~al.(2023)Modarressi, Fayyaz, Aghazadeh, Yaghoobzadeh, and Pilehvar}]{modarressi-etal-2023-decompx}
Ali Modarressi, Mohsen Fayyaz, Ehsan Aghazadeh, Yadollah Yaghoobzadeh, and Mohammad~Taher Pilehvar. 2023.
\newblock \href {https://doi.org/10.18653/v1/2023.acl-long.149} {{D}ecomp{X}: Explaining transformers decisions by propagating token decomposition}.
\newblock In \emph{Proceedings of the 61st Annual Meeting of the Association for Computational Linguistics (Volume 1: Long Papers)}, pages 2649--2664, Toronto, Canada. Association for Computational Linguistics.

\bibitem[{Mohebbi et~al.(2023)Mohebbi, Zuidema, Chrupa{\l}a, and Alishahi}]{mohebbi-etal-2023-quantifying}
Hosein Mohebbi, Willem Zuidema, Grzegorz Chrupa{\l}a, and Afra Alishahi. 2023.
\newblock \href {https://doi.org/10.18653/v1/2023.eacl-main.245} {Quantifying context mixing in transformers}.
\newblock In \emph{Proceedings of the 17th Conference of the European Chapter of the Association for Computational Linguistics}, pages 3378--3400, Dubrovnik, Croatia. Association for Computational Linguistics.

\bibitem[{Olsson et~al.(2022)Olsson, Elhage, Nanda, Joseph, DasSarma, Henighan, Mann, Askell, Bai, Chen, Conerly, Drain, Ganguli, Hatfield-Dodds, Hernandez, Johnston, Jones, Kernion, Lovitt, Ndousse, Amodei, Brown, Clark, Kaplan, McCandlish, and Olah}]{olsson2022context}
Catherine Olsson, Nelson Elhage, Neel Nanda, Nicholas Joseph, Nova DasSarma, Tom Henighan, Ben Mann, Amanda Askell, Yuntao Bai, Anna Chen, Tom Conerly, Dawn Drain, Deep Ganguli, Zac Hatfield-Dodds, Danny Hernandez, Scott Johnston, Andy Jones, Jackson Kernion, Liane Lovitt, Kamal Ndousse, Dario Amodei, Tom Brown, Jack Clark, Jared Kaplan, Sam McCandlish, and Chris Olah. 2022.
\newblock In-context learning and induction heads.
\newblock \emph{Transformer Circuits Thread}.

\bibitem[{Park et~al.(2019)Park, Na, Jo, Shin, Yoo, Kwon, Zhao, Noh, Lee, and Choo}]{park2019sanvis}
Cheonbok Park, Inyoup Na, Yongjang Jo, Sungbok Shin, Jaehyo Yoo, Bum~Chul Kwon, Jian Zhao, Hyungjong Noh, Yeonsoo Lee, and Jaegul Choo. 2019.
\newblock \href {https://doi.org/10.1109/VISUAL.2019.8933677} {Sanvis: Visual analytics for understanding self-attention networks}.
\newblock In \emph{2019 IEEE Visualization Conference (VIS)}, pages 146--150.

\bibitem[{Petroni et~al.(2019)Petroni, Rockt{\"a}schel, Riedel, Lewis, Bakhtin, Wu, and Miller}]{petroni-etal-2019-language}
Fabio Petroni, Tim Rockt{\"a}schel, Sebastian Riedel, Patrick Lewis, Anton Bakhtin, Yuxiang Wu, and Alexander Miller. 2019.
\newblock \href {https://doi.org/10.18653/v1/D19-1250} {Language models as knowledge bases?}
\newblock In \emph{Proceedings of the 2019 Conference on Empirical Methods in Natural Language Processing and the 9th International Joint Conference on Natural Language Processing (EMNLP-IJCNLP)}, pages 2463--2473, Hong Kong, China. Association for Computational Linguistics.

\bibitem[{Qian et~al.(2024)Qian, Zhang, Yao, Liu, Yin, Qiao, Liu, and Shao}]{qian2024towards}
Chen Qian, Jie Zhang, Wei Yao, Dongrui Liu, Zhenfei Yin, Yu~Qiao, Yong Liu, and Jing Shao. 2024.
\newblock Towards tracing trustworthiness dynamics: Revisiting pre-training period of large language models.
\newblock \emph{arXiv preprint arXiv:2402.19465}.

\bibitem[{Ren et~al.(2023)Ren, Li, Chen, Deng, and Zhang}]{ren2023defining}
Jie Ren, Mingjie Li, Qirui Chen, Huiqi Deng, and Quanshi Zhang. 2023.
\newblock Defining and quantifying the emergence of sparse concepts in dnns.
\newblock In \emph{{CVPR}}, pages 20280--20289. {IEEE}.

\bibitem[{Ren et~al.(2021)Ren, Zhang, Wang, Chen, Zhou, Chen, Cheng, Wang, Zhou, Shi, and Zhang}]{ren2021towards}
Jie Ren, Die Zhang, Yisen Wang, Lu~Chen, Zhanpeng Zhou, Yiting Chen, Xu~Cheng, Xin Wang, Meng Zhou, Jie Shi, and Quanshi Zhang. 2021.
\newblock Towards a unified game-theoretic view of adversarial perturbations and robustness.
\newblock In \emph{NeurIPS}, volume~34, pages 3797--3810.

\bibitem[{Ren et~al.(2024)Ren, Gao, Shao, Yan, Tan, Lam, and Ma}]{ren2024exploring}
Qibing Ren, Chang Gao, Jing Shao, Junchi Yan, Xin Tan, Wai Lam, and Lizhuang Ma. 2024.
\newblock Exploring safety generalization challenges of large language models via code.
\newblock \emph{arxiv preprint arxiv:2403.07865}.

\bibitem[{Ribeiro et~al.(2016)Ribeiro, Singh, and Guestrin}]{ribeiro2016should}
Marco~Tulio Ribeiro, Sameer Singh, and Carlos Guestrin. 2016.
\newblock " why should i trust you?" explaining the predictions of any classifier.
\newblock In \emph{Proceedings of the 22nd ACM SIGKDD international conference on knowledge discovery and data mining}, pages 1135--1144.

\bibitem[{Simonyan et~al.(2013)Simonyan, Vedaldi, and Zisserman}]{simonyan2013deep}
Karen Simonyan, Andrea Vedaldi, and Andrew Zisserman. 2013.
\newblock Deep inside convolutional networks: Visualising image classification models and saliency maps.
\newblock \emph{arXiv preprint arXiv:1312.6034}.

\bibitem[{Soboleva et~al.(2023)Soboleva, Al-Khateeb, Myers, Steeves, Hestness, and Dey}]{cerebras2023slimpajama}
Daria Soboleva, Faisal Al-Khateeb, Robert Myers, Jacob~R Steeves, Joel Hestness, and Nolan Dey. 2023.
\newblock \href {https://huggingface.co/datasets/cerebras/SlimPajama-627B} {{SlimPajama: A 627B token cleaned and deduplicated version of RedPajama}}.

\bibitem[{Sundararajan et~al.(2017)Sundararajan, Taly, and Yan}]{sundararajan2017axiomatic}
Mukund Sundararajan, Ankur Taly, and Qiqi Yan. 2017.
\newblock Axiomatic attribution for deep networks.
\newblock In \emph{Proceedings of the 34th International Conference on Machine Learning, {ICML}}, volume~70, pages 3319--3328. {PMLR}.

\bibitem[{Team(2023)}]{2023internlm}
InternLM Team. 2023.
\newblock Internlm: A multilingual language model with progressively enhanced capabilities.
\newblock \url{https://github.com/InternLM/InternLM}.

\bibitem[{Thomas et~al.(2023)Thomas, Agustin, Louis, Thibaut, David, Julien, Rémi, and Thomas}]{fel2023craft}
Fel Thomas, Picard Agustin, Bethune Louis, Boissin Thibaut, Vigouroux David, Colin Julien, Cadène Rémi, and Serre Thomas. 2023.
\newblock Craft: Concept recursive activation factorization for explainability.
\newblock In \emph{Proceedings of the IEEE Conference on Computer Vision and Pattern Recognition (CVPR)}.

\bibitem[{Touvron et~al.(2023)Touvron, Lavril, Izacard, Martinet, Lachaux, Lacroix, Rozi{\`e}re, Goyal, Hambro, Azhar et~al.}]{touvron2023llama}
Hugo Touvron, Thibaut Lavril, Gautier Izacard, Xavier Martinet, Marie-Anne Lachaux, Timoth{\'e}e Lacroix, Baptiste Rozi{\`e}re, Naman Goyal, Eric Hambro, Faisal Azhar, et~al. 2023.
\newblock Llama: Open and efficient foundation language models.
\newblock \emph{arXiv preprint arXiv:2302.13971}.

\bibitem[{Vig(2019)}]{vig-2019-multiscale}
Jesse Vig. 2019.
\newblock \href {https://doi.org/10.18653/v1/P19-3007} {A multiscale visualization of attention in the transformer model}.
\newblock In \emph{Proceedings of the 57th Annual Meeting of the Association for Computational Linguistics: System Demonstrations}, pages 37--42, Florence, Italy. Association for Computational Linguistics.

\bibitem[{Wang et~al.(2023)Wang, Variengien, Conmy, Shlegeris, and Steinhardt}]{wang2023interpretability}
Kevin~Ro Wang, Alexandre Variengien, Arthur Conmy, Buck Shlegeris, and Jacob Steinhardt. 2023.
\newblock Interpretability in the wild: a circuit for indirect object identification in {GPT}-2 small.
\newblock In \emph{The Eleventh International Conference on Learning Representations}.

\bibitem[{Wei et~al.(2022)Wei, Tay, Bommasani, Raffel, Zoph, Borgeaud, Yogatama, Bosma, Zhou, Metzler, Chi, Hashimoto, Vinyals, Liang, Dean, and Fedus}]{wei2022emergent}
Jason Wei, Yi~Tay, Rishi Bommasani, Colin Raffel, Barret Zoph, Sebastian Borgeaud, Dani Yogatama, Maarten Bosma, Denny Zhou, Donald Metzler, Ed~H. Chi, Tatsunori Hashimoto, Oriol Vinyals, Percy Liang, Jeff Dean, and William Fedus. 2022.
\newblock Emergent abilities of large language models.
\newblock \emph{Trans. Mach. Learn. Res.}, 2022.

\bibitem[{Yang et~al.(2023)Yang, Huang, Zou, Zhang, Dai, and Chen}]{yang-etal-2023-local}
Sen Yang, Shujian Huang, Wei Zou, Jianbing Zhang, Xinyu Dai, and Jiajun Chen. 2023.
\newblock \href {https://doi.org/10.18653/v1/2023.acl-long.572} {Local interpretation of transformer based on linear decomposition}.
\newblock In \emph{Proceedings of the 61st Annual Meeting of the Association for Computational Linguistics (Volume 1: Long Papers)}, pages 10270--10287, Toronto, Canada. Association for Computational Linguistics.

\bibitem[{Yeh et~al.(2024)Yeh, Chen, Wu, Chen, Viégas, and Wattenberg}]{yeh2023attentionviz}
Catherine Yeh, Yida Chen, Aoyu Wu, Cynthia Chen, Fernanda Viégas, and Martin Wattenberg. 2024.
\newblock \href {https://doi.org/10.1109/TVCG.2023.3327163} {Attentionviz: A global view of transformer attention}.
\newblock \emph{IEEE Transactions on Visualization and Computer Graphics}, 30(1):262--272.

\bibitem[{Zhang et~al.(2022)Zhang, Wang, Chen, and Zhang}]{zhang-etal-2022-probing}
Lining Zhang, Mengchen Wang, Liben Chen, and Wenxin Zhang. 2022.
\newblock \href {https://doi.org/10.18653/v1/2022.blackboxnlp-1.24} {Probing {GPT}-3{'}s linguistic knowledge on semantic tasks}.
\newblock In \emph{Proceedings of the Fifth BlackboxNLP Workshop on Analyzing and Interpreting Neural Networks for NLP}, pages 297--304, Abu Dhabi, United Arab Emirates (Hybrid). Association for Computational Linguistics.

\bibitem[{Zhou et~al.(2024)Zhou, Zhang, Deng, Liu, Shen, Chan, and Zhang}]{zhou2024explaining}
Huilin Zhou, Hao Zhang, Huiqi Deng, Dongrui Liu, Wen Shen, Shih-Han Chan, and Quanshi Zhang. 2024.
\newblock Explaining generalization power of a dnn using interactive concepts.
\newblock In \emph{Proceedings of the AAAI Conference on Artificial Intelligence}, volume~38, pages 17105--17113.

\end{thebibliography}
